\documentclass[10pt,twocolumn,letterpaper]{article}

\usepackage{cvpr}
\usepackage{times}
\usepackage{epsfig}
\usepackage{graphicx}
\usepackage{amsmath}
\usepackage{amssymb}

\usepackage[pagebackref=true,breaklinks=true,letterpaper=true,colorlinks,bookmarks=false]{hyperref}

\cvprfinalcopy % *** Uncomment this line for the final submission

\def\cvprPaperID{9} % *** Enter the CVPR Paper ID here

\ifcvprfinal\fi
\begin{document}

%%%%%%%%% TITLE
\title{Unsupervised Domain Adaptation for Multispectral Pedestrian Detection}

%\author{Dayan Guan \and Xing Luo \and Yanpeng Cao \and Jiangxin Yang \and Yanlong Cao \and
%	Zhejiang University, {\tt\small caopy@zju.edu.cn}
%	% For a paper whose authors are all at the same institution,
%	% omit the following lines up until the closing ``}''.
%	% Additional authors and addresses can be added with ``\and'',
%	% just like the second author.
%	% To save space, use either the email address or home page, not both
%	\and 
%	George Vosselman \and Michael Ying Yang \and
%	University of Twente, {\tt\small michael.yang@utwente.nl}
%}

\author{Dayan Guan$^{1}$ \and Xing Luo$^{1}$ \and Yanpeng Cao$^{1}$ \and Jiangxin Yang$^{1}$ \and Yanlong Cao$^{1}$ \and George Vosselman$^{2}$ \quad Michael Ying Yang$^{2}$\\ 
	$^{1}$Zhejiang University, {\tt\small \{11725001, luoxing, caopy, yangjx, sdcaoyl\}@zju.edu.cn}\\
	$^{2}$University of Twente, {\tt\small \{george.vosselman, michael.yang\}@utwente.nl}
}

\maketitle
%\thispagestyle{empty}

%%%%%%%%% ABSTRACT
\begin{abstract}

Multimodal information (e.g., visible and thermal) can generate robust pedestrian detections to facilitate around-the-clock computer vision applications, such as autonomous driving and video surveillance. However, it still remains a crucial challenge to train a reliable detector working well in different multispectral pedestrian datasets without manual annotations. In this paper, we propose a novel unsupervised domain adaptation framework for multispectral pedestrian detection, by iteratively generating pseudo annotations and updating the parameters of our designed multispectral pedestrian detector on target domain. Pseudo annotations are generated using the detector trained on source domain, and then updated by fixing the parameters of detector and minimizing the cross entropy loss without back-propagation. Training labels are generated using the pseudo annotations by considering the characteristics of similarity and complementarity between well-aligned visible and infrared image pairs. The parameters of detector are updated using the generated labels by minimizing our defined multi-detection loss function with back-propagation. The optimal parameters of detector can be obtained after iteratively updating the pseudo annotations and parameters. Experimental results show that our proposed unsupervised multimodal domain adaptation method achieves significantly higher detection performance than the approach without domain adaptation, and is competitive with the supervised multispectral pedestrian detectors.

\end{abstract}

%%%%%%%%% BODY TEXT

\begin{figure}[t]
	\centering
	\begin{minipage}{0.9\linewidth}
		\begin{minipage}{0.49\linewidth}
			\includegraphics[width=0.99\linewidth,clip]{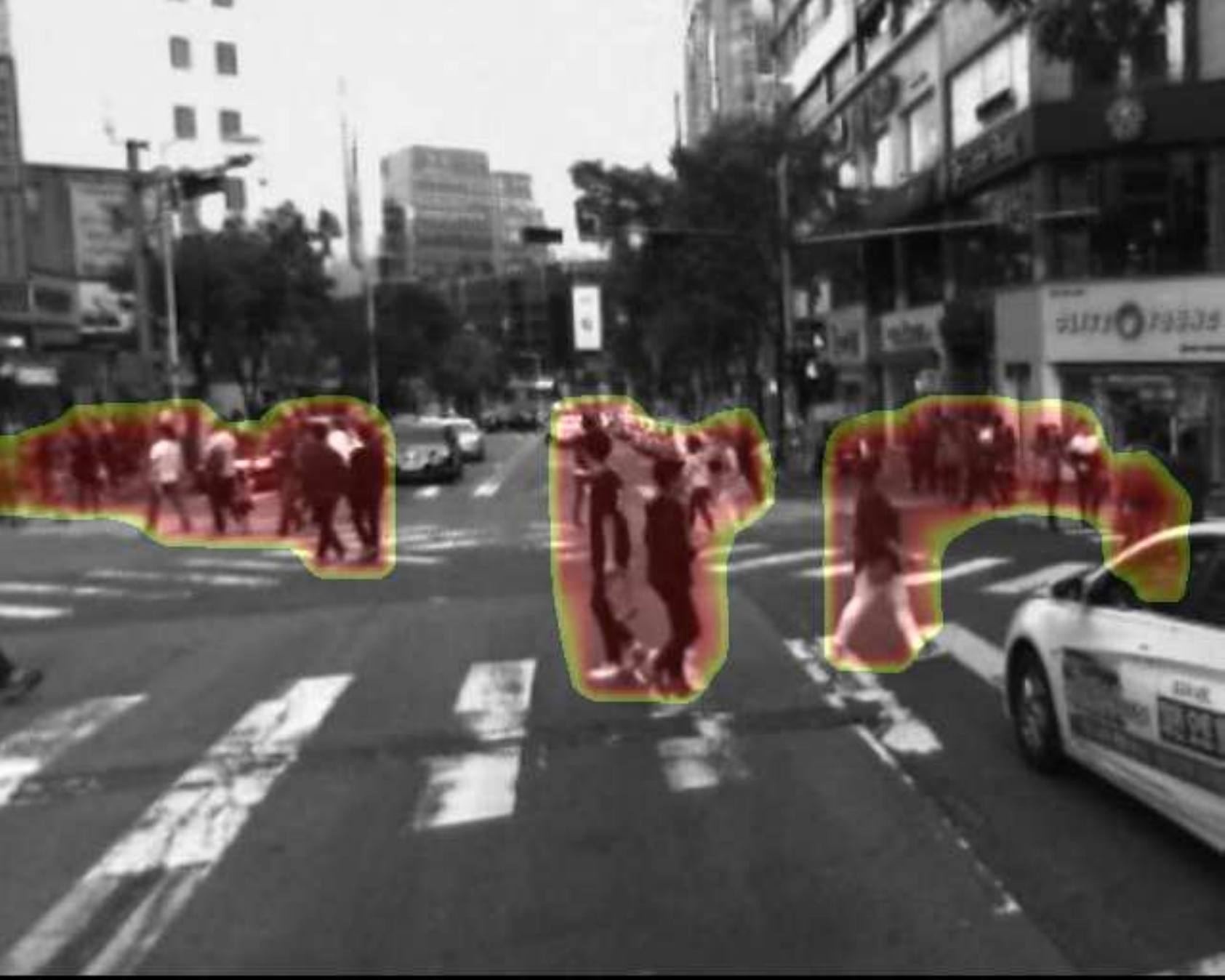}
		\end{minipage}
		\begin{minipage}{0.49\linewidth}
			\includegraphics[width=0.99\linewidth,clip]{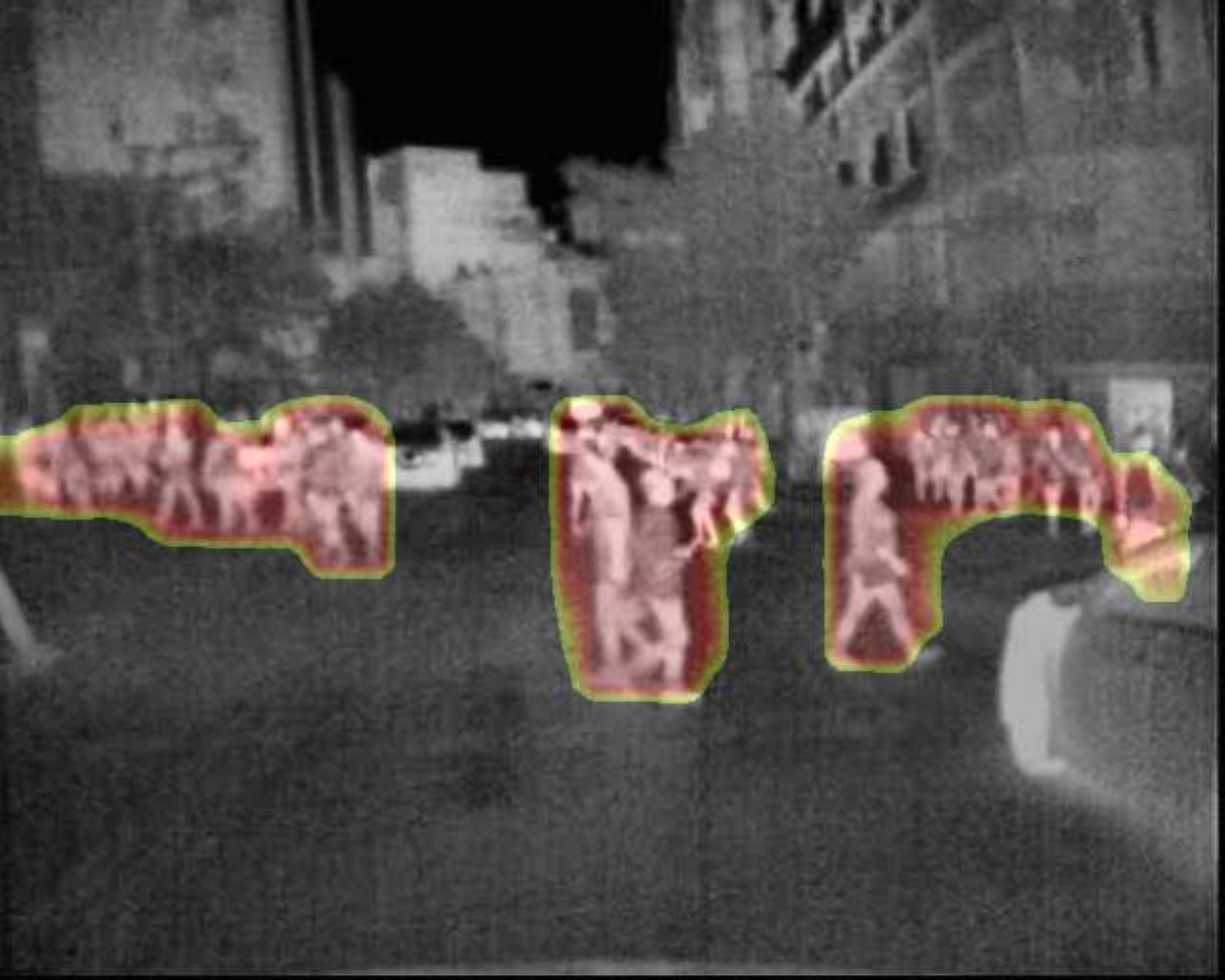}
		\end{minipage}
		\centering{(a)}
		\vspace{1mm}
	\end{minipage}
	\begin{minipage}{0.9\linewidth}
		\begin{minipage}{0.49\linewidth}
			\includegraphics[width=0.99\linewidth,clip]{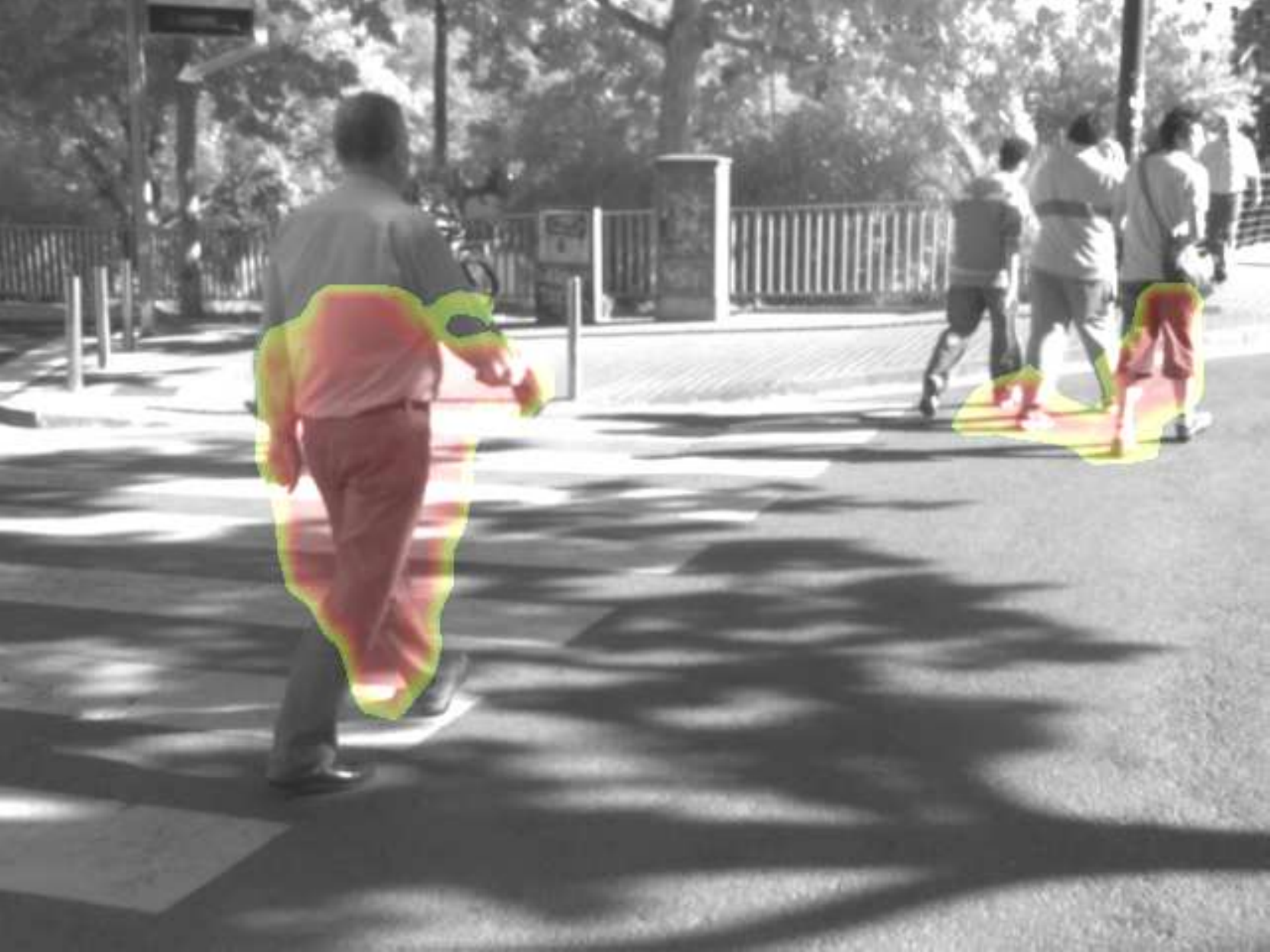}
		\end{minipage}
		\begin{minipage}{0.49\linewidth}
			\includegraphics[width=0.99\linewidth,clip]{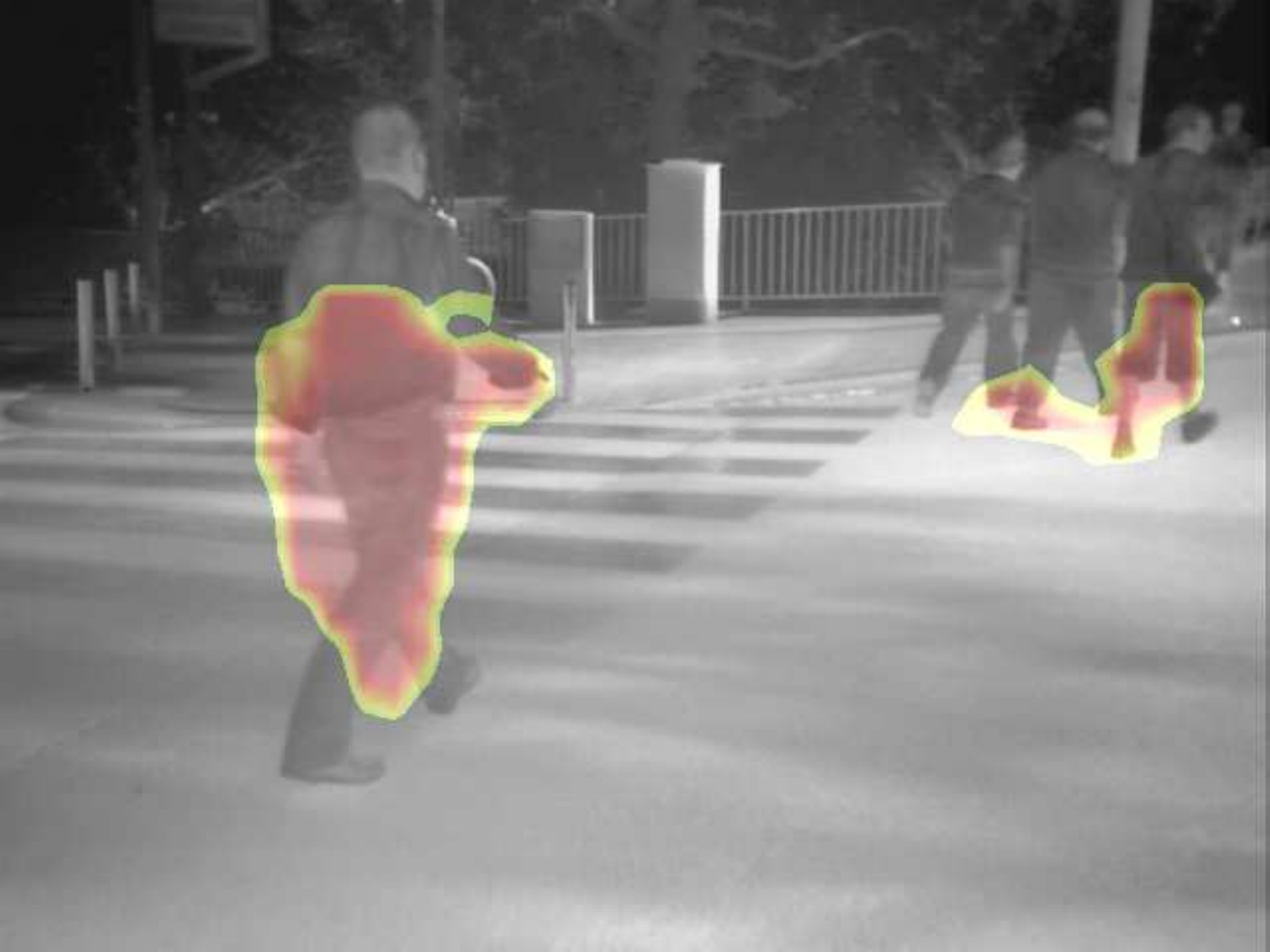}
		\end{minipage}
		\centering{(b)}
		\vspace{1mm}
	\end{minipage}
	\caption{Detection results of the current state-of-the-art multispectral pedestrian detector well-trained using visible and thermal image pairs from KAIST~\cite{hwang2015multispectral} dataset following the method presented by Cao~\emph{et al.}~\cite{cao2019box}. (a) Results on the KAIST dataset; (b) Results on the CVC-14~\cite{gonzalez2016pedestrian} dataset. Please note that the visible images in KAIST dataset are transferred from RGB to gray level in order to decrease domain differences between these two datasets.}
	\label{fig1}
\end{figure}

\begin{figure*}[ht]
	\begin{center}
		\includegraphics[width=0.99\linewidth]{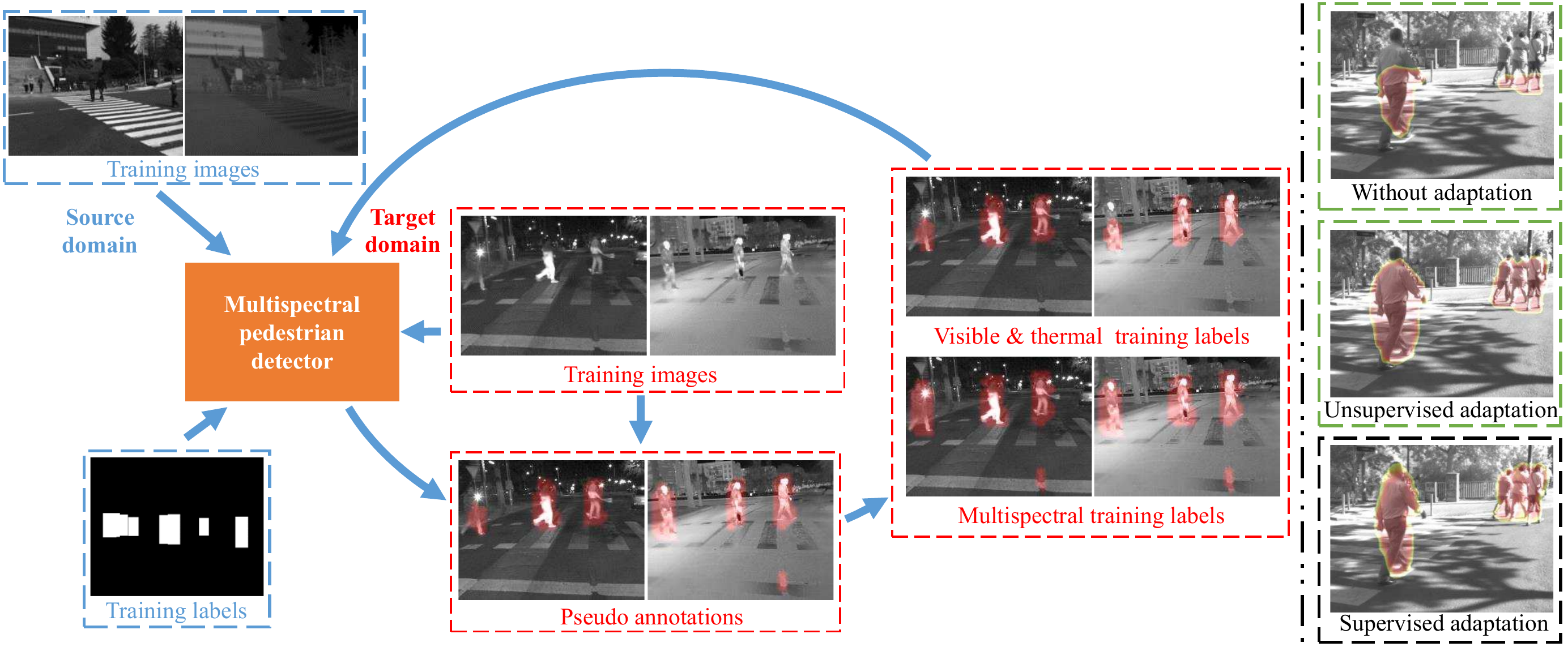}
	\end{center}
	\caption{Illustration of our proposed unsupervised domain adaptation framework for multispectral pedestrian detection. Pseudo annotations are generated using the detector trained on source domain, and then updated by fixing the parameters of detector and minimizing the cross entropy loss without back-propagation. Training labels are generated using the pseudo annotations by considering the characteristics of similarity and complementarity between well-aligned visible and infrared image pairs. The parameters of detector are updated using the generated labels by minimizing our defined multi-detection loss function with back-propagation. The optimal parameters of detector can be obtained by iteratively generating pseudo annotations and updating the parameters of our designed multispectral pedestrian detector on target domain.}
	\label{fig2}
\end{figure*}

\section{Introduction}
Pedestrian detection has become an important and popular topic within the field of computer vision community over the past few years~\cite{dollar2009pedestrian, benenson2012pedestrian, tian2015pedestrian, zhang2016far, zhang2017citypersons, zhang2018towards}. Given sensing images captured in complex and changing real-world environment, pedestrian detection solution is required to predict the regions of human. It provides significant information for various human-centric sensing applications. In order to facilitate the around-the-clock robotic applications, such as autonomous driving and video surveillance, multimodal information (e.g., visible and thermal) have applied to generate more robust and reliable pedestrian detection results in the recent years~\cite{krotosky2007color, hwang2015multispectral, liu2016multispectral, xu2017learning, konig2017fully, guan2018fusion}.

Although significant improvements have been accomplished in the research area of multispectral pedestrian detection recently, it still remains a crucial challenge to train a reliable multispectral pedestrian detector working well in different open benchmark datasets simultaneously. The detection performance of a multispectral pedestrian detector well-trained on one benchmark dataset may drop significantly when applying to another one. Specifically, we utilize the KAIST~\cite{hwang2015multispectral} and CVC-14~\cite{gonzalez2016pedestrian} benchmark datasets to display this phenomenon. Considering that the visible images from CVC-14 dataset are gray scale without much color information, we transfer the visible images in KAIST dataset from RGB to gray level in order to decrease domain differences between these two datasets. For example, as shown in Fig.~\ref{fig1}, the current state-of-the-art multispectral pedestrian detector~\cite{cao2019box} well-trained in the KAIST~\cite{hwang2015multispectral} dataset can't generate reliable detection results on the images from CVC-14~\cite{gonzalez2016pedestrian} dataset. This is because multispectral pedestrian detectors well-trained on major open dataset tend to overfit the training data, which is usually biased to specific environments \cite{zou2018unsupervised}. Different benchmark datasets exist domain differences caused by varying conditions of viewpoints, cameras, weather and etc.

To improve multispectral pedestrian detection performance on target domain, multiple-cue information should be generated to update the detector on target data. A nature idea is to annotate data on the target domain. However, densely annotating images is costly and unscalable, since the target domain might change frequently. To overcome this limitation, we propose a novel unsupervised domain adaptation framework for multispectral pedestrian detection, by iteratively generating pseudo annotations and updating the parameters of our designed multispectral pedestrian detector on target domain, as shown in Fig.~\ref{fig2}. Pseudo annotations are generated using the detector trained on source domain, and then updated by fixing the parameters of detector and minimizing the cross entropy loss without back-propagation. Training labels are generated using the pseudo annotations by considering the characteristics of similarity and complementarity between well-aligned visible and infrared image pairs. The parameters of detector are updated using the generated labels by minimizing our defined multi-detection loss function with back-propagation. We show in the experimental part that our designed unsupervised multimodal domain adaptation method achieves significantly higher detection performance than the approach without domain adaptation, which can verify the effectiveness of our proposed approach. 
Comparing with the supervised multimodal domain adaptation method which need extremely time-consuming manual annotating effort, our proposed unsupervised method barely increase the training time  because the additional processing time is caused by the optimization of pseudo annotations without back-propagation.

Overall, the \textbf{contributions} of this paper are summarized as follows:

\begin{itemize}
	\item[1] we demonstrate the usefulness of visible and thermal data for the task of unsupervised domain adaptation for multispectral pedestrian detection. Characteristics of similarity and complementarity between well-aligned visible and infrared image pairs can be used to adapt the detector trained on an annotated source domain to a target one without manual annotations. To the best of our knowledge, this is the first attempt to explore characteristics of visible and infrared images on the task of unsupervised domain adaptation for multispectral pedestrian detection.
	
	\item[2] We propose a novel unsupervised domain adaptation framework for multispectral pedestrian detection, by iteratively generating training labels and updating the parameters of our designed multispectral pedestrian detector on target domain. Training labels are generated using the pseudo annotations, which are updated by fixing the parameters of detector and minimizing the cross entropy loss without back-propagation. The parameters of detector are updated using the generated labels by minimizing our defined multi-detection loss function with back-propagation.
	
	\item[3] Our proposed unsupervised multimodal domain adaptation method achieves significantly higher detection performance than the approach without domain adaptation, and is competitive with the supervised multispectral pedestrian detectors. Comparing with the supervised approach which need extremely time-consuming manual annotating effort, our proposed unsupervised method barely increase the training time.
		
\end{itemize}

\section{Related works}

Pedestrian detection applications in intelligent robotics, urban surveillance, and self-driving vehicles have been widely spread. Constantly emerging pedestrian detectors and related improvements have accelerated its practical application. 
%Zhang \emph{et al.}~\cite{zhang2016faster} evaluated the reason why specified application of general object detection in pedestrian aspect drops significantly. The authors adopted a hybrid strategy that extracting the candidate regions utilizing region proposal networks (RPN) along with boosted classifiers, which then classifies the high-resolution convolutional human-related feature. 
Zhang \emph{et al.}~\cite{zhang2016faster} adopted a hybrid strategy that extracting the candidate regions utilizing region proposal networks~\cite{ren2015faster} along with boosted classifiers~\cite{viola2001rapid}. 
Mao \emph{et al.}~\cite{Mao2017CVPR} proposed a powerful framework which implements representations of channel features to benefit the detection by additionally learning extra features to assist inference. 
Brazil \emph{et al.}~\cite{Brazil2017ICCV} put detection and segmentation together during the training period, where it suggests that weak box-level annotations could bring benefit to the improvement of detection accuracy. 
Wang \emph{et al.}~\cite{wang2018repulsion} designed a novel repulsion loss to restrain the predicted boxes from shifting to surrounding ground truth boxes. The superior visible pedestrian detection performance had achieved with the detectors trained using the repulsion loss.

With the complementary informations given by infrared images, multispectral pedestrian detection expands the research field beyond the traditional visible images and turns to be a potential solution to shrink the gap between machine and human observers. 
Hwang \emph{et al.}~\cite{hwang2015multispectral} noticed the phenomenon and released the first large-scale multispectral pedestrian dataset (KAIST), containing well-aligned visible and infrared image pairs annotated densely. 
Liu \emph{et al.}~\cite{liu2016multispectral} methodically explored the performance of two-stream deep convolutional neural networks where the multi-information feature integrates, showing the architecture that merges two-branch features on the middle-level convolutional layers outperforms any other ones. 
K{\"o}nig \emph{et al.}~\cite{konig2017fully} adopted the architecture of RPN+BDT~\cite{zhang2016faster} in a fusion way, which merges the features generated by two-branch middle-level convolutional layers, in the purpose of multispectral pedestrian detection. 
%Furthermore, researchers also paid attention to the main difference between visible and infrared images, they heeded the illumination features of objects at daytime and nighttime varies, therefore they proposed illumination-aware weighting mechanism to give extra information to detector, attaining much advance in detection~\cite{guan2018fusion, li2018illumination}. 
Researchers also paid attention to the main difference between visible and infrared images, and proposed illumination-aware weighting mechanism to give extra information to detectors~\cite{guan2018fusion, li2018illumination}. 
Guan \emph{et al.}~\cite{guan2018exploiting} presented a unified multispectral fusion framework, which infuses the multispectral semantic segmentation masks as supervision for learning human-related features, getting more accurate detection results.  
Li \emph{et al.}~\cite{li2018multispectral} designed a cascaded multispectral classification network to distinguish hard negatives sample from pedestrian and human-like instances. 
Cao \emph{et al.}~\cite{cao2019box} developed a novel box-level segmentation supervised networks, which can generate more accurate multispectral pedestrian detections on small-size training images. Experimental results showed that their proposed approach achieved the current state-of-the-art pedestrian detection performance using visible and thermal images on both accuracy and speed.

\begin{figure*}[ht]
	\begin{center}
		\includegraphics[width=0.6\linewidth]{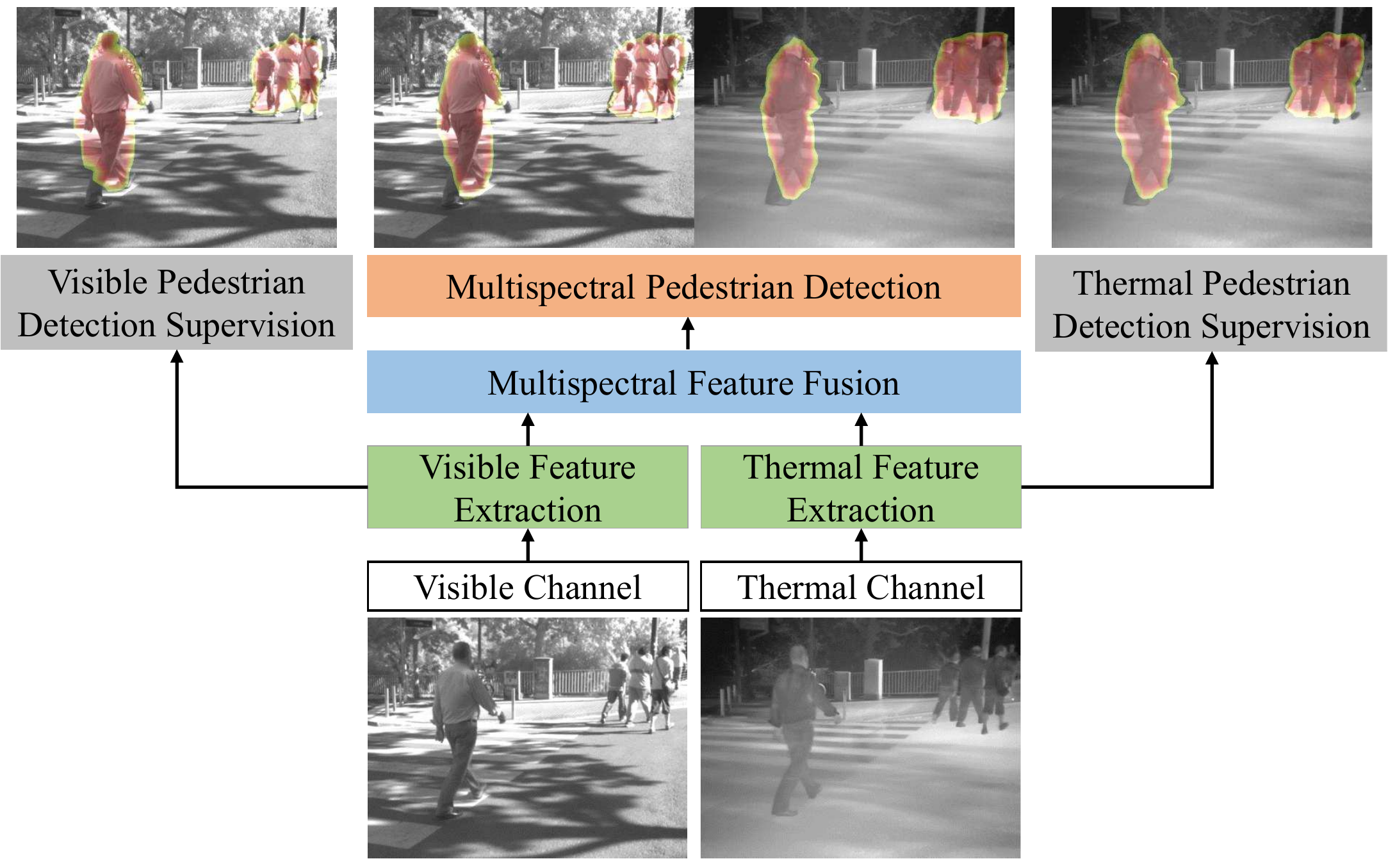}
	\end{center}
	\caption{Architecture of our proposed detector for joint training of multispectral pedestrian detections associated with visible and thermal detection supervisions. It contains four major components: feature extraction, feature fusion, pedestrian detection, and detection supervision. The feature extraction module learn features from visible and thermal channels individually; the feature fusion module integrate the visible and thermal features to generate multimodal feature maps; the pedestrian detection module learn the generated multimodal features to produce multispectral pedestrian detections; the detection supervision module learn the individual features to generate visible and thermal pedestrian detections, which provide additional feature information to facilitate the training of multispectral pedestrian detector. }
	\label{fig3}
\end{figure*}

Although significant improvements have been accomplished in the research area of pedestrian detection recently, it still remains a crucial challenge to train a reliable pedestrian detector working well in different open benchmark datasets simultaneously. In the past few years, some researchers have developed different unsupervised domain adaptation schemes in order to avoid the annotation effort. 
Wang \emph{et al.}~\cite{wang2014scene} presented a new method to achieve unsupervised domain adaptation for a scene-specific pedestrian detector. The approach explores multiple context cues (e.g., structures, locations and sizes) in the static video surveillance to select high-confident training sample on target domain. 
Liu \emph{et al.}~\cite{liu2016unsupervised} proposed an effect algorithm to iteratively select negative annotations on source domain and annotate positive labels with high score on target domain as the training samples on the task of unsupervised domain adaptation for pedestrian detection in surveillance situations. 
Wu \emph{et al.}~\cite{wu2018exploiting} designed a selective ensemble algorithm to adapt the human detector based on Haar-like features~\cite{zhang2014informed} and boosted classifier~\cite{viola2001rapid} to target domain. The selective ensemble algorithm recombined the useful components that are capable of generating human-related characteristics related to target domain. 
Cao \emph{et al.}~\cite{cao2019pedestrian} developed a novel unsupervised domain adaptation method to adapt a visible pedestrian detector on source domain to a multispectral pedestrian detector on target domain without using any annotations. An auto-annotation framework was designed to iteratively annotate pedestrian labels. 

%Our method is different from the above approach distinctly by proposing an unsupervised multimodal domain adaptation framework for multispectral pedestrian detection, by iteratively generating pseudo annotations and updating the parameters of our designed multispectral pedestrian detector on target domain.
As far as we know, it has not been solved yet to adapt a multispectral pedestrian detector on source domain to target one without manual annotations. Thus, we propose an unsupervised domain adaptation framework for multispectral pedestrian detection in this paper.

\section{Our approach}

We first design a new multispectral pedestrian detector and train it on source domain. Based on our designed detector, a novel unsupervised adaptation framework for multispectral pedestrian detection is proposed by iteratively generating training labels and updating the parameters of detector on target domain. Visible and thermal pseudo annotations are generated using our designed detector trained on source domain, and then updated by fixing the parameters of detector and minimizing the cross entropy loss without back-propagation. Training labels in visible, thermal and multispectral channels are generated using the pseudo annotations by considering the characteristics of similarity and complementarity respectively, which are existing in well-aligned visible and infrared image pairs. The parameters of detector are updated using the generated labels by minimizing our defined multi-detection loss function with back-propagation.

\begin{figure*}[!ht]
	\begin{center}
		\includegraphics[width=0.61\linewidth]{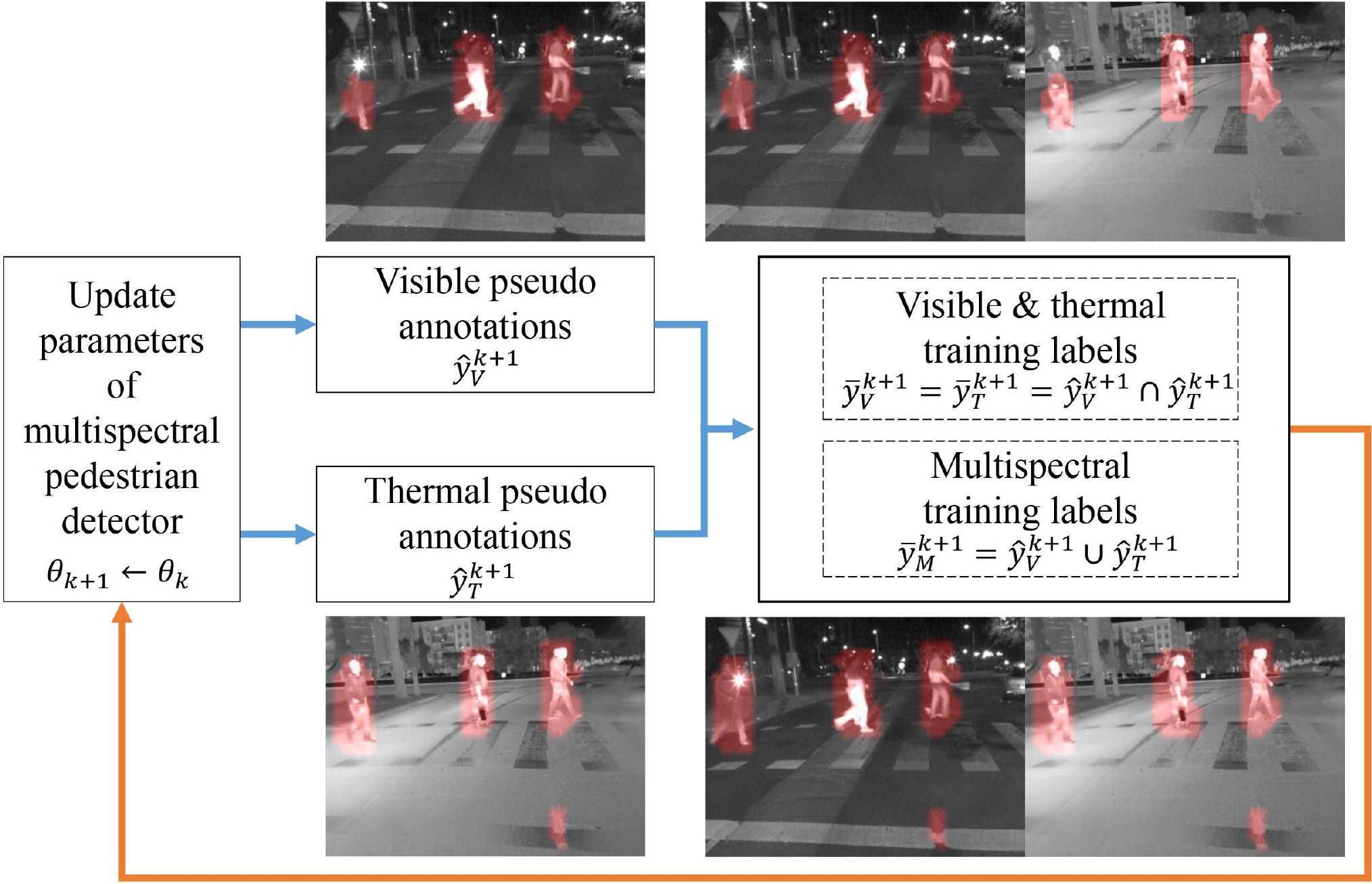}
	\end{center}
	\caption{Framework of our proposed unsupervised multimodal domain adaptation model. At each iteration step, visible and thermal pseudo annotations are updated by fixing the parameters of detector and minimizing the cross entropy loss without back-propagation. Training labels in visible, thermal and multispectral channels are generated using the pseudo annotations by considering the characteristics of similarity and complementarity respectively, which are existing in well-aligned visible and infrared image pairs. The parameters of detector are updated using the generated labels by minimizing our defined multi-detection loss function with back-propagation.}
	\label{fig4}
\end{figure*}

\subsection{Multispectral pedestrian detector}

Inspired by the multi-task framework for joint training of multispectral pedestrian detection and semantic segmentation~\cite{guan2018exploiting}, we combine the visible and thermal pedestrian detection supervision module with the box-level segmentation supervised deep neural networks~\cite{cao2019pedestrian} to build multispectral pedestrian detector, as illustrated in Fig.~\ref{fig3}. During the training procedure, the bounding box annotations are used to generate box-level segmentation mask as training labels for the detector on multispectral, visible and thermal channels simultaneously. 

Let $\{x,\overline{y}\}$ denote the training images $x$ with box-level segmentation masks $\overline{y}=\{\overline{y}_{i},i=1,...,I\}$ ($I$ pixels), where $\overline{y}_{i}=1$ represents the foreground pixel and $\overline{y}_{i}=0$ denotes the background one. At each iteration step, the parameters $\theta$ is updated by minimizing a multi-detection loss function, which is defined as:
\begin{equation}
\begin{aligned}
\theta^{k+1} = \mathop{\arg\min}_{\theta^{k}}(\mathcal{L}_{c}(y_{M},\overline{y},I\mathop{|}x;\theta^{k})\\ + \mathcal{L}_{c}(y_{V},\overline{y},I\mathop{|}x;\theta^{k}) + \mathcal{L}_{c}(y_{T},\overline{y},I\mathop{|}x;\theta^{k})) ;
\label{eq1}
\end{aligned}
\end{equation}
where $y_{M}$, $y_{V}$ and $y_{T}$ represent the prediction of pedestrian regions on multispectral, visible and thermal channels; $I$ represents the set of training pixels on the box-level segmentation masks; and $\mathcal{L}_{c}(y,\overline{y},I\mathop{|}x;\theta)$ is the cross entropy loss for classification which is defined as:
\begin{equation}
\begin{aligned}
\mathcal{L}_{c}(y,\overline{y},I\mathop{|}x;\theta) = -\sum_{i\in{I}}(\overline{y}_{i}\text{log}(y_{i}) + (1-\overline{y}_{i})\text{log}(1-y_{i})) ,
\label{eq2}
\end{aligned}
\end{equation}
where ${y}_{i}\in[0,1]$ represents the confident score which predicts the probability of the corresponding pixel belonging to pedestrian regions, $\overline{y}_{i}=1$ presents the foreground pixel and $\overline{y}_{i}=0$ denotes the background pixel.

The optimal parameters of detector $\theta^{*}$ can be obtained after iteratively updating the parameters $\theta$. During the testing phase, the output of our designed detector is multispectral pedestrian detections, which are form of full-size heat map predictions. 

\subsection{Unsupervised multimodal domain adaptation}

Based on our designed multispectral pedestrian detector, we propose an unsupervised domain adaptation framework for multispectral pedestrian detection by iteratively generating training labels and updating the parameters of detector on target domain, as illustrated in Fig.~\ref{fig4}. Firstly, the visible and thermal pseudo annotations $\{\hat{y}_{V}^{0},\hat{y}_{T}^{0}\}$ are initialized using the visible and thermal pedestrian detection supervision module of our designed detector, which has been trained on source domain. 
%At each iteration step $k$, the visible and thermal pseudo annotations $\{\hat{y}_{V}^{k+1},\hat{y}_{T}^{k+1}\}$ are generated by fixing the parameters of detector $\theta^{k}$ and minimizing the cross entropy loss function $\mathcal{L}_{c}$ which is defined in Eq.~\ref{eq2} without back-propagation.
At each iteration step $k$, the most confident pseudo labels can be selected by fixing the parameters of detector $\theta^{k}$ and minimizing the cross entropy loss function $\mathcal{L}_{c}$. The visible and thermal pseudo annotations $\{\hat{y}_{V}^{k+1},\hat{y}_{T}^{k+1}\}$ are generated by adding the most confident pseudo labels into the existing set $\{\hat{y}_{V}^{k},\hat{y}_{T}^{k}\}$.

Thus, the optimization of visible pseudo annotations $\hat{y}_{V}^{k+1}\in\{0,1\}$ is defined as:
\begin{equation}
\hat{y}_{V}^{k+1} = \mathop{\arg\min}_{\hat{y}_{V}}(\mathcal{L}_{c}(y_{V},\hat{y}_{V},I\mathop{|}x;\theta^{k})) \cup \hat{y}_{V}^{k} ,
\label{eq3}
\end{equation}
and the thermal pseudo annotations $\hat{y}_{T}^{k+1}\in\{0,1\}$ is optimized as:
\begin{equation}
\hat{y}_{T}^{k+1} = \mathop{\arg\min}_{\hat{y}_{T}}(\mathcal{L}_{c}(y_{T},\hat{y}_{T},I\mathop{|}x;\theta^{k})) \cup \hat{y}_{T}^{k} .
\label{eq4}
\end{equation}

The optimized visible and thermal pseudo annotations are used to generate training labels on visible, thermal and multispectral channels, by considering the characteristics of similarity and complementarity between well-aligned visible and thermal image pairs. 
The similarity means that there exists obvious human-related features on both visible and thermal channels, which can be used as a cue to train the visible and thermal pedestrian detection supervision module simultaneously. We consider the intersection of visible and thermal pseudo annotations as the regions that exist obvious human-related features on both channels. Thus, the visible and thermal training labels are generated as:                                                                                                                                                                                                                                                                                                                                                                                                                                                                                                         
\begin{equation}
\overline{y}_{V}^{k+1} = \overline{y}_{T}^{k+1} = \hat{y}_{V}^{k+1} \cap \hat{y}_{T}^{k+1} .
\label{eq5}
\end{equation}

Considering that the pseudo annotations should not be considered as negative training labels, we define the set of visible training pixels as:
\begin{equation}
I_{V} = I-\hat{y}_{V}^{k+1}+\overline{y}_{V}^{k+1},
\label{eq6}
\end{equation}
and the set of thermal training pixels as:
\begin{equation}
I_{T} = I-\hat{y}_{V}^{k+1}+\overline{y}_{T}^{k+1}.
\label{eq7}
\end{equation}

The complementarity means that visible and thermal data can provide complementary information about objects of interest to improve the detection accuracy. We consider the union of visible and thermal pseudo annotations as the complementary information to update the parameters of multispectral pedestrian detector. Thus, the multispectral training labels are generated as:
\begin{equation}
\overline{y}_{M}^{k+1} = \hat{y}_{V}^{k+1} \cup \hat{y}_{T}^{k+1} .
\label{eq8}
\end{equation}

The parameters of detector are updated using the generated training labels by minimizing a multi-detection loss function with back-propagation, which is defined as:
\begin{equation}
\begin{aligned}
\theta^{k+1} = \mathop{\arg\min}_{\theta^{k}}(\mathcal{L}_{c}(y_{M},\overline{y}_{M}^{k+1},I\mathop{|}x;\theta^{k})\\ + \mathcal{L}_{c}(y_{V},\overline{y}_{V}^{k+1},I_{V}\mathop{|}x;\theta^{k})\\ + \mathcal{L}_{c}(y_{T},\overline{y}_{T}^{k+1},I_{T}\mathop{|}x;\theta^{k})) .
\label{eq9}
\end{aligned}
\end{equation}

The optimal parameters of detector $\theta^{*}$ can be obtained after iteratively updating the visible and thermal pseudo annotations $\{\hat{y}_{V},\hat{y}_{T}\}$ and the parameters $\theta$.

\section{Experiments}

\subsection{Datasets}
In order to conduct our experiments on multimodal domain adaptation, we utilize the KAIST~\cite{hwang2015multispectral} and CVC-14~\cite{gonzalez2016pedestrian} multispectral pedestrian benchmarks as the source and target domain datasets respectively. 

The KAIST training dataset contains 50172 well-aligned color visible and thermal infrared sequential image pairs with 13853 dense pedestrian annotations. The images on KAIST dataset were captured in various traffic environments with a resolution of $640 \times 512$. Considering that the visible images from CVC-14 dataset are gray scale without much color information, we transfer the visible images in KAIST dataset from RGB to gray level in order to decrease domain differences between these two datasets. According to the current state-of-the-art multispectral pedestrian detector designed by Cao \emph{et al.}~\cite{cao2019box}, the training images are downscaled to the resolution of $320 \times 256$ through bilinear interpolation and the bounding box annotations are transfered to box-level segmentation masks to train the detectors on source domain.

The CVC-14 training dataset consists of 7085 aligned gray visible and thermal infrared sequential image pairs with 8105 dense pedestrian annotations. It should be noted that the manual annotations on the CVC-14 training dataset are abandoned in our designed unsupervised multimodal domain adaptation method. The CVC-14 testing dataset contains 1433 aligned image pairs in which 706 pairs were captured during daytime and others in nighttime. All the images on CVC-14 dataset were captured in city traffic environments with a resolution of $640 \times 480$. For a fair comparison with the current state-of-the-art multispectral pedestrian detector designed by Cao \emph{et al.}~\cite{cao2019box}, we downscale the images to the resolution of $320 \times 240$ through bilinear interpolation during training and testing phase on target domain. The annotations of CVC-14 test set under the reasonable setting (pedestrians larger than 50 pixels~\cite{gonzalez2016pedestrian}) are used to evaluate detection performance.

\subsection{Implementation Details}
All the detectors are trained and tested using the Caffe~\cite{jia2014caffe} deep learning framework with the image-centric strategy to generate mini-batches. We set the batch size to one. For the supervised multimodal domain adaptation, the box-level segmentation masks are generated as training labels using the bounding box annotations, following the multispectral pedestrian detection method designed by Cao \emph{et al.}~\cite{cao2019box}. Each stream in multispectral deep neural networks is initialized using the parameters in VGG-16~\cite{simonyan2014very}  pre-trained on the ImageNet dataset~\cite{russakovsky2015imagenet} and the other convolutional layers are initialized according to Xavier initialization following~\cite{glorot2010understanding}. The multispectral pedestrian detector on source domain is trained with stochastic gradient descent (SGD) algorithm~\cite{zinkevich2010parallelized} for the first 2 epochs with learning rate (LR) 0.001 and 1 more epoch with LR 0.0001 following~\cite{cao2019box}. The multispectral pedestrian detector on target domain is fine-tuned with SGD algorithm for 4 epochs with a low LR of 0.00005. In order to avoid gradient exploding, we utilize the adjustable gradient clipping method~\cite{pascanu2013difficulty} in the training procedure to suppress exploding gradients.

\subsection{Evaluation Metric}
The final output of our approach contains human regions and background regions classified by the confident scores according to our framework’s prediction in a heat map style, following the method presented by Cao \emph{et al.}~\cite{cao2019box}. Considering the difference of results depicted in heat map and traditional bounding box style, to compare impartially, we turn detection results in bounding box into heat map representation depending on the prediction scores. As it’s utilized diffusely, we also adopt the average precision (AP)~\cite{deerwester1990indexing, cao2019box} in pixel-level as metric to quantify the comparison result between our method and others. More specifically, the average precision (AP) refers to 4 concepts including true positive (TP), true negative (TN), false positive (FP), false negative (FN). Provided with the human-target labels, some bounding box in this case, we call the pixels in bounding box containing human-target foreground pixels, while those surrounding ones is treated as background pixels. After we get the final heat map, true positive (TP) counts those pixels belonging to human-targets inferred correctly, true negative (TN) counts those pixels that is not belonging to human-targets but gets inferred, false positive (FP) counts those background pixels inferred incorrectly, false negative (FN) counts those background pixels inferred correctly. The precision is the ratio TP / (TP + FP), while the recall is the ratio TP / (TP + FN). The AP depicts the shape of the precision/recall curve, and is defined as the mean precision at each recalls by varying the threshold on detection scores. 

\subsection{Evaluation of UMDA}

In order to verify the effectiveness of our proposed approach, we evaluate the detection performance of our proposed unsupervised multimodal domain adaptation (UMDA) model with the detection model without multimodal domain adaptation (WMDA). The quantitative performance (pixel-wise AP~\cite{cao2019box}) of UMDA and WMDA are compared in Tab.~\ref{tab1}. It is observed that our proposed unsupervised multimodal domain adaptation method achieves multispectral pedestrian detection performance significantly higher than the approach without domain adaptation, pixel-level AP~\cite{cao2019box} of UMDA is 31.37\% higher than the results of WMDA.

\begin{table}[!h]
	\centering
	\caption{ Comparing the quantitative performance (pixel-wise AP~\cite{cao2019box}) of UMDA and WMDA.}
	\begin{tabular}{cccc}
		\hline
		Model &All-day	&Daytime	&Nighttime	\\
		\hline
		{WMDA}	&0.4886	&0.3986	&0.5584	\\ 
		\textbf{UMDA} & \textbf{0.8023}	&\textbf{0.7688}	&\textbf{0.8503} \\  
		\hline		
	\end{tabular}
	\label{tab1}
\end{table}

\begin{figure}[t]
	\centering	{ WMDA  \qquad \qquad \qquad \qquad \quad UMDA }
	\begin{minipage}{0.99\linewidth}
		\begin{minipage}{0.24\linewidth}
			\includegraphics[width=1\linewidth,trim=160 120 0 0,clip]{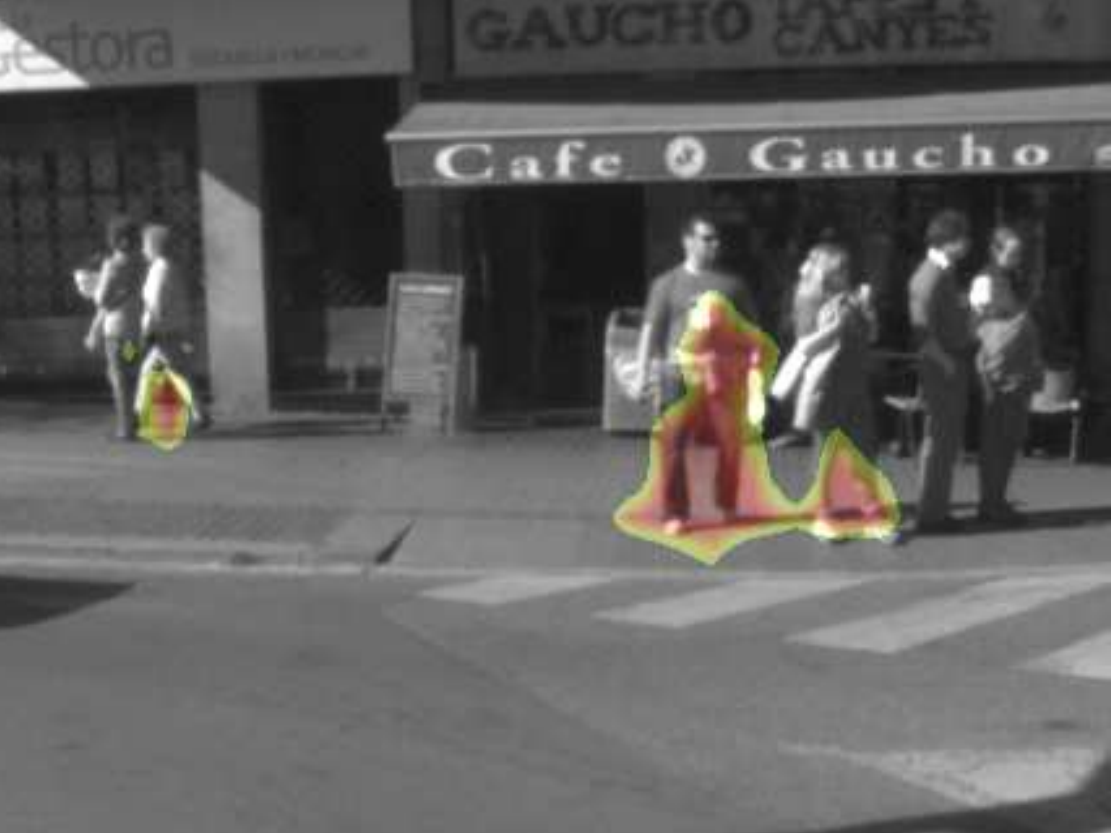}
		\end{minipage}
		\begin{minipage}{0.24\linewidth}
			\includegraphics[width=1\linewidth,trim=160 120 0 0,clip]{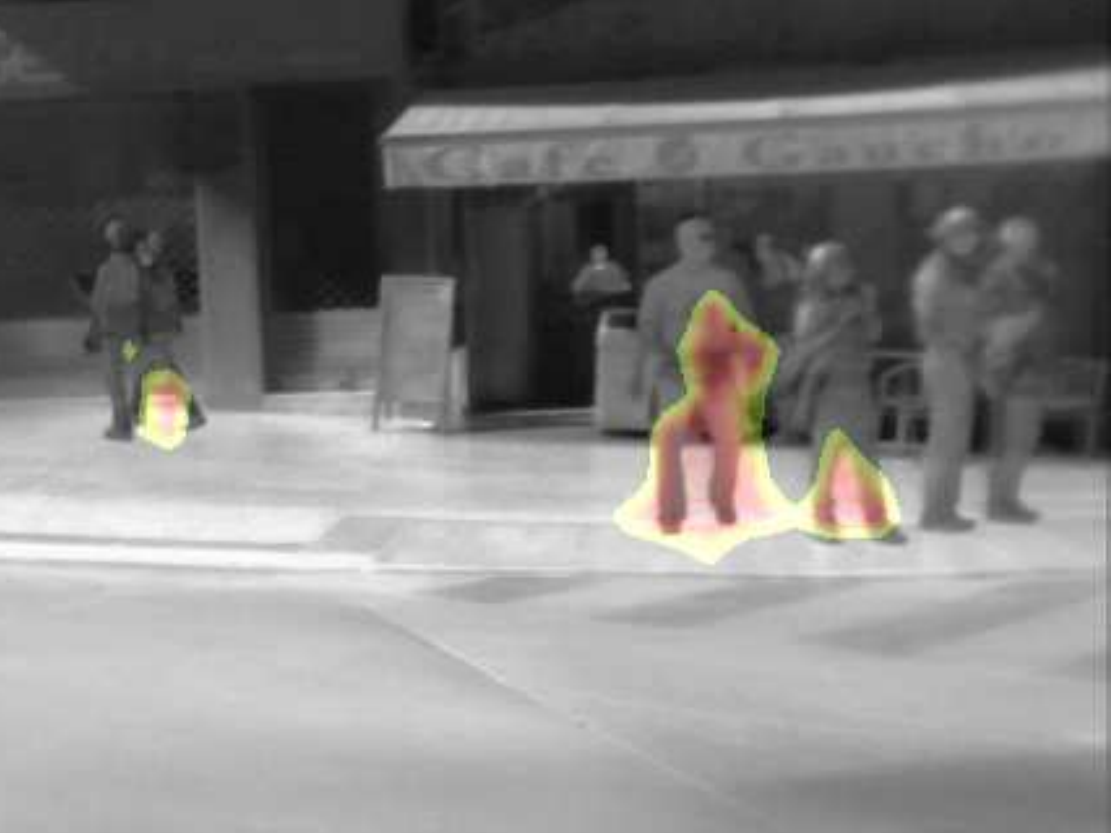}
		\end{minipage}
		\hspace{0.5mm}
		\begin{minipage}{0.24\linewidth}
			\includegraphics[width=1\linewidth,trim=160 120 0 0,clip]{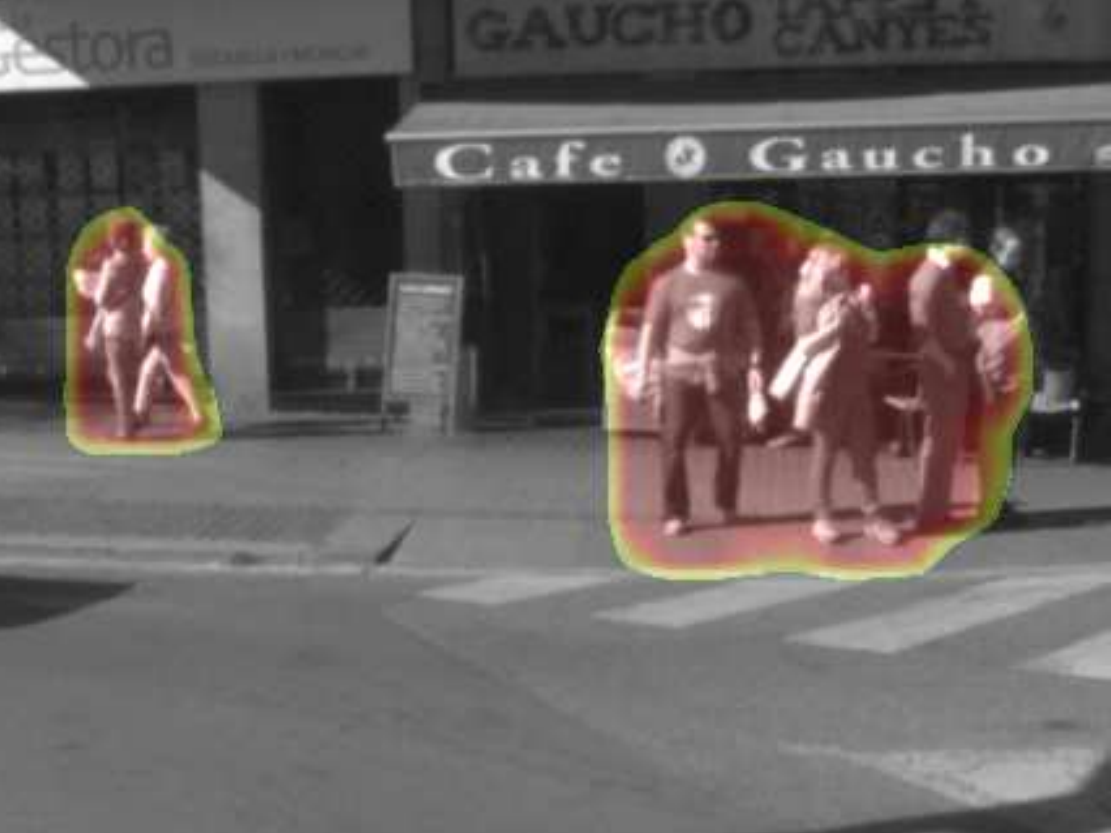}
		\end{minipage}
		\begin{minipage}{0.24\linewidth}
			\includegraphics[width=1\linewidth,trim=160 120 0 0,clip]{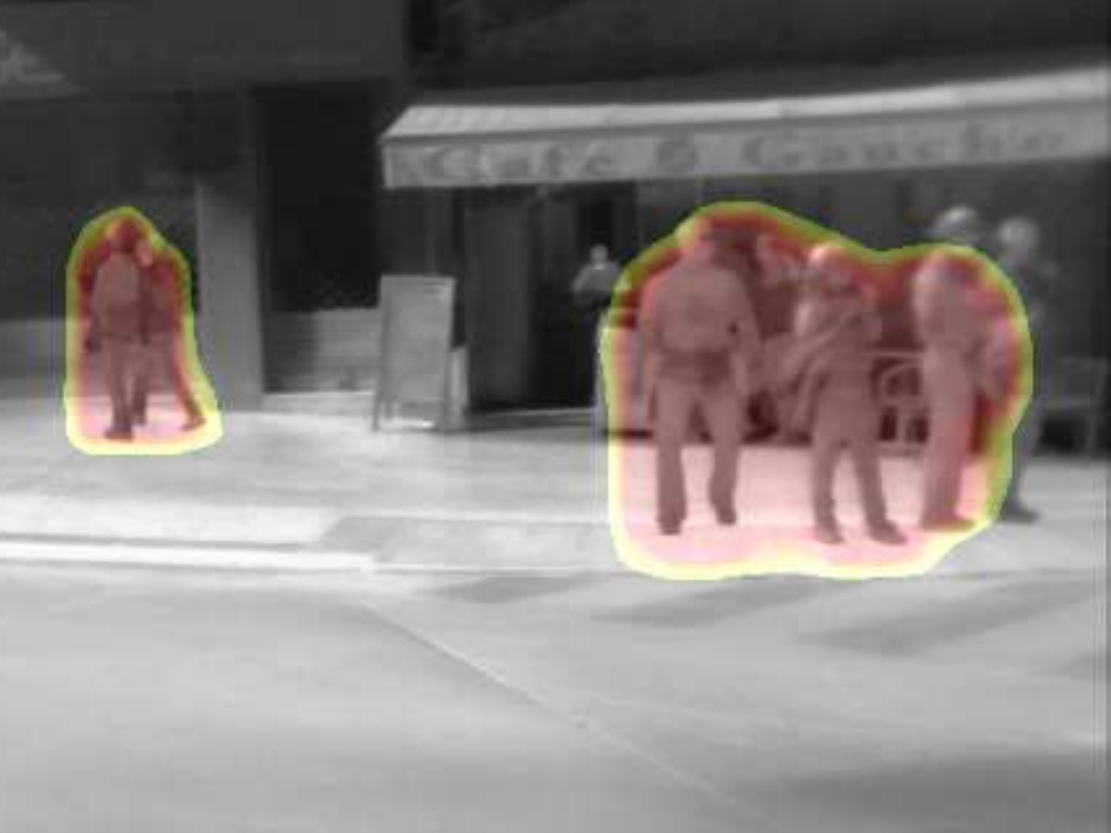}
		\end{minipage}
		\vspace{0.5mm}
	\end{minipage}
	\begin{minipage}{0.99\linewidth}
		\begin{minipage}{0.24\linewidth}
			\includegraphics[width=1\linewidth,trim=0 120 160 0,clip]{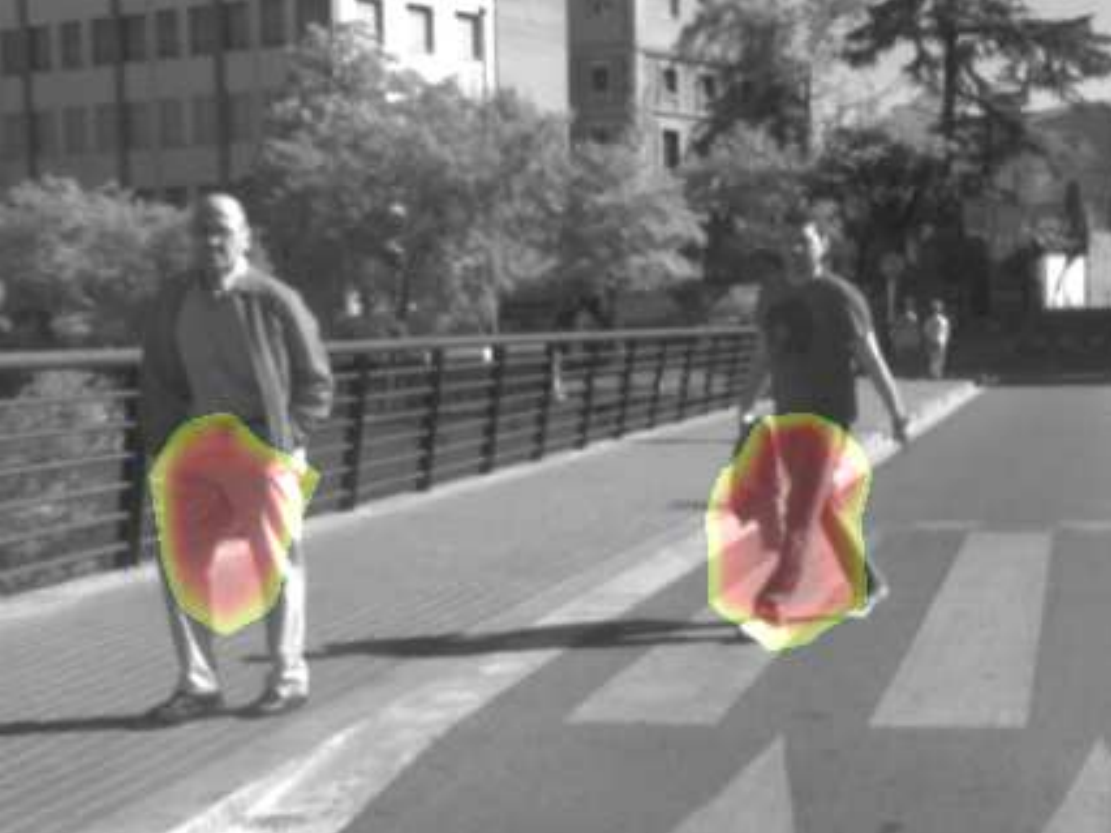}
		\end{minipage}
		\begin{minipage}{0.24\linewidth}
			\includegraphics[width=1\linewidth,trim=0 120 160 0,clip]{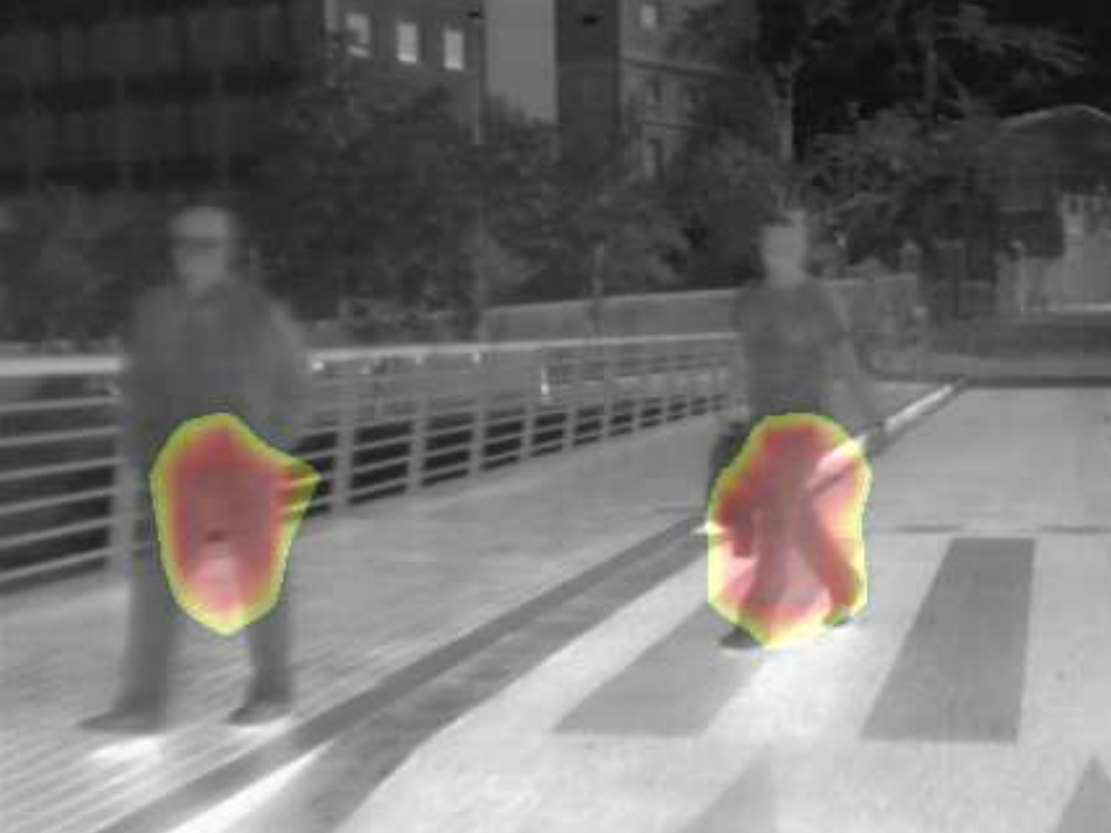}
		\end{minipage}
		\hspace{0.5mm}
		\begin{minipage}{0.24\linewidth}
			\includegraphics[width=1\linewidth,trim=0 120 160 0,clip]{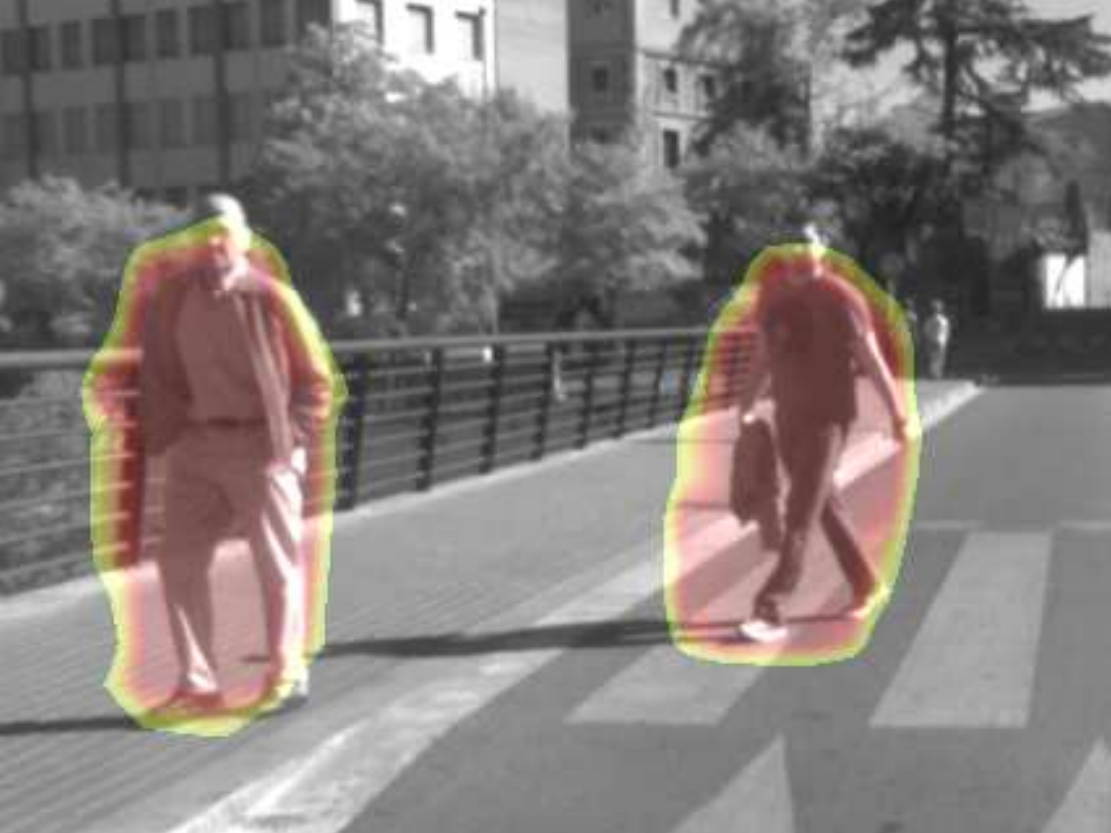}
		\end{minipage}
		\begin{minipage}{0.24\linewidth}
			\includegraphics[width=1\linewidth,trim=0 120 160 0,clip]{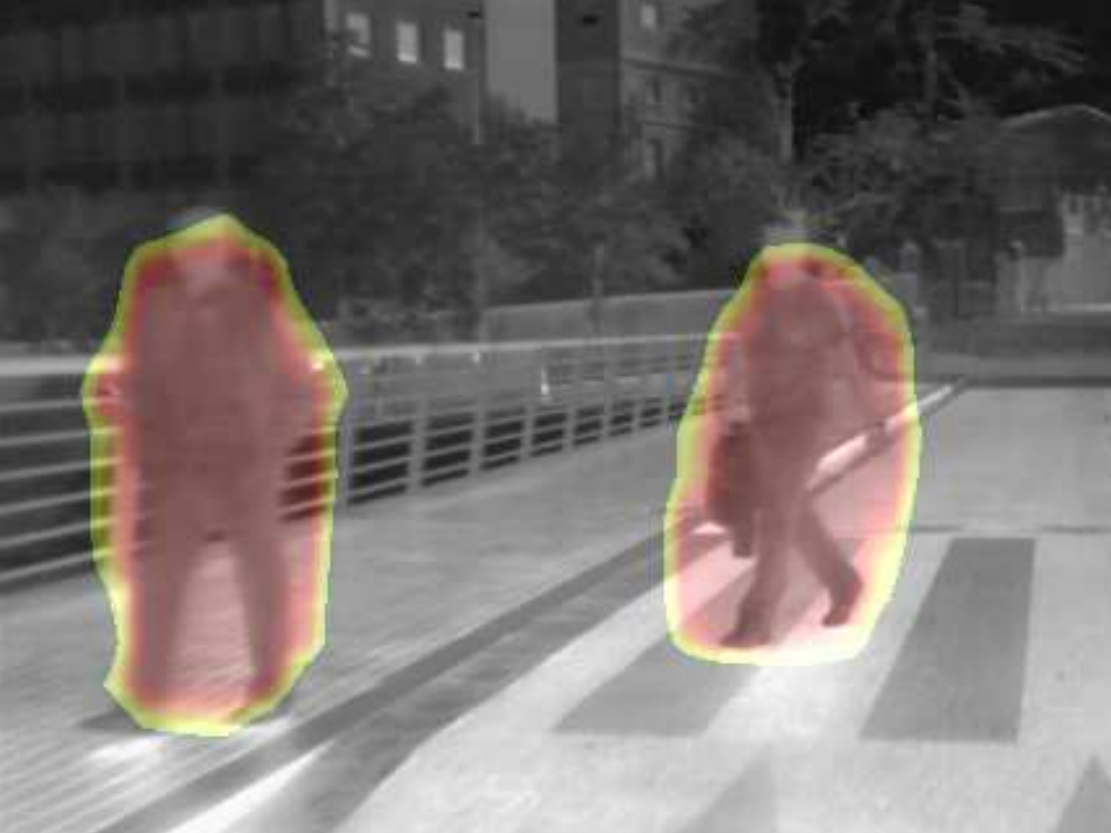}
		\end{minipage}
		\vspace{0.5mm}
	\end{minipage}
	\begin{minipage}{0.99\linewidth}
		\begin{minipage}{0.24\linewidth}
			\includegraphics[width=1\linewidth,trim=80 120 80 0,clip]{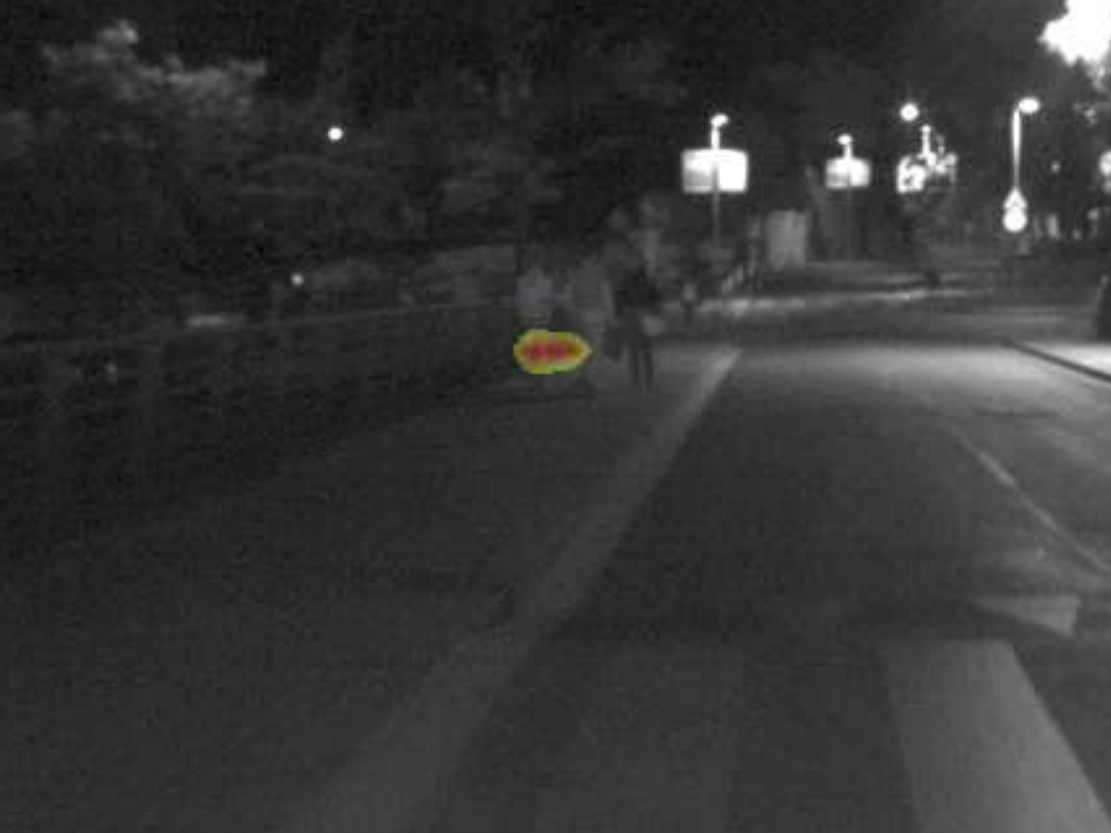}
		\end{minipage}
		\begin{minipage}{0.24\linewidth}
			\includegraphics[width=1\linewidth,trim=80 120 80 0,clip]{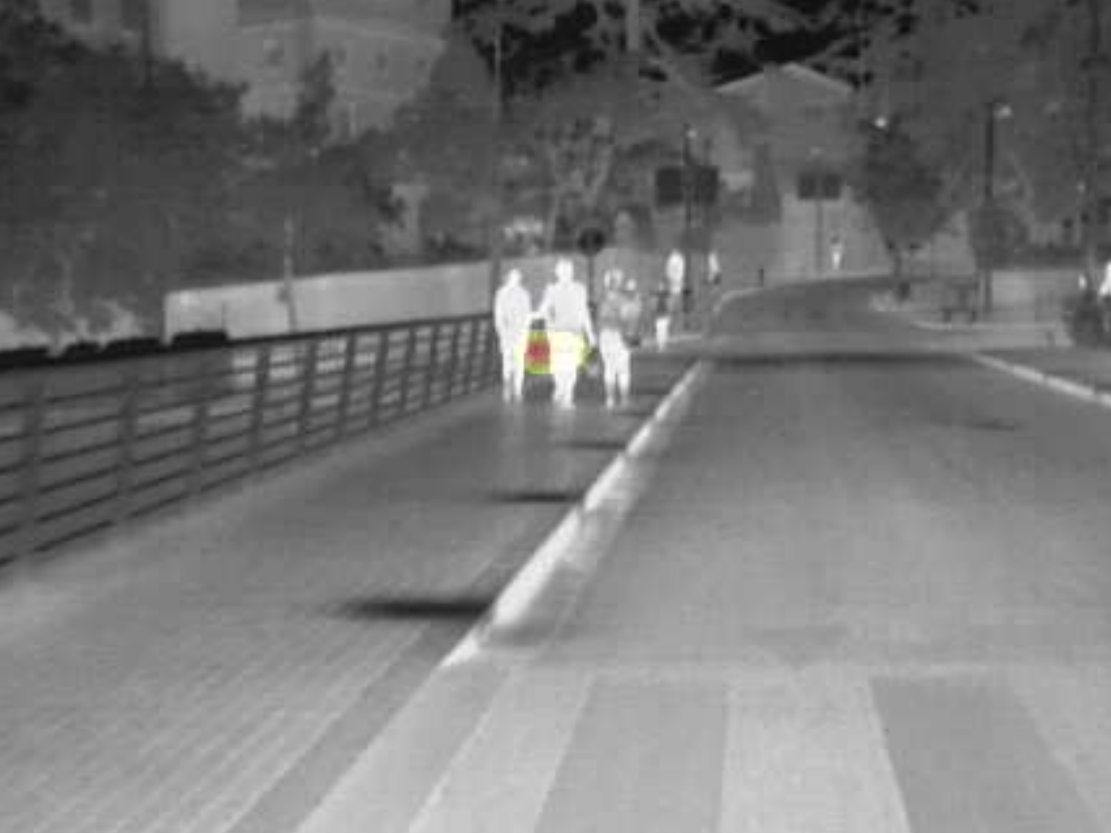}
		\end{minipage}
		\hspace{0.5mm}
		\begin{minipage}{0.24\linewidth}
			\includegraphics[width=1\linewidth,trim=80 120 80 0,clip]{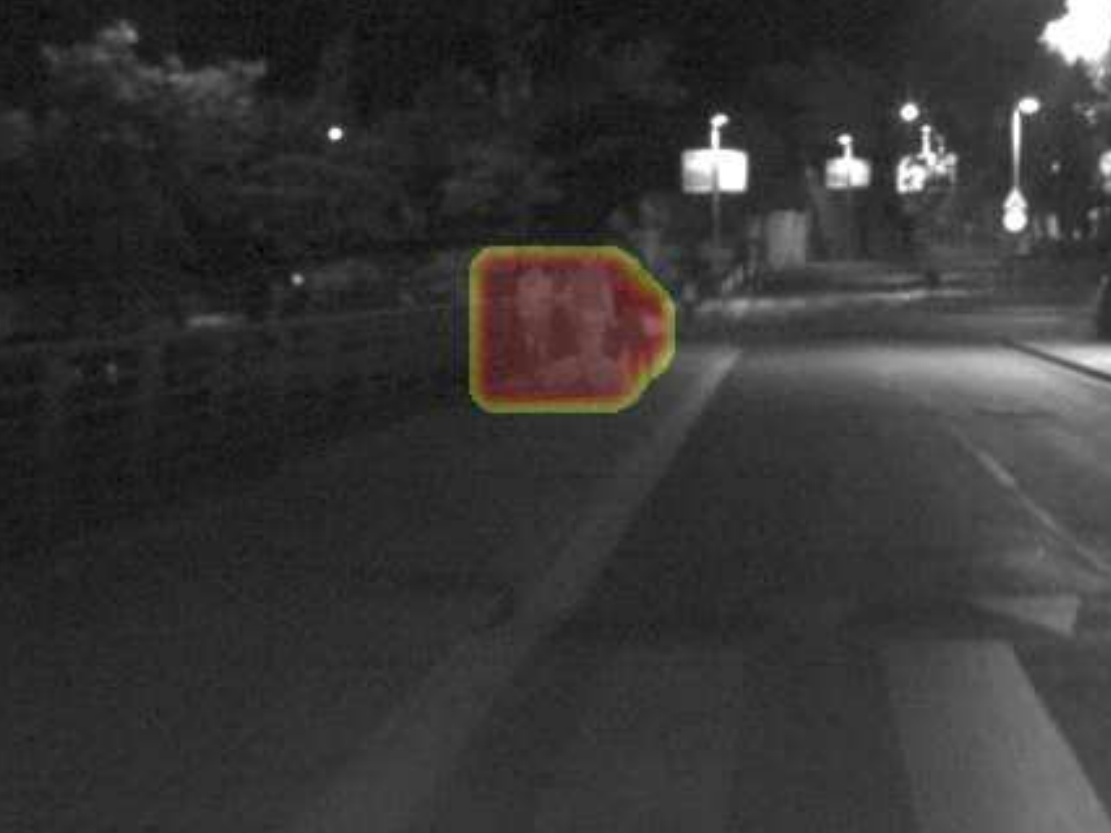}
		\end{minipage}
		\begin{minipage}{0.24\linewidth}
			\includegraphics[width=1\linewidth,trim=80 120 80 0,clip]{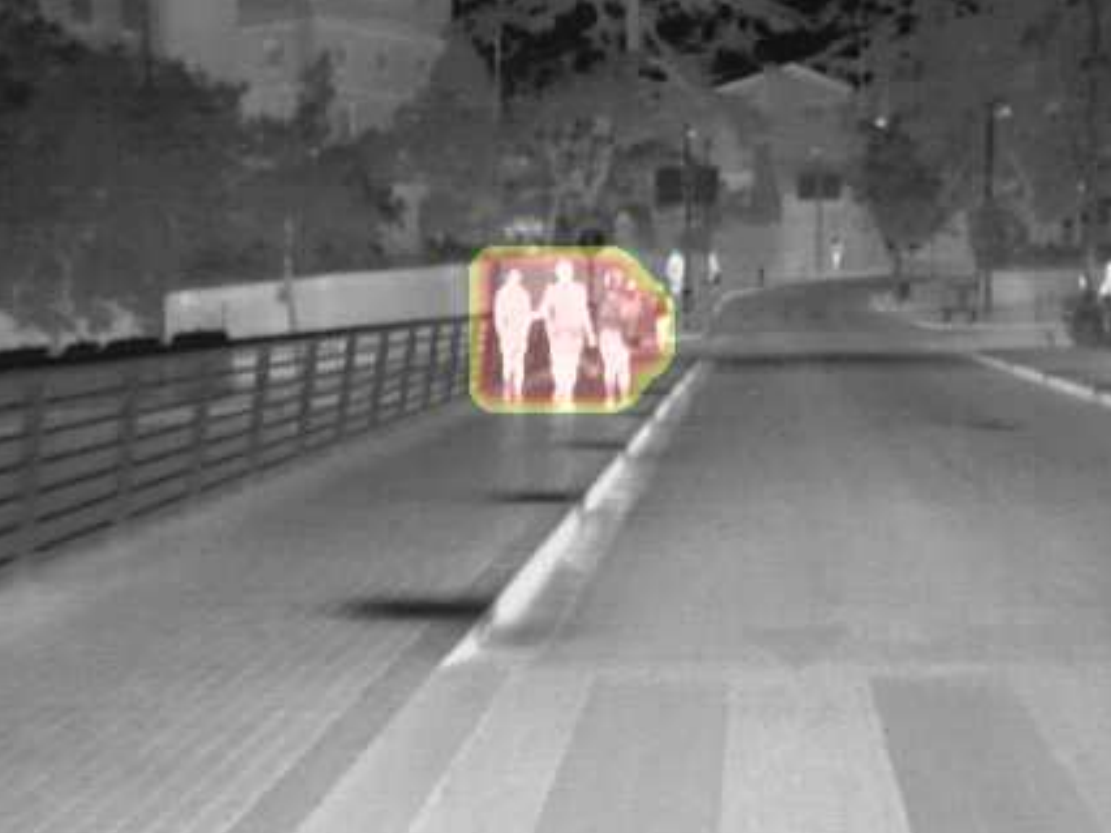}
		\end{minipage}
		\vspace{0.5mm}
	\end{minipage}
	\begin{minipage}{0.99\linewidth}
		\begin{minipage}{0.24\linewidth}
			\includegraphics[width=1\linewidth,trim=160 120 0 0,clip]{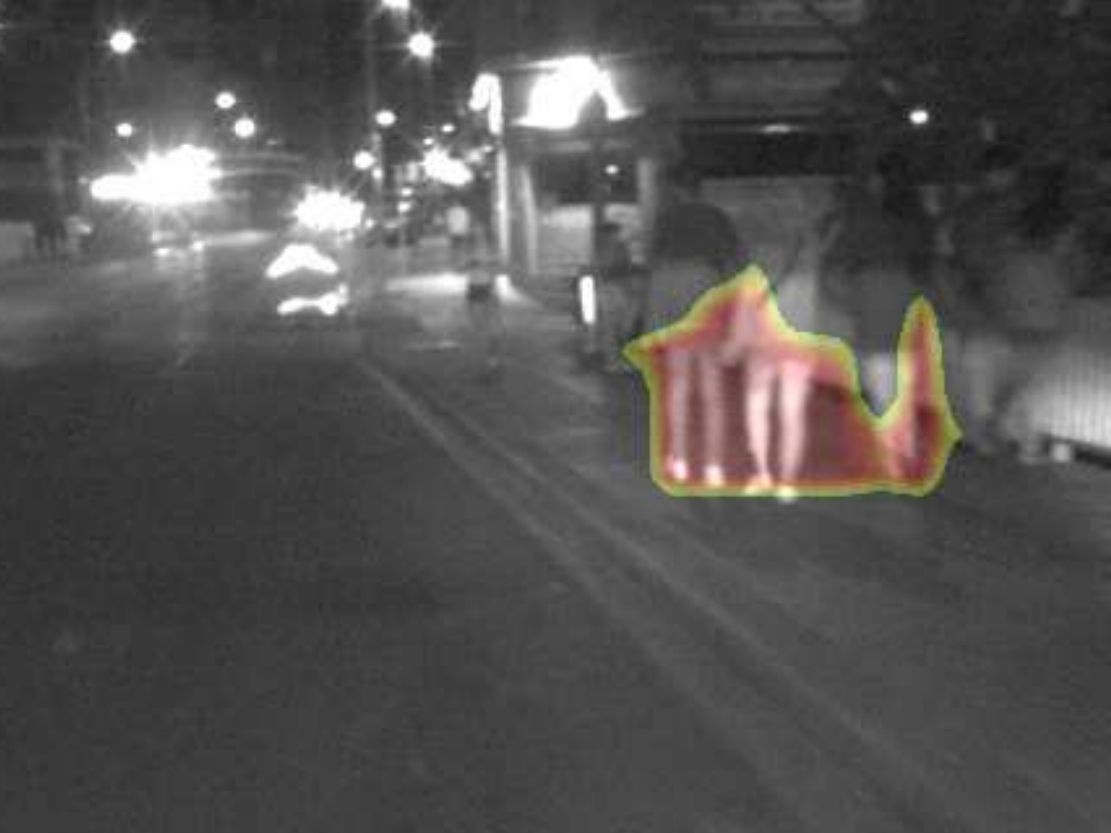}
		\end{minipage}
		\begin{minipage}{0.24\linewidth}
			\includegraphics[width=1\linewidth,trim=160 120 0 0,clip]{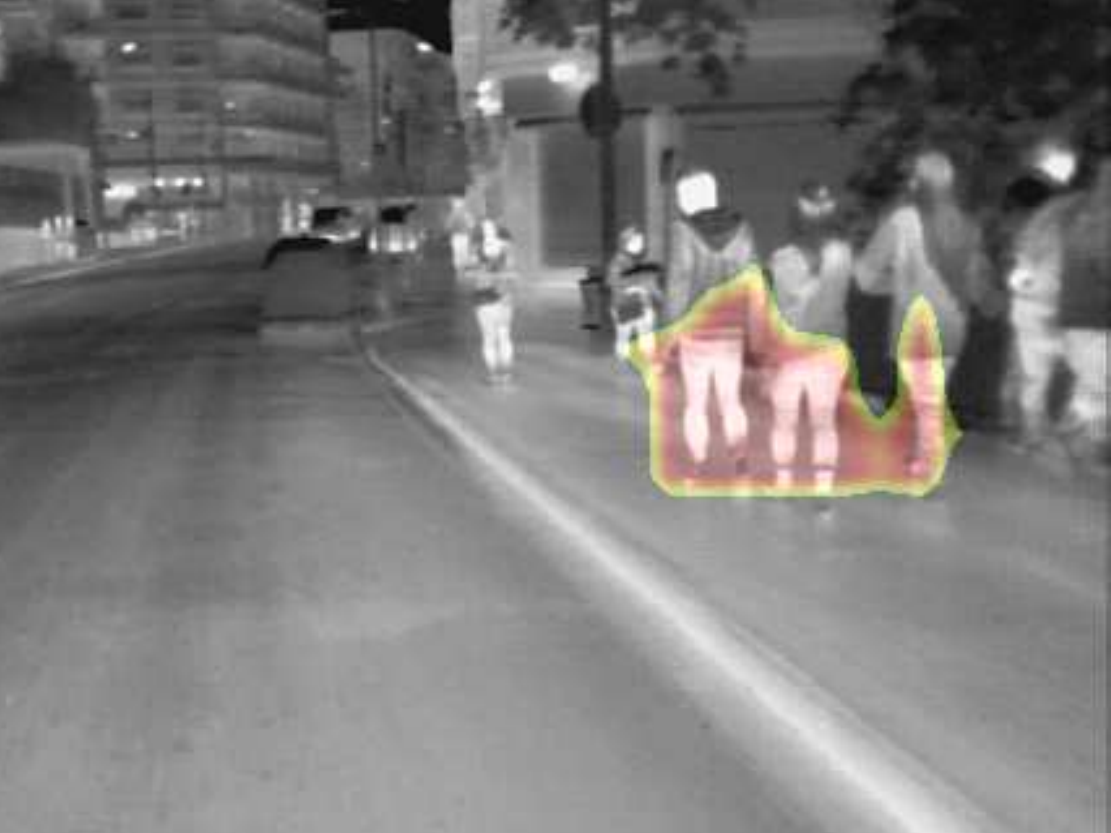}
		\end{minipage}
		\hspace{0.5mm}
		\begin{minipage}{0.24\linewidth}
			\includegraphics[width=1\linewidth,trim=160 120 0 0,clip]{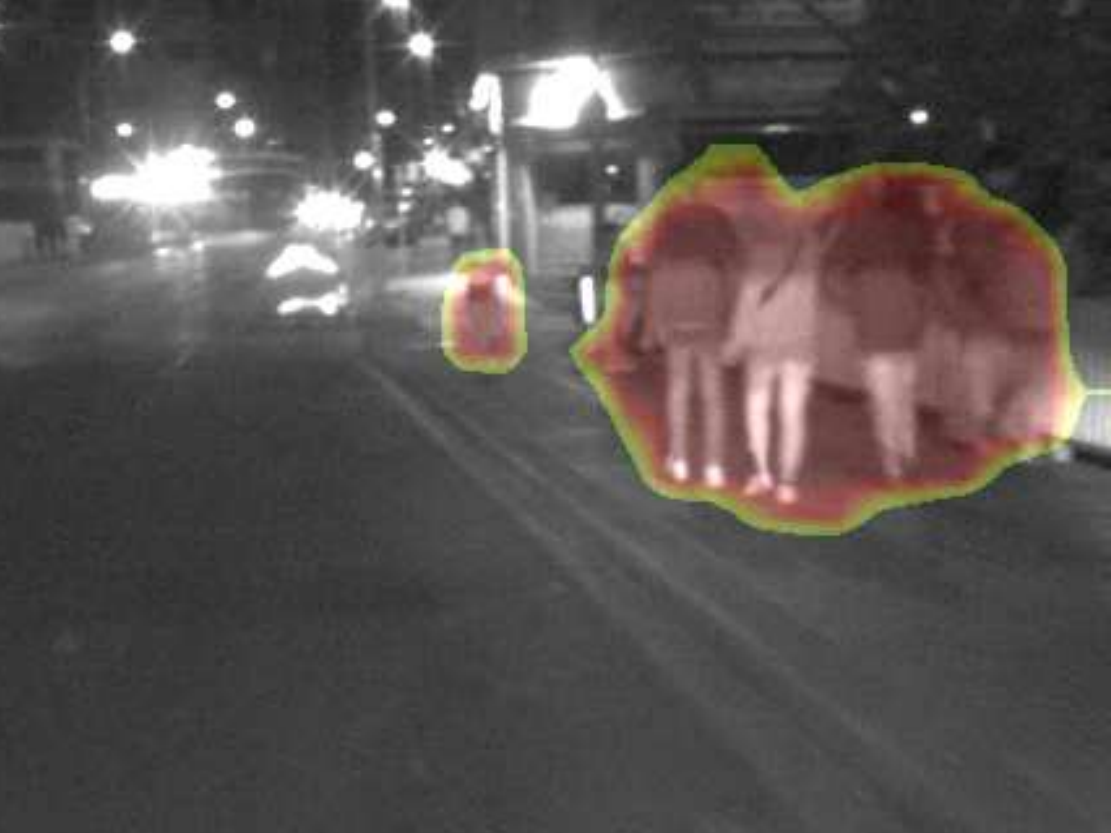}
		\end{minipage}
		\begin{minipage}{0.24\linewidth}
			\includegraphics[width=1\linewidth,trim=160 120 0 0,clip]{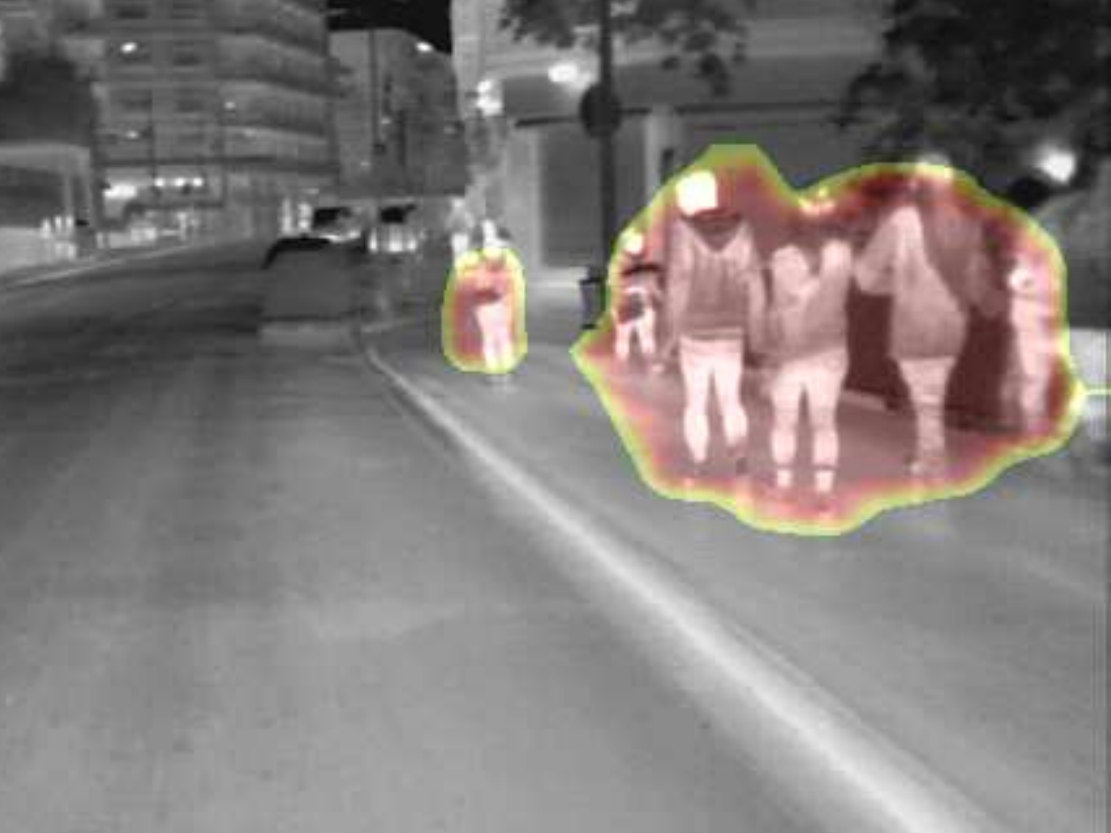}
		\end{minipage}
		\vspace{0.5mm}
	\end{minipage}
	\centering{(a)}
	\begin{minipage}{0.99\linewidth}
		\begin{minipage}{0.24\linewidth}
			\includegraphics[width=1\linewidth,trim=160 120 0 0,clip]{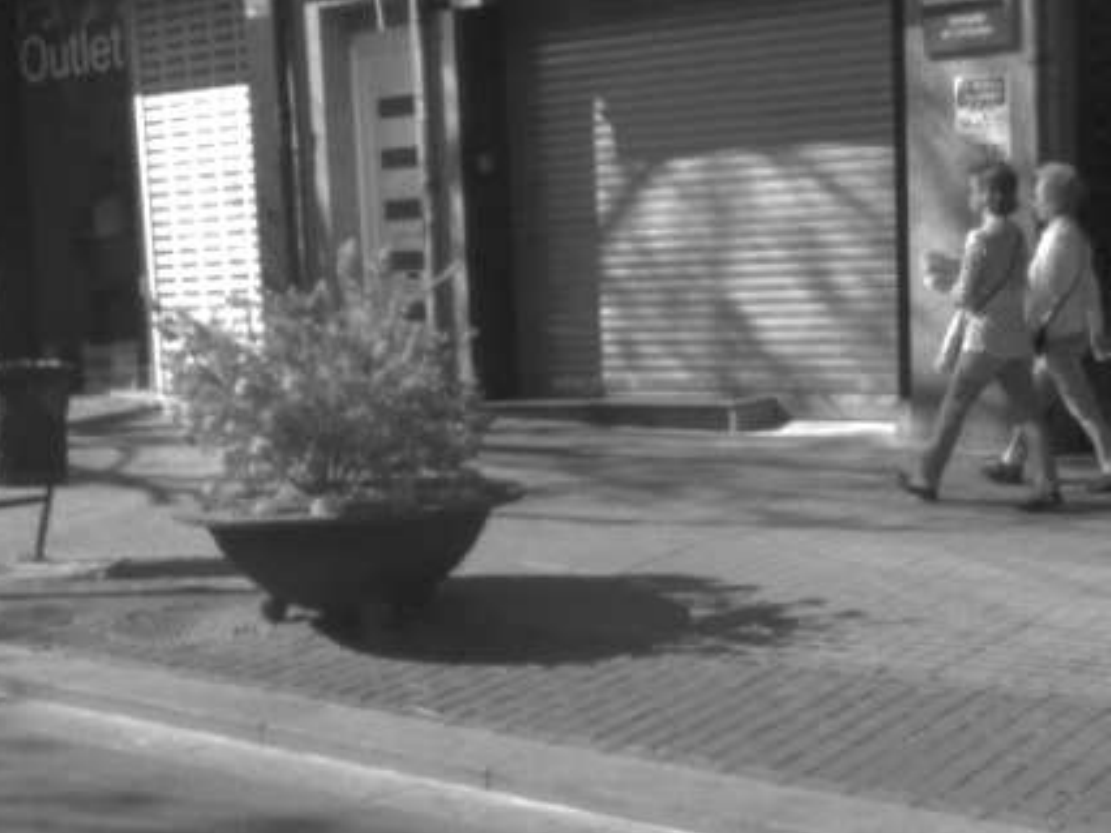}
		\end{minipage}
		\begin{minipage}{0.24\linewidth}
			\includegraphics[width=1\linewidth,trim=160 120 0 0,clip]{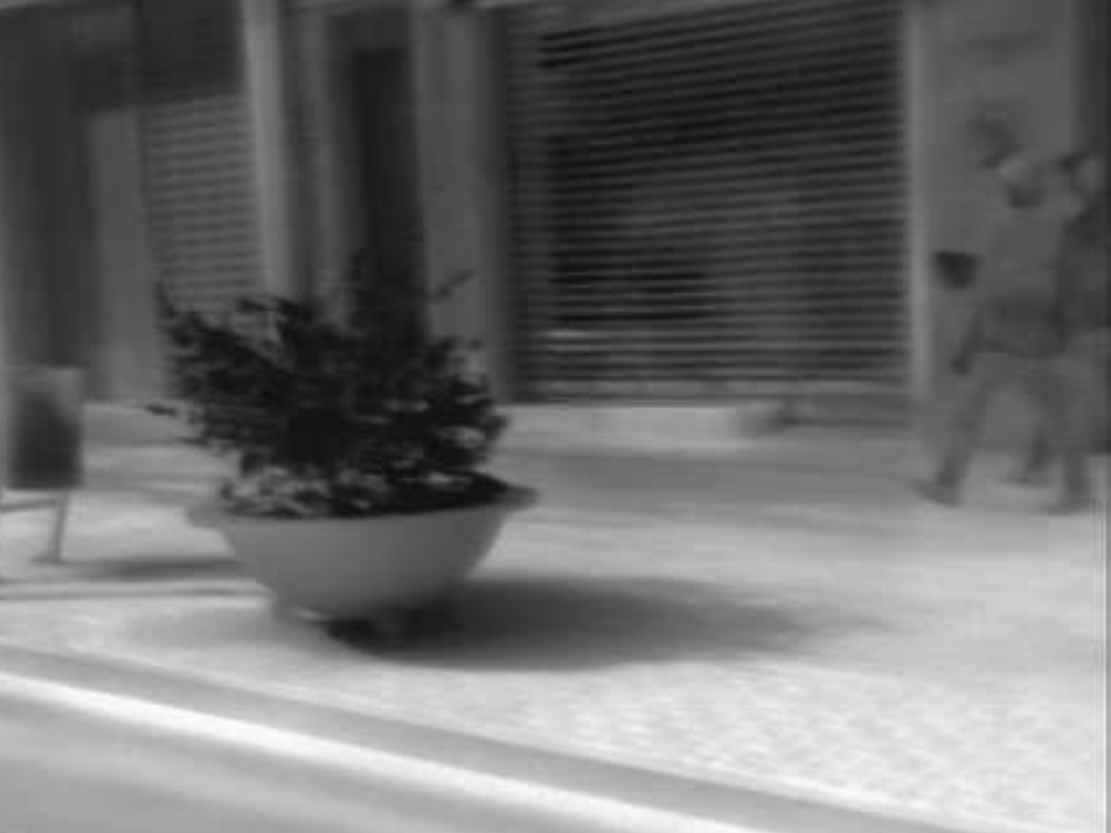}
		\end{minipage}
		\hspace{0.5mm}
		\begin{minipage}{0.24\linewidth}
			\includegraphics[width=1\linewidth,trim=160 120 0 0,clip]{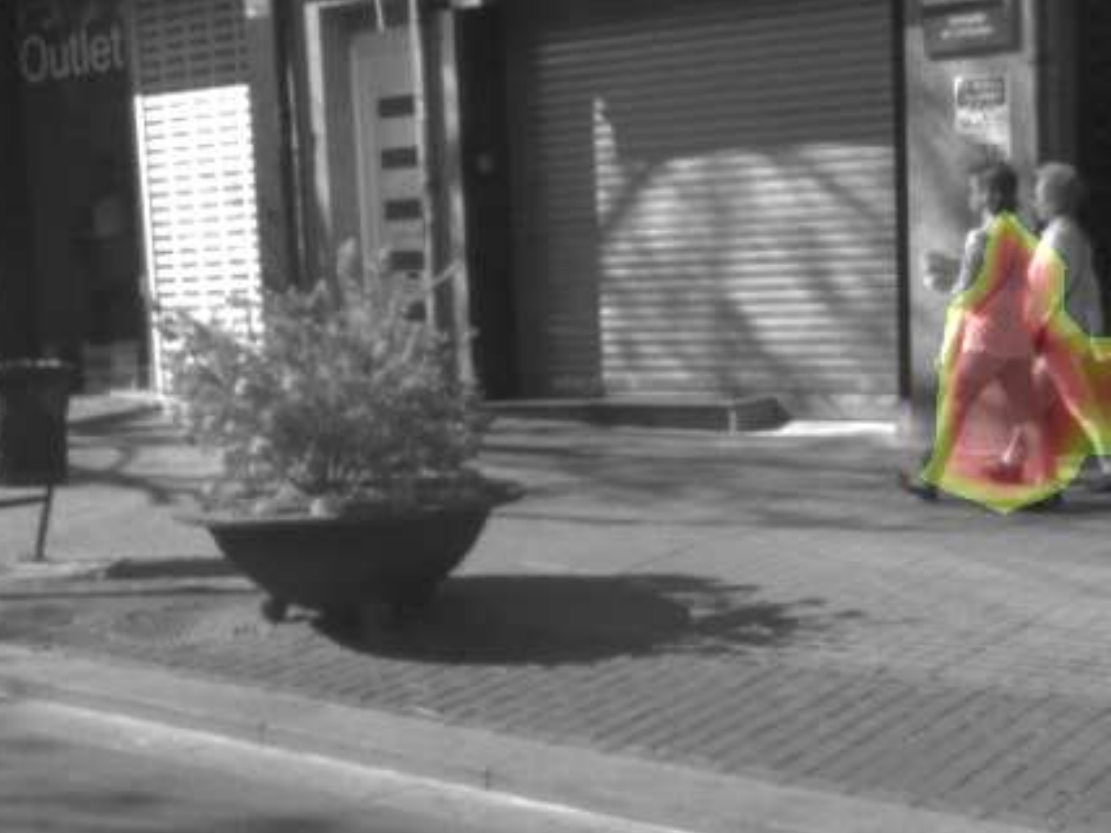}
		\end{minipage}
		\begin{minipage}{0.24\linewidth}
			\includegraphics[width=1\linewidth,trim=160 120 0 0,clip]{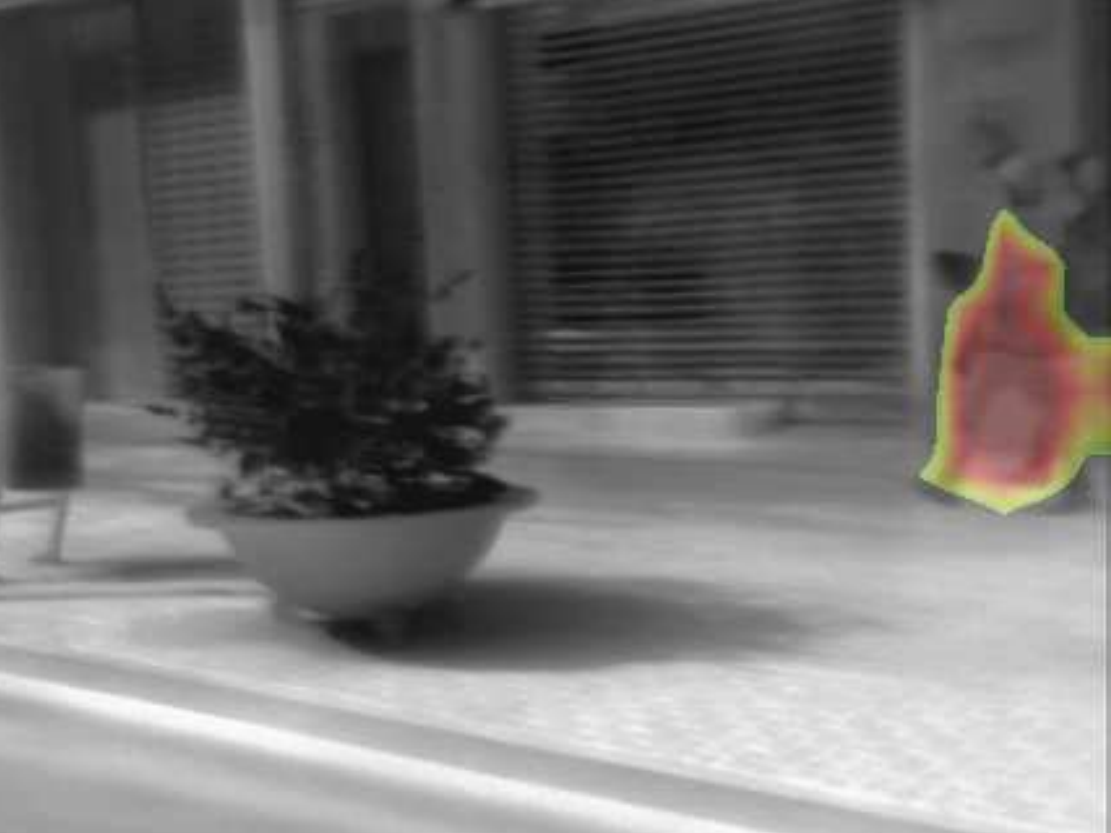}
		\end{minipage}
		\vspace{0.5mm}
	\end{minipage}
	\begin{minipage}{0.99\linewidth}
		\begin{minipage}{0.24\linewidth}
			\includegraphics[width=1\linewidth,trim=0 120 160 0,clip]{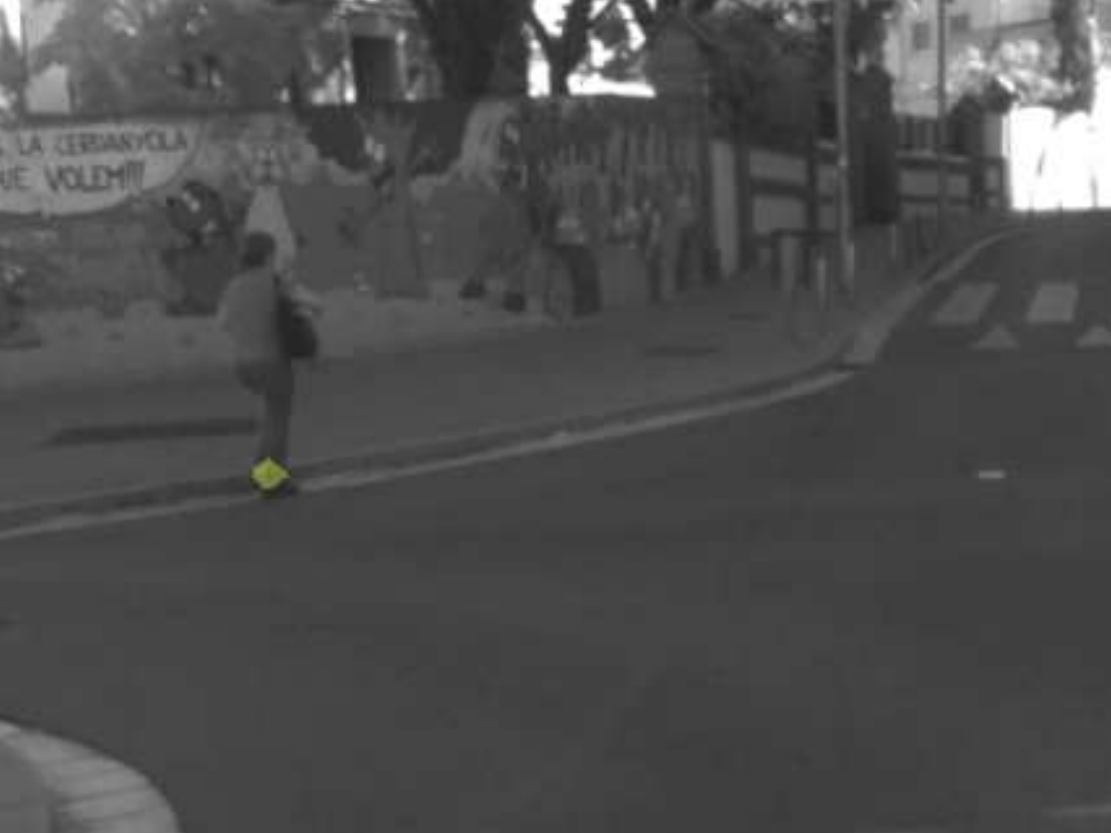}
		\end{minipage}
		\begin{minipage}{0.24\linewidth}
			\includegraphics[width=1\linewidth,trim=0 120 160 0,clip]{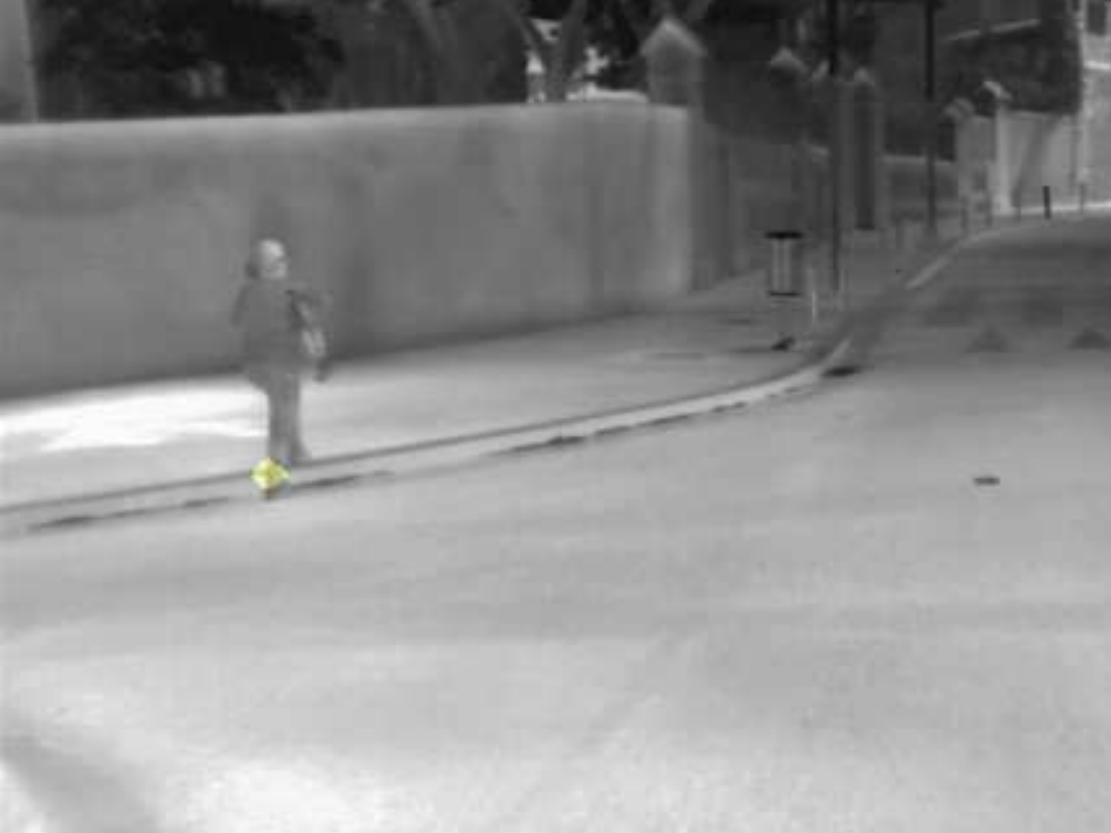}
		\end{minipage}
		\hspace{0.5mm}
		\begin{minipage}{0.24\linewidth}
			\includegraphics[width=1\linewidth,trim=0 120 160 0,clip]{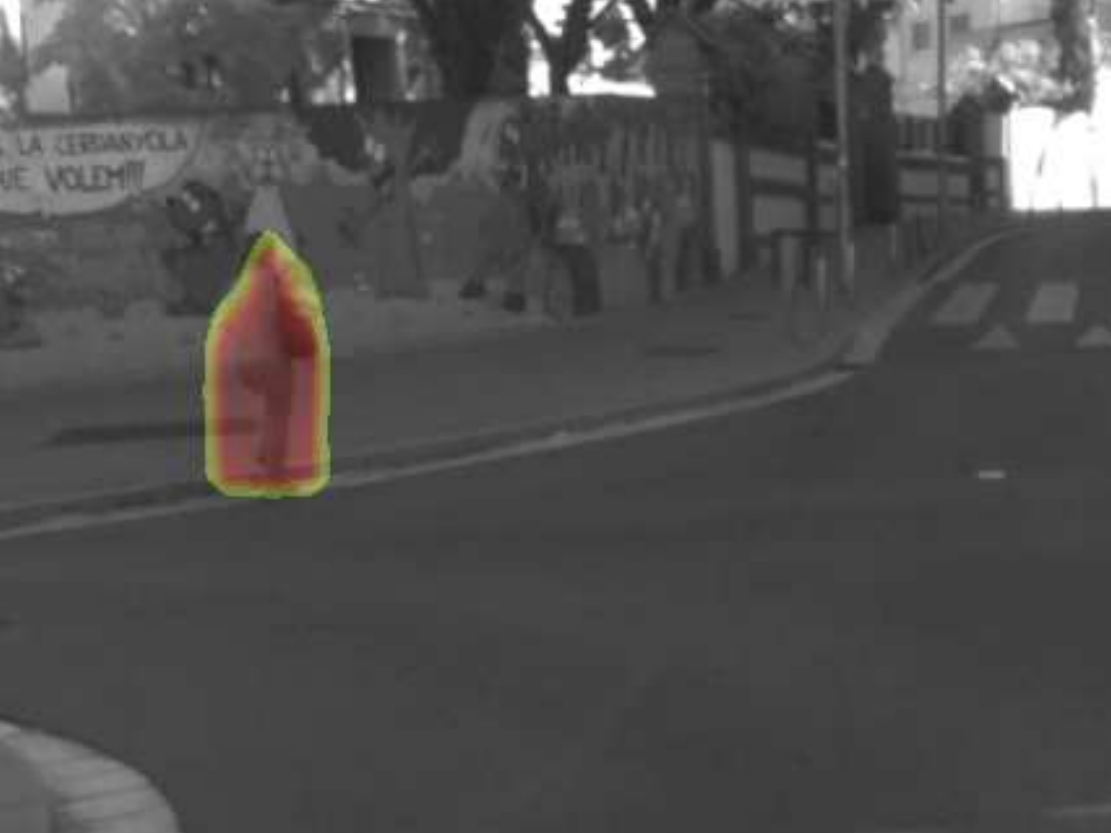}
		\end{minipage}
		\begin{minipage}{0.24\linewidth}
			\includegraphics[width=1\linewidth,trim=0 120 160 0,clip]{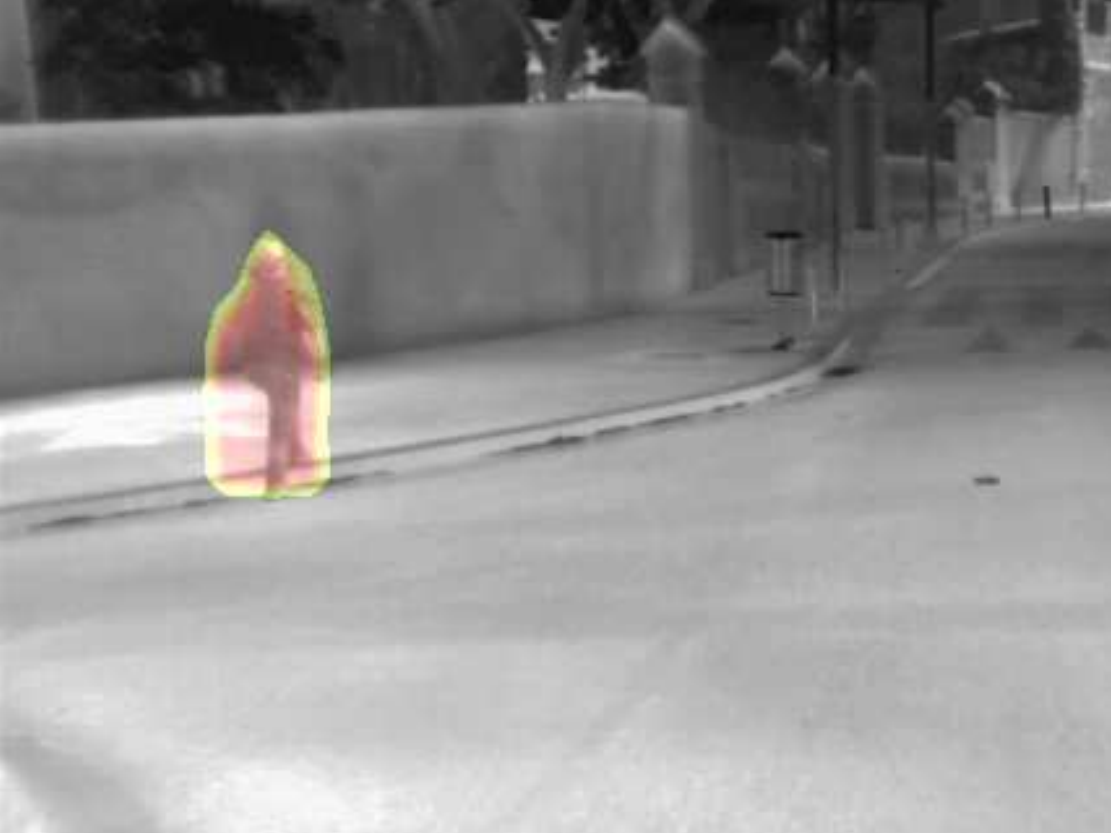}
		\end{minipage}
		\vspace{0.5mm}
	\end{minipage}
	\begin{minipage}{0.99\linewidth}
		\begin{minipage}{0.24\linewidth}
			\includegraphics[width=1\linewidth,trim=80 120 80 0,clip]{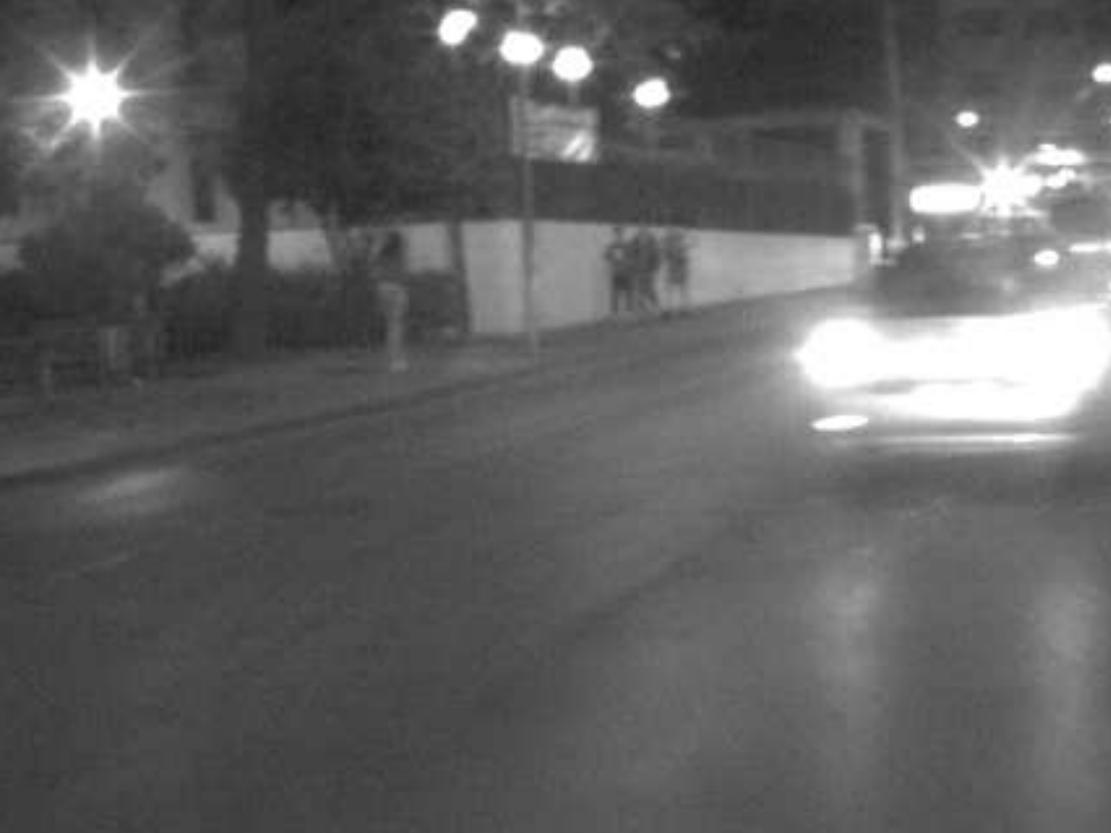}
		\end{minipage}
		\begin{minipage}{0.24\linewidth}
			\includegraphics[width=1\linewidth,trim=80 120 80 0,clip]{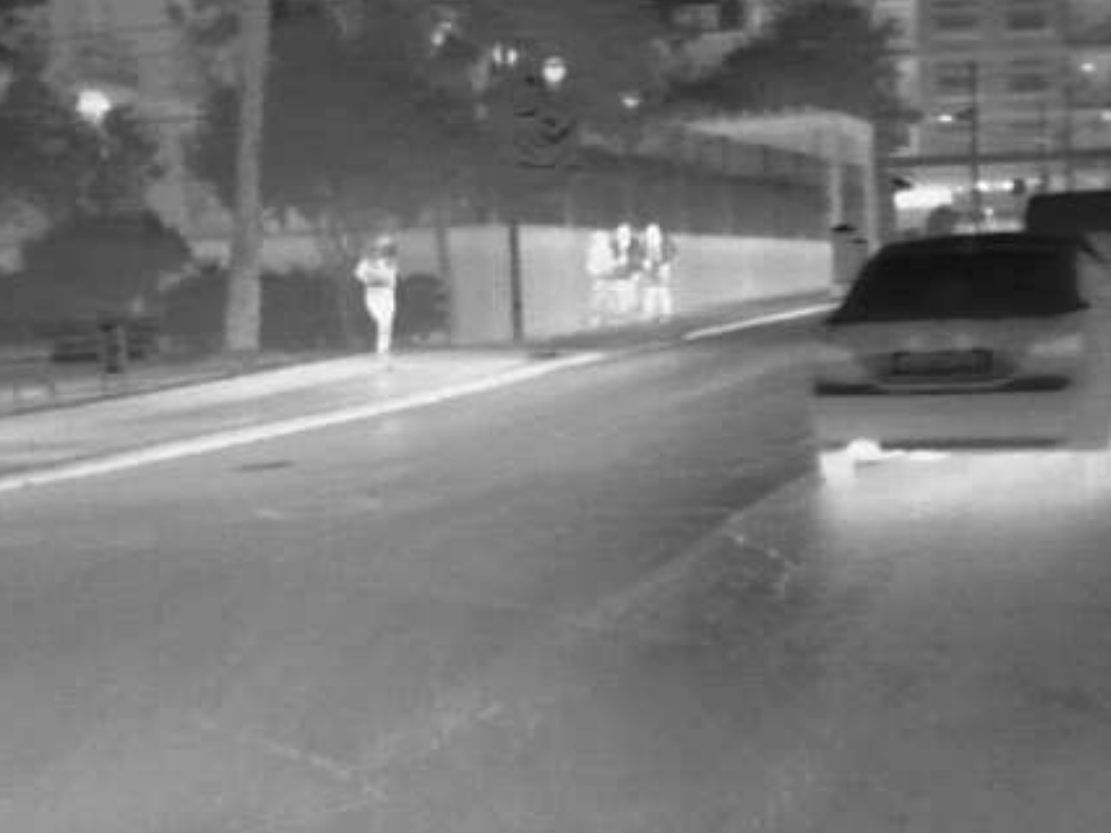}
		\end{minipage}
		\hspace{0.5mm}
		\begin{minipage}{0.24\linewidth}
			\includegraphics[width=1\linewidth,trim=80 120 80 0,clip]{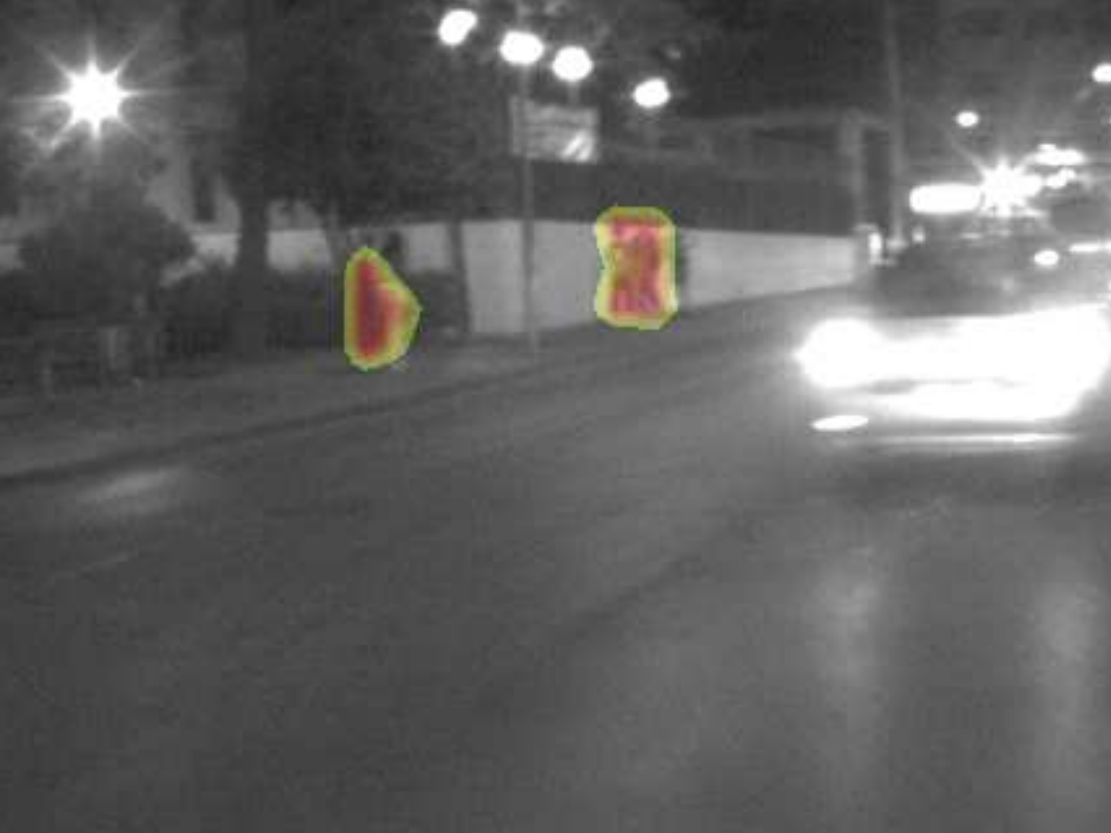}
		\end{minipage}
		\begin{minipage}{0.24\linewidth}
			\includegraphics[width=1\linewidth,trim=80 120 80 0,clip]{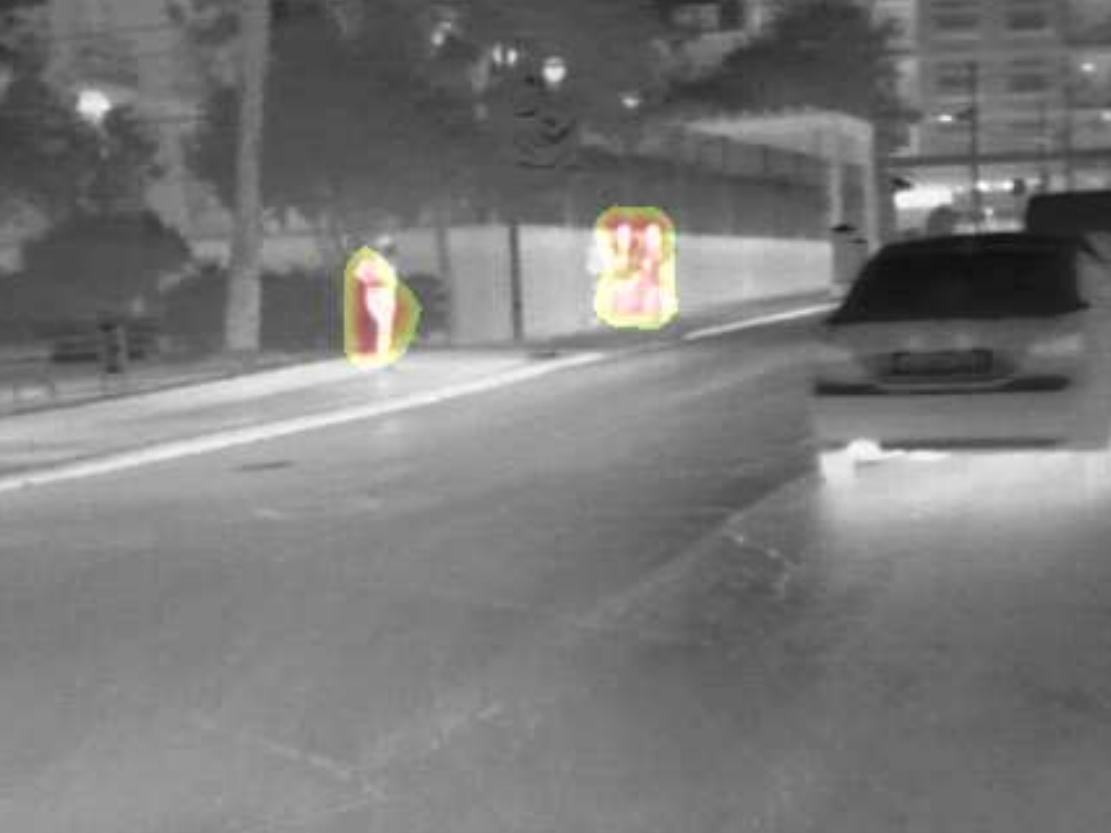}
		\end{minipage}
		\vspace{0.5mm}
	\end{minipage}
	\begin{minipage}{0.99\linewidth}
		\begin{minipage}{0.24\linewidth}
			\includegraphics[width=1\linewidth,trim=80 120 80 0,clip]{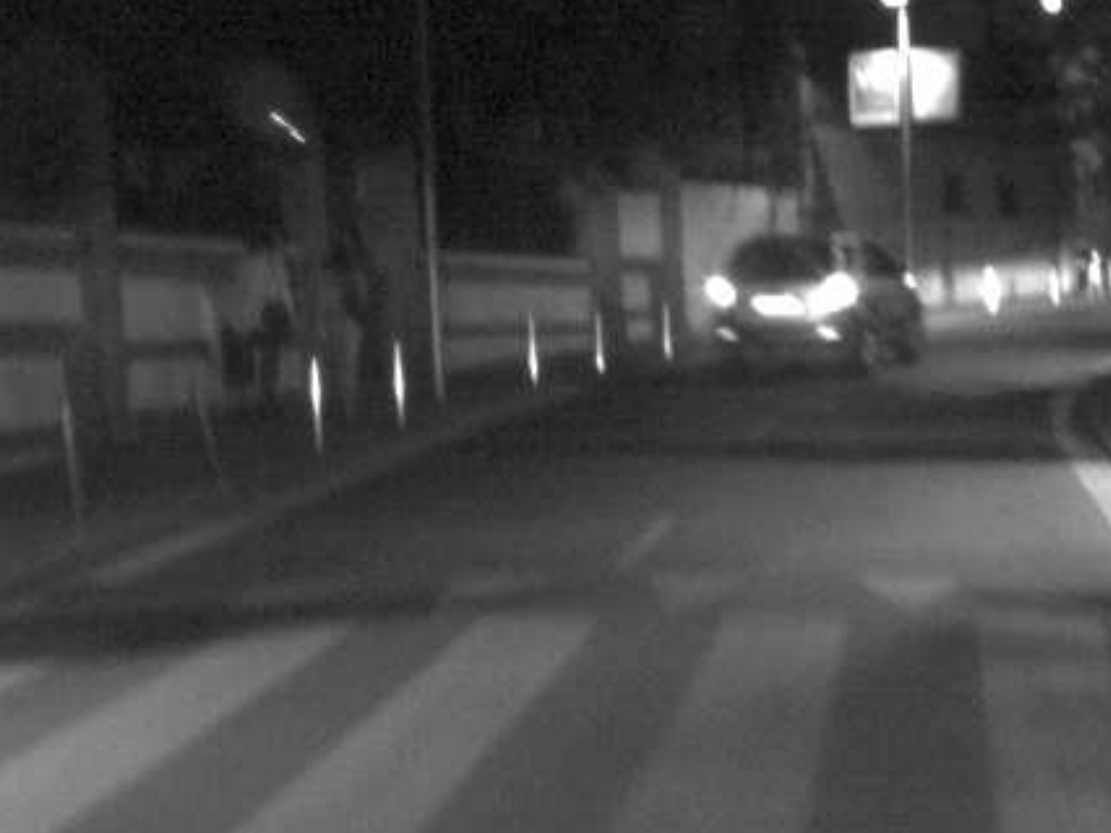}
		\end{minipage}
		\begin{minipage}{0.24\linewidth}
			\includegraphics[width=1\linewidth,trim=80 120 80 0,clip]{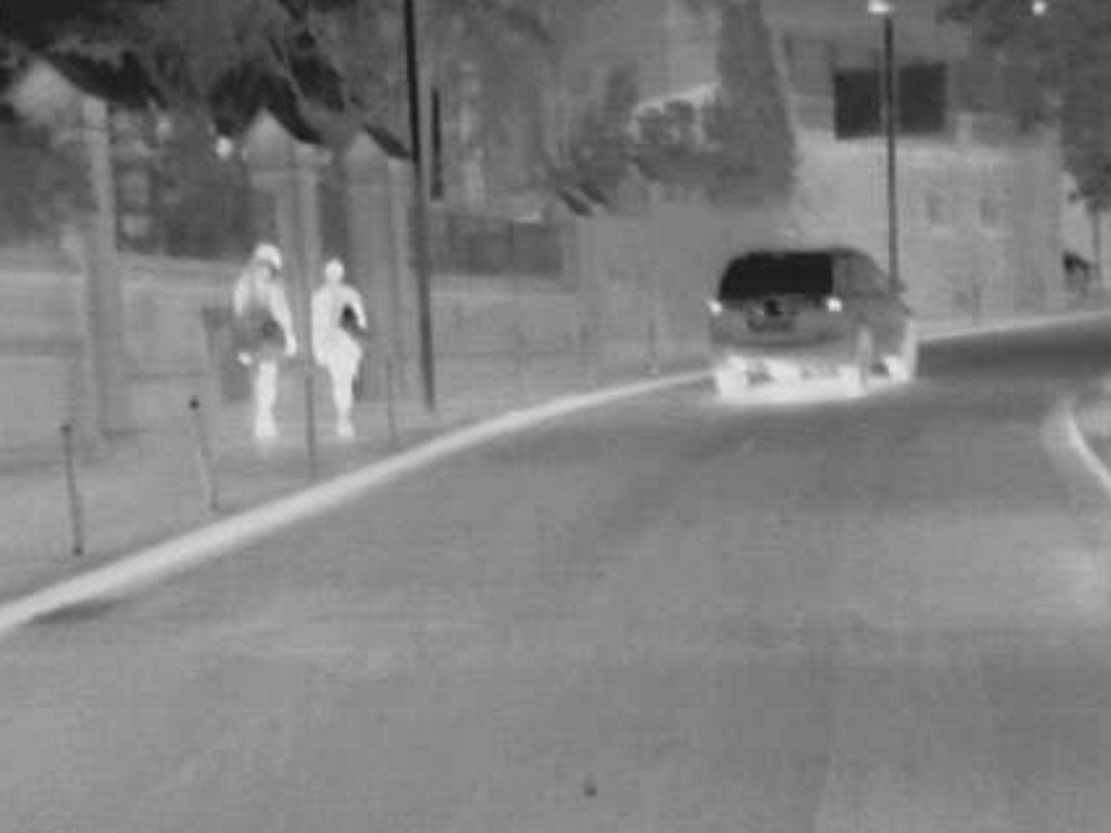}
		\end{minipage}
		\hspace{0.5mm}
		\begin{minipage}{0.24\linewidth}
			\includegraphics[width=1\linewidth,trim=80 120 80 0,clip]{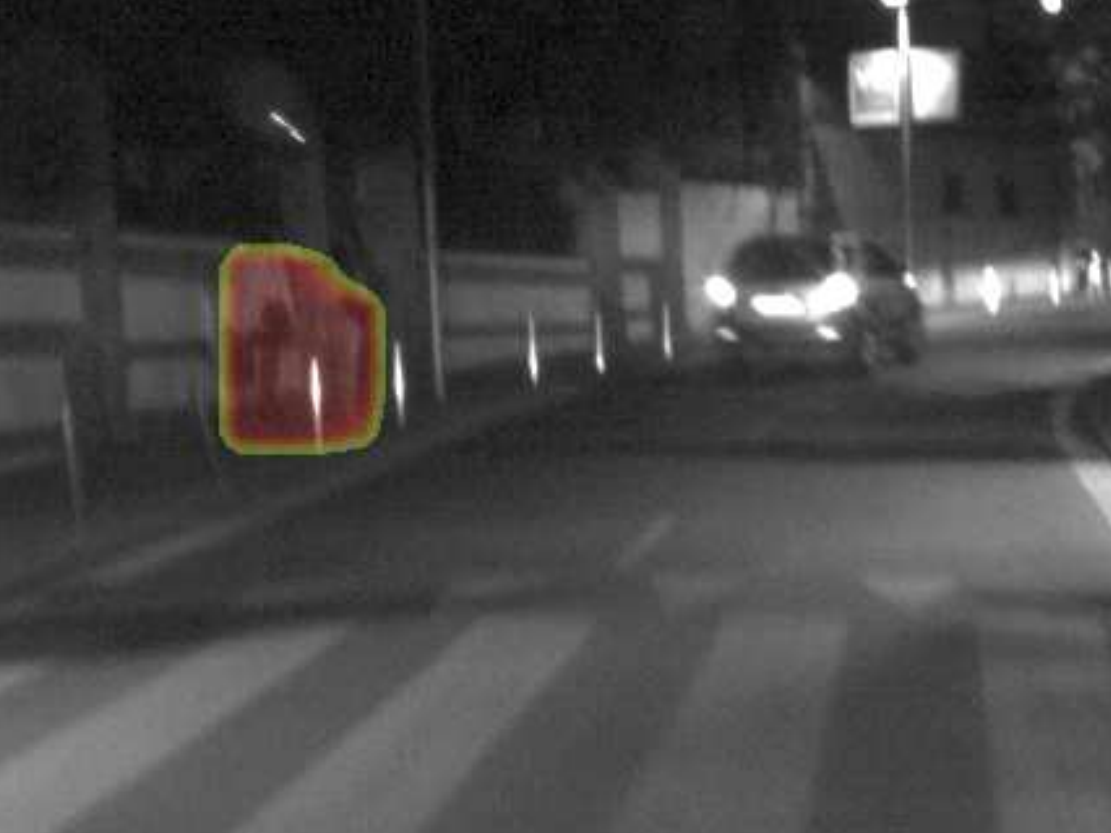}
		\end{minipage}
		\begin{minipage}{0.24\linewidth}
			\includegraphics[width=1\linewidth,trim=80 120 80 0,clip]{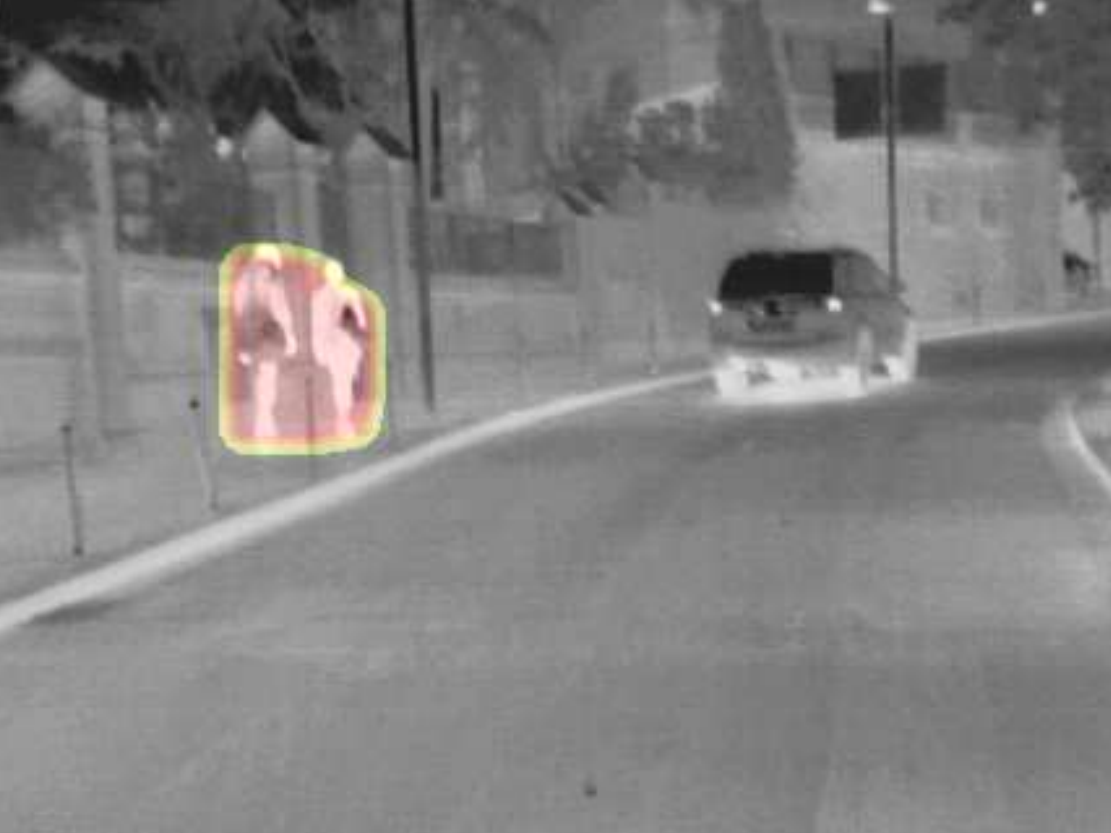}
		\end{minipage}
		\vspace{0.5mm}
	\end{minipage}
	\centering{(b)}
	\caption{ Comparing the qualitative performance of multispectral pedestrian detection results of UMDA and WMDA. (a) The performance of WMDA is not satisfactory while the UMDA can generate accurate detections; (b) WMDA can't detect any pedestrian regions while the UMDA is capable of generating accurate detections. }
	\label{fig5}
\end{figure}

In addition, the qualitative performance of multispectral pedestrian detection results of UMDA and WMDA are compared in Fig.~\ref{fig5}. We can observe that the UMDA can generate accurate detections in the case that the performance of WMDA is not satisfactory, as shown in Fig.~\ref{fig5}~(a). Even in the situation when WMDA can't detect any pedestrian regions, the UMDA is capable of generating accurate detections. 

The quantitative and qualitative comparison of UMDA and WMDA can prove that our proposed unsupervised multimodal domain adaptation framework is able to improve the multispectral pedestrian detection performance on target domain with a large margin.

\subsection{Comparison with the State-of-the-art}

We define the supervised multimodal domain adaptation (SMDA) model as our proposed multispectral pedestrian detector trained on target domain with manual annotations. The proposed SMDA and unsupervised multimodal domain adaptation (UMDA) models are compared with the current state-of-the-art multispectral pedestrian detectors, such as ACF+T+THOG~\cite{hwang2015multispectral}, Fusion RPN+BDT \cite{konig2017fully}, and HMFFN320~\cite{cao2019box} on target domain. Considering that these detectors were trained on the KAIST multispectral pedestrian benchmark, we fine-tune these detectors on the CVC-14 multispectral pedestrian dataset with supervised multimodal domain adaptation method.

The quantitative performance (pixel-wise AP~\cite{cao2019box}) of different multispectral pedestrian detectors are compared in Tab.~\ref{tab2}. It is observed that our proposed SMDA model outperforms the current state-of-the-art supervised multispectral pedestrian detectors, pixel-level AP~\cite{cao2019box} of SMDA is 0.91\% higher than the results of HMFFN320~\cite{cao2019box} and 5.18\% higher than the ones of Fusion RPN+BDT~\cite{konig2017fully}. Considering that our proposed SMDA model incorporates the visible and thermal pedestrian detection supervision module into the HMFFN320~\cite{cao2019box} model, we can prove that the two-stream detection supervision module is able to provide additional feature information to facilitate the training of multispectral pedestrian detectors. 

\begin{table}[h]
	\centering
	\caption{ Comparing the quantitative performance (pixel-wise AP~\cite{cao2019box}) of UMDA and SMDA with the current state-of-the-art methods. Please note that ACF+T+THOG~\cite{hwang2015multispectral}, Fusion RPN+BDT \cite{konig2017fully}, HMFFN320~\cite{cao2019box}, and SMDA are supervised multispectral pedestrian detectors; UMDA is unsupervised domain adaptation model for multispectral pedestrian detection.}
	\begin{tabular}{cccc}
		\hline
		Model &All-day	&Daytime	&Nighttime \\
		\hline
		{ACF+T+THOG~\cite{hwang2015multispectral}}	&0.7139	&0.6926	&0.7334	\\ 
		{Fusion RPN+BDT~\cite{konig2017fully}}	&0.8172	&0.8103	&0.8241	\\ 
%		{IATDNN+IAMSS~\cite{guan2018fusion}}	&0.8333	&0.8272	&0.8419	\\ 
		{HMFFN320~\cite{cao2019box}}	&0.8599	&0.8355	&0.8942	\\ 
		\textbf{SMDA (ours)} & \textbf{0.8690} &\textbf{0.8485}	&\textbf{0.8944} \\ 
		\textbf{UMDA (ours)} & \textbf{0.8023} &\textbf{0.7688}	&\textbf{0.8503} \\  
		\hline		
	\end{tabular}
	\label{tab2}
\end{table}

\begin{figure}[t]
	\centering	{ SMDA  \qquad \qquad \qquad \qquad \quad UMDA }
	\begin{minipage}{0.99\linewidth}
		\begin{minipage}{0.24\linewidth}
			\includegraphics[width=1\linewidth,trim=160 120 0 0,clip]{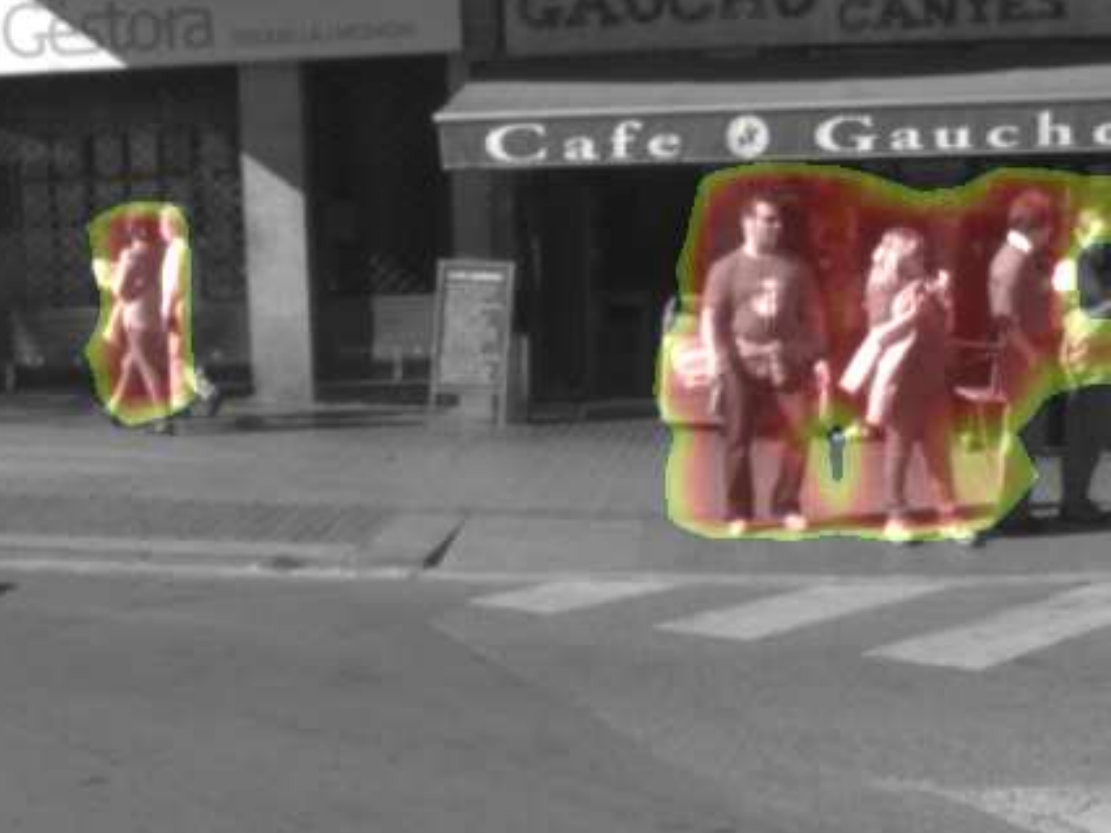}
		\end{minipage}
		\begin{minipage}{0.24\linewidth}
			\includegraphics[width=1\linewidth,trim=160 120 0 0,clip]{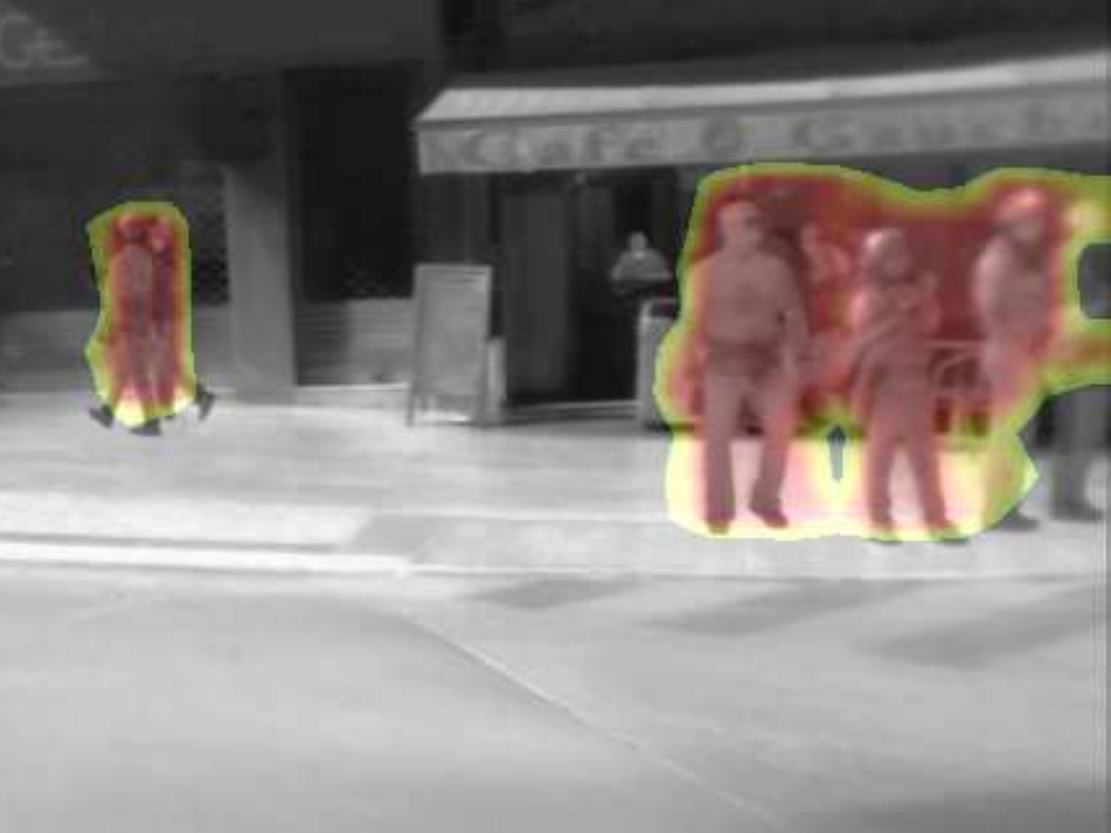}
		\end{minipage}
		\hspace{0.5mm}
		\begin{minipage}{0.24\linewidth}
			\includegraphics[width=1\linewidth,trim=160 120 0 0,clip]{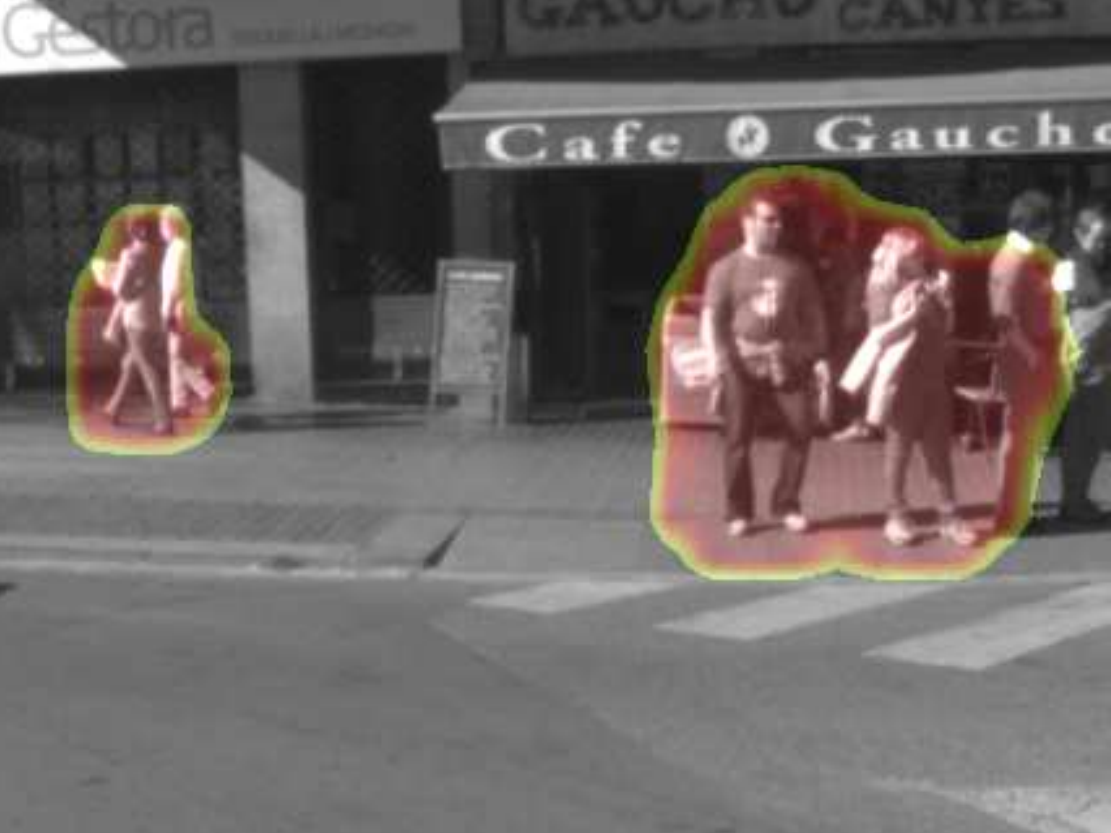}
		\end{minipage}
		\begin{minipage}{0.24\linewidth}
			\includegraphics[width=1\linewidth,trim=160 120 0 0,clip]{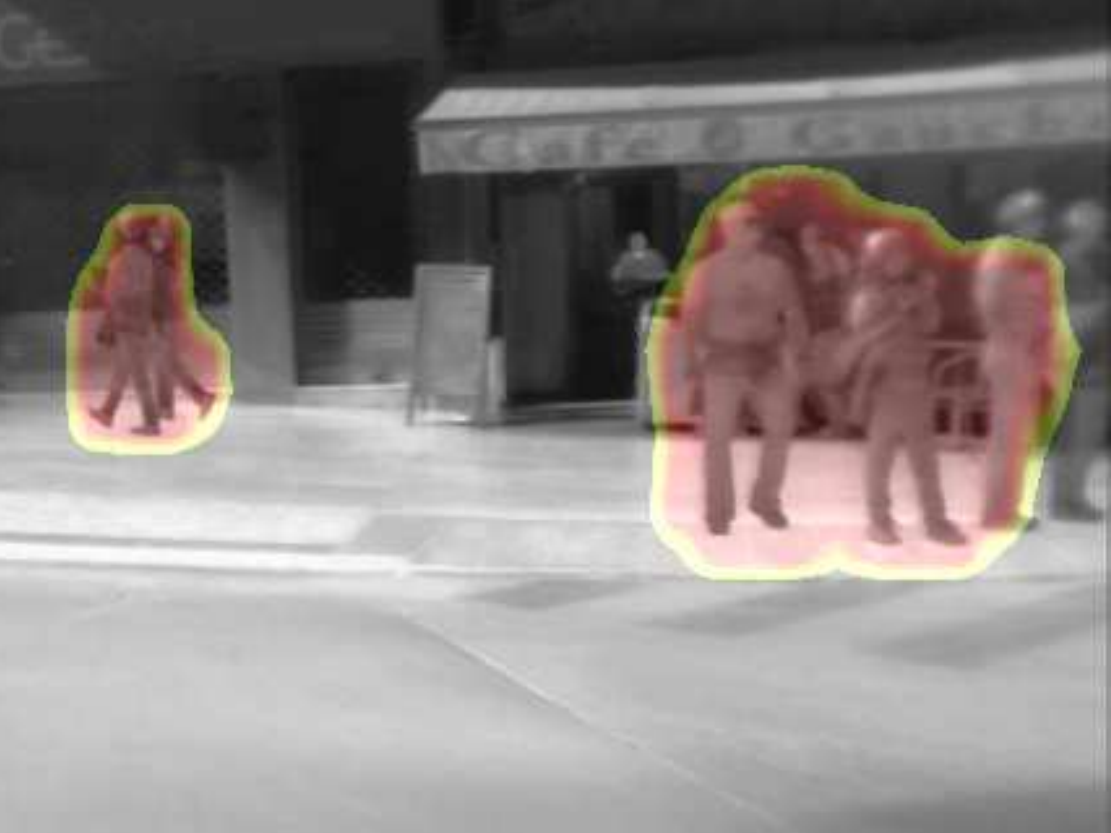}
		\end{minipage}
		\vspace{0.5mm}
	\end{minipage}
	\begin{minipage}{0.99\linewidth}
		\begin{minipage}{0.24\linewidth}
			\includegraphics[width=1\linewidth,trim=0 0 0 0,clip]{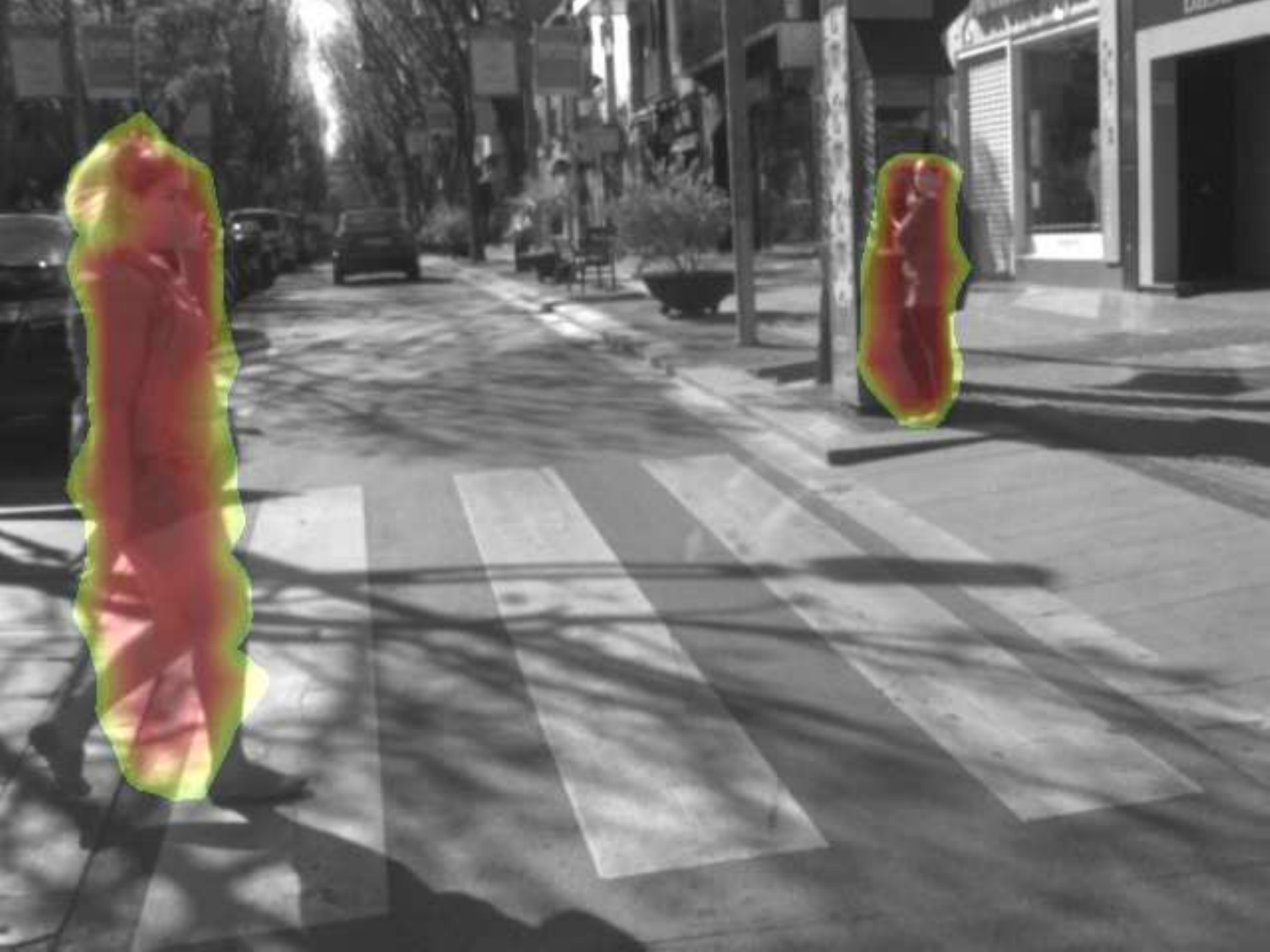}
		\end{minipage}
		\begin{minipage}{0.24\linewidth}
			\includegraphics[width=1\linewidth,trim=0 0 0 0,clip]{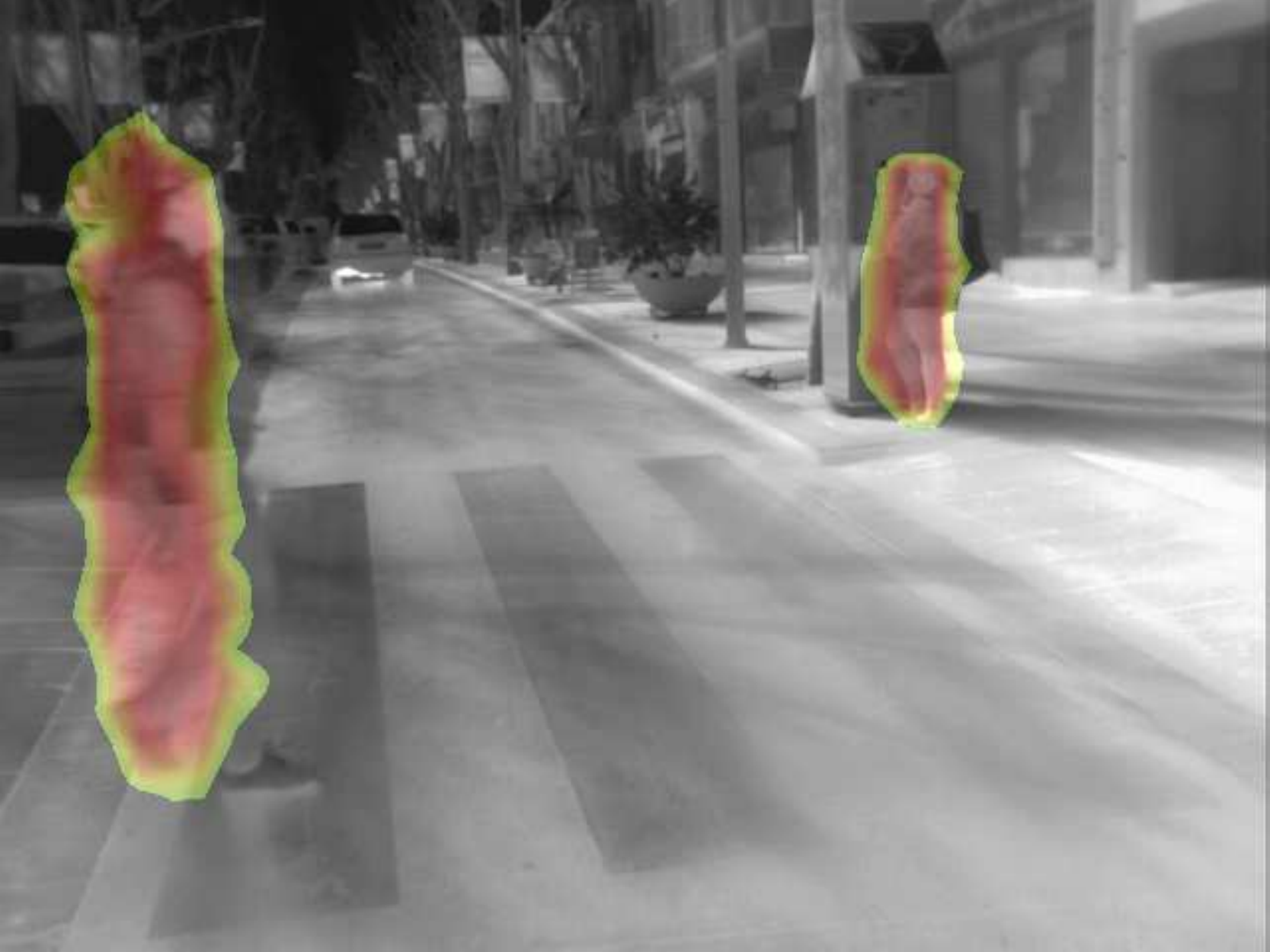}
		\end{minipage}
		\hspace{0.5mm}
		\begin{minipage}{0.24\linewidth}
			\includegraphics[width=1\linewidth,trim=0 0 0 0,clip]{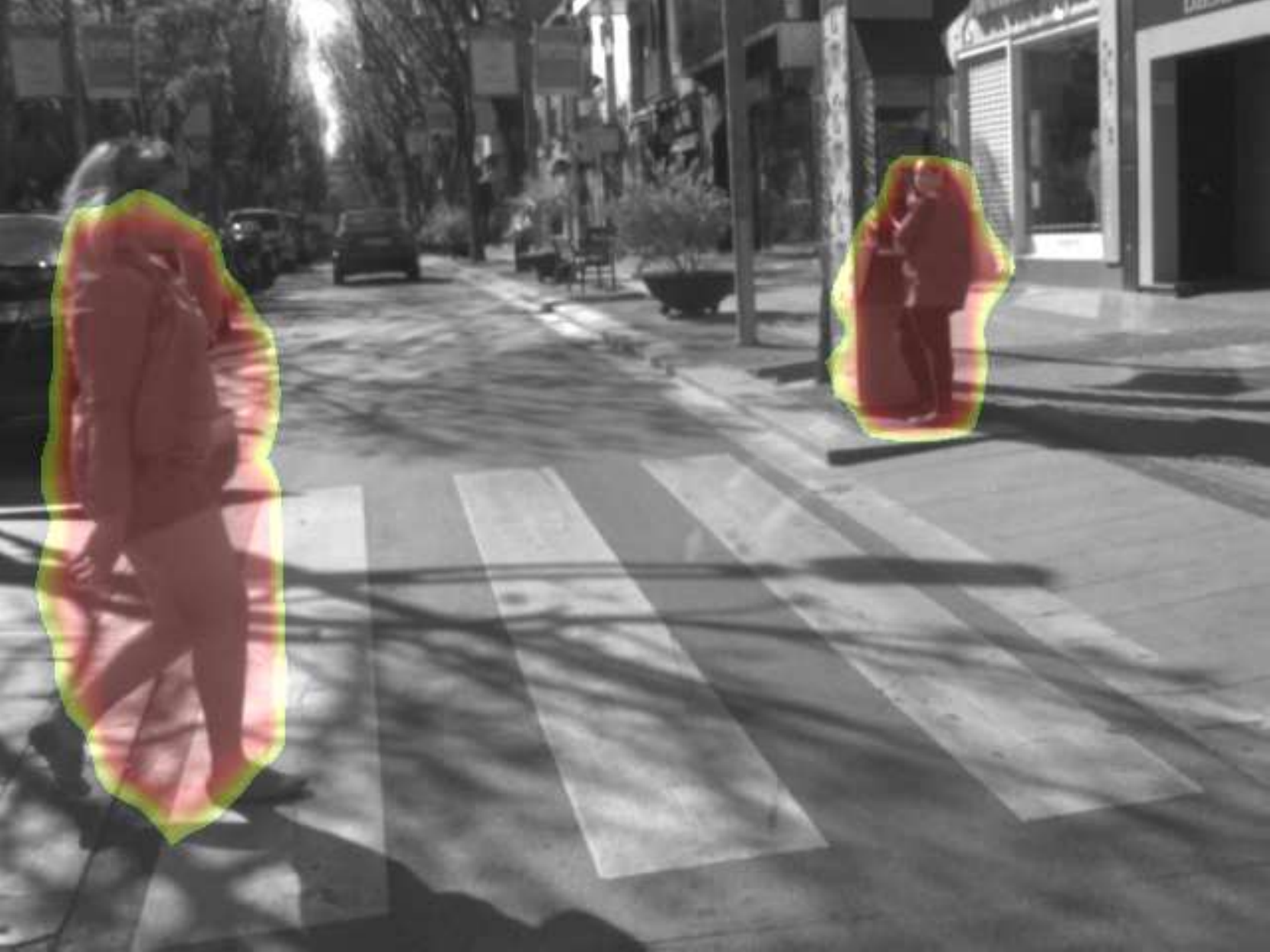}
		\end{minipage}
		\begin{minipage}{0.24\linewidth}
			\includegraphics[width=1\linewidth,trim=0 0 0 0,clip]{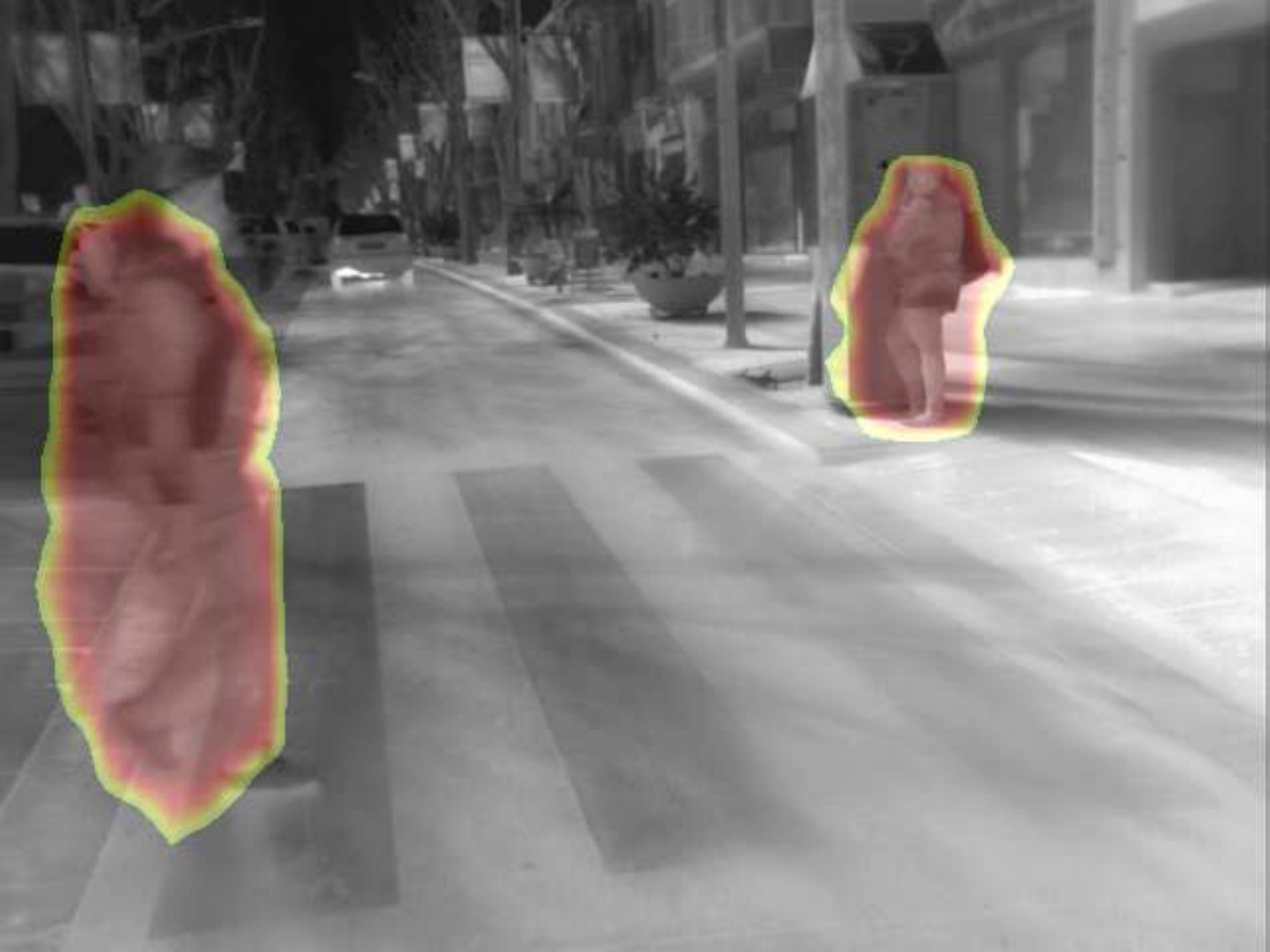}
		\end{minipage}
		\vspace{0.5mm}
	\end{minipage}
	\begin{minipage}{0.99\linewidth}
		\begin{minipage}{0.24\linewidth}
			\includegraphics[width=1\linewidth,trim=150 120 10 0,clip]{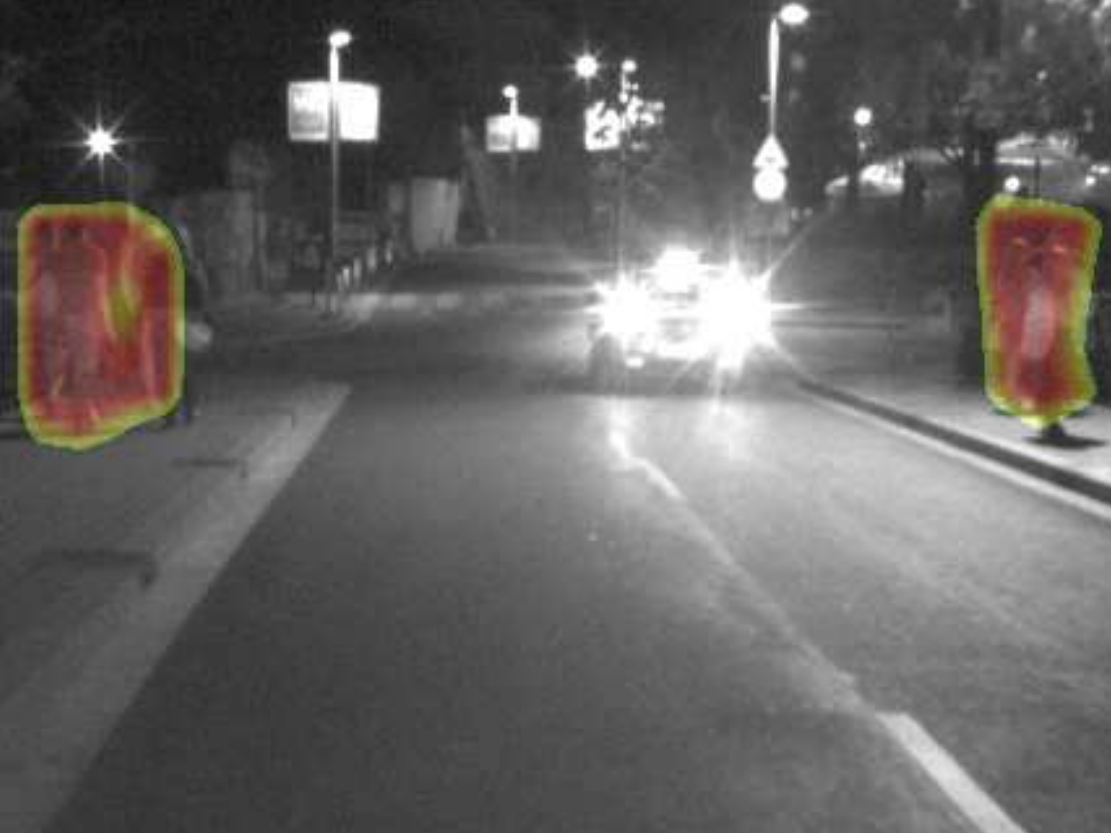}
		\end{minipage}
		\begin{minipage}{0.24\linewidth}
			\includegraphics[width=1\linewidth,trim=150 120 10 0,clip]{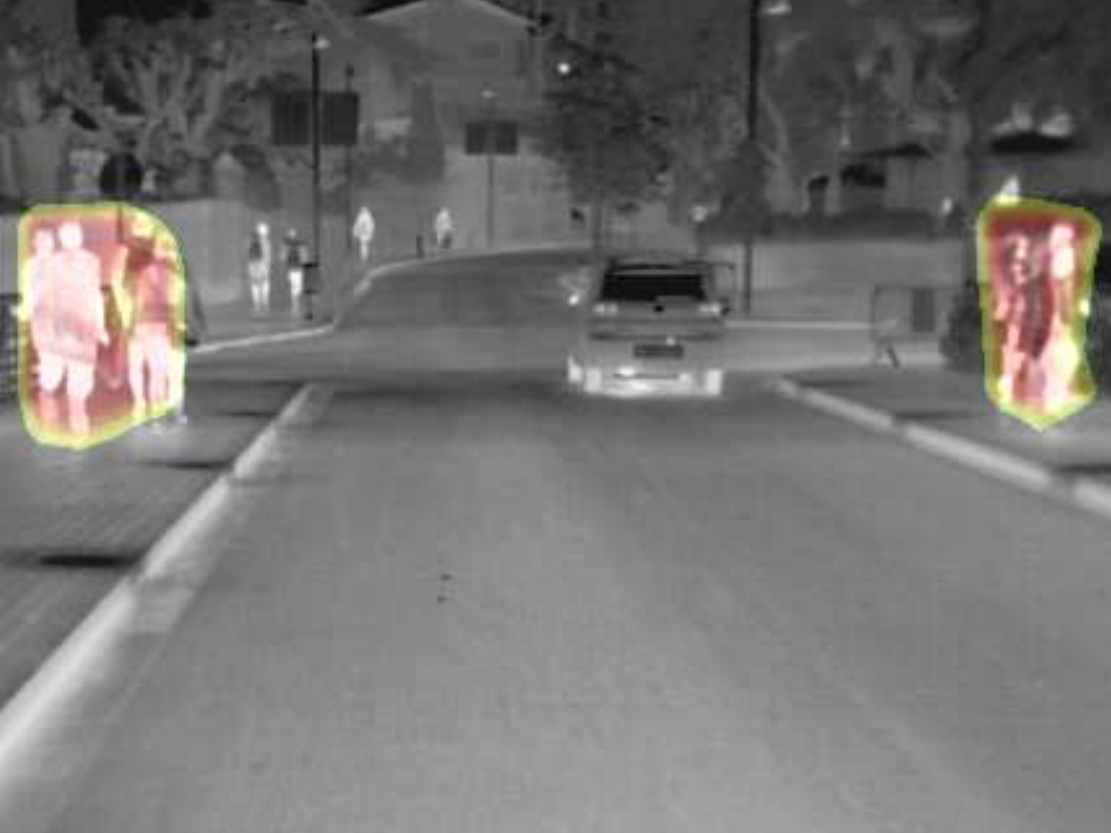}
		\end{minipage}
		\hspace{0.5mm}
		\begin{minipage}{0.24\linewidth}
			\includegraphics[width=1\linewidth,trim=150 120 10 0,clip]{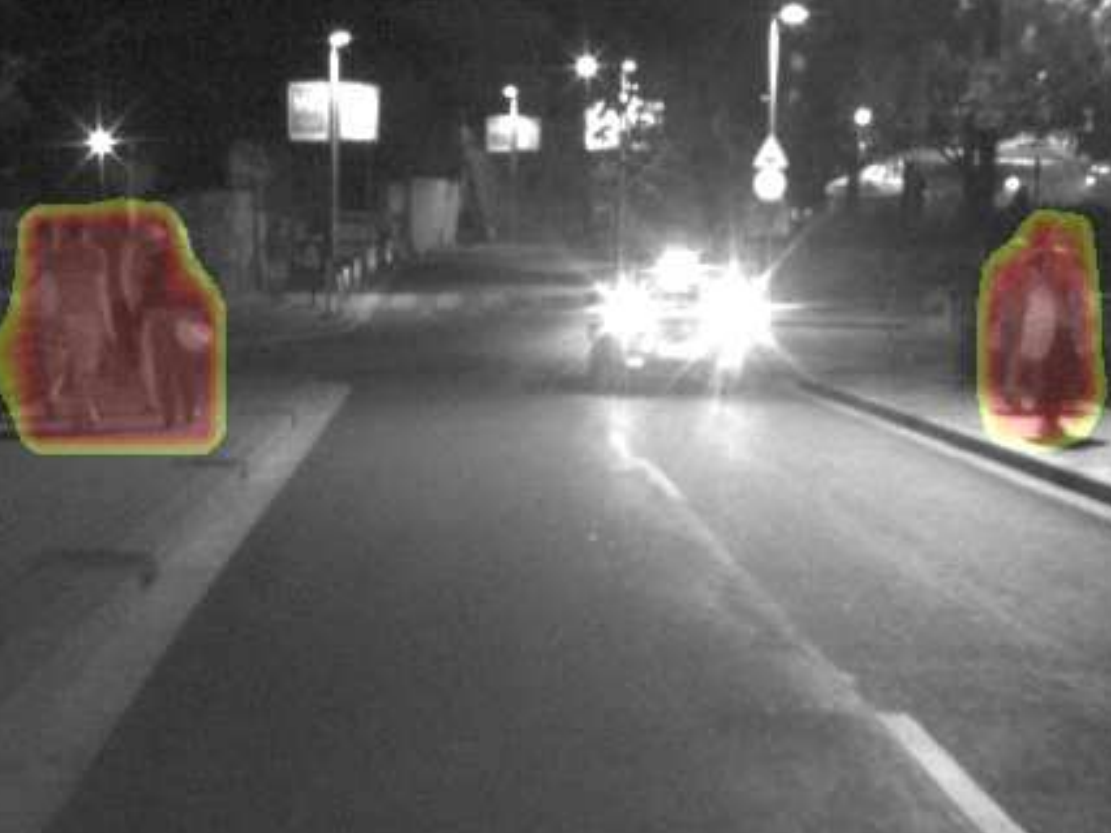}
		\end{minipage}
		\begin{minipage}{0.24\linewidth}
			\includegraphics[width=1\linewidth,trim=150 120 10 0,clip]{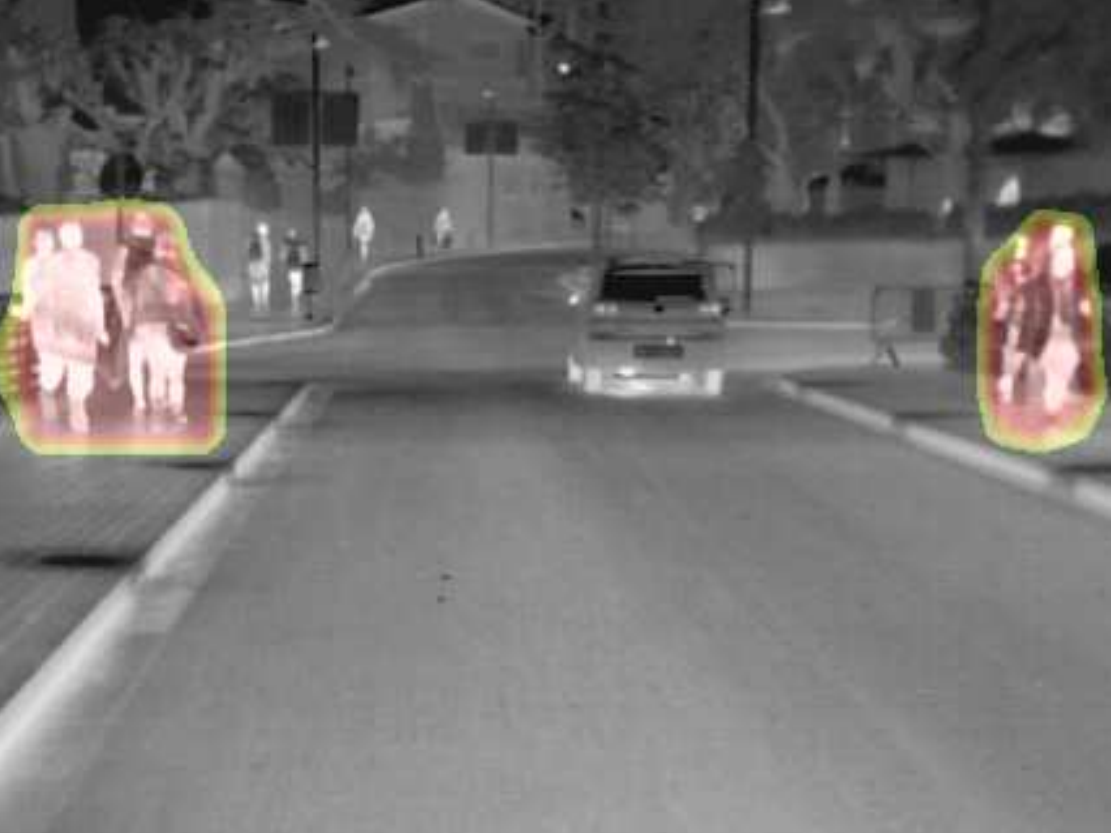}
		\end{minipage}
		\vspace{0.5mm}
	\end{minipage}
	\begin{minipage}{0.99\linewidth}
		\begin{minipage}{0.24\linewidth}
			\includegraphics[width=1\linewidth,trim=0 120 160 0,clip]{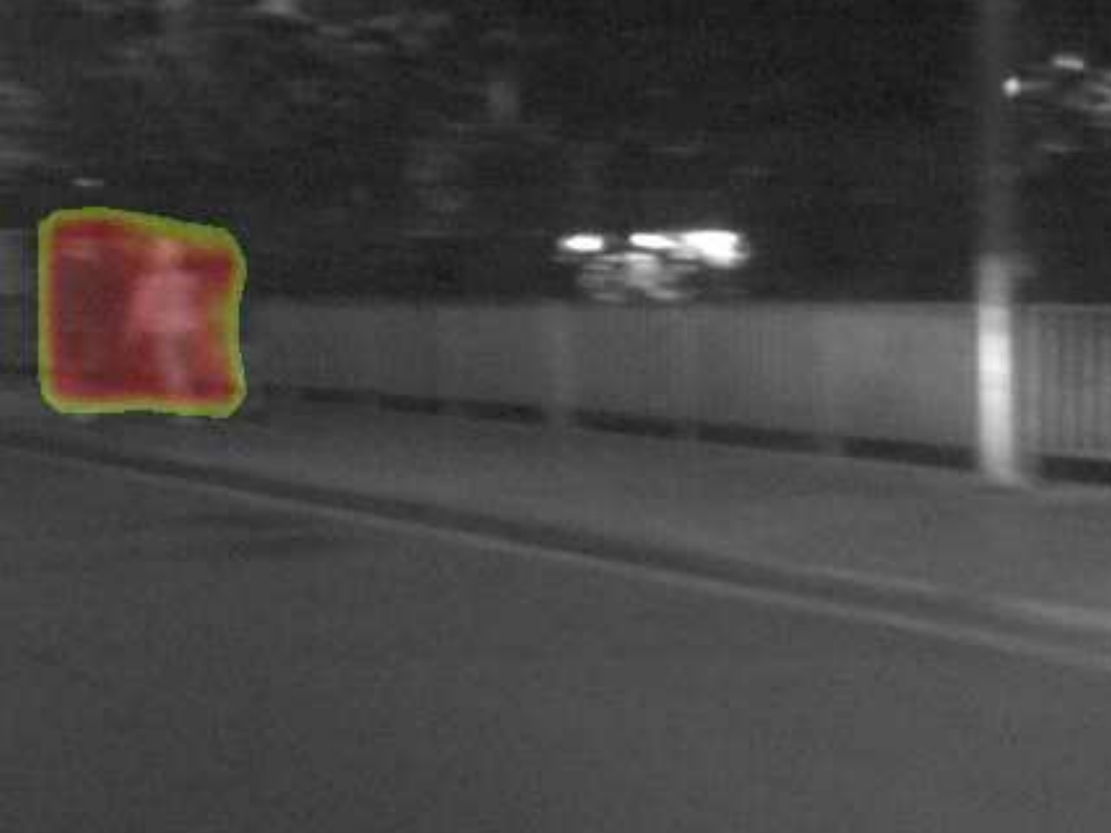}
		\end{minipage}
		\begin{minipage}{0.24\linewidth}
			\includegraphics[width=1\linewidth,trim=0 120 160 0,clip]{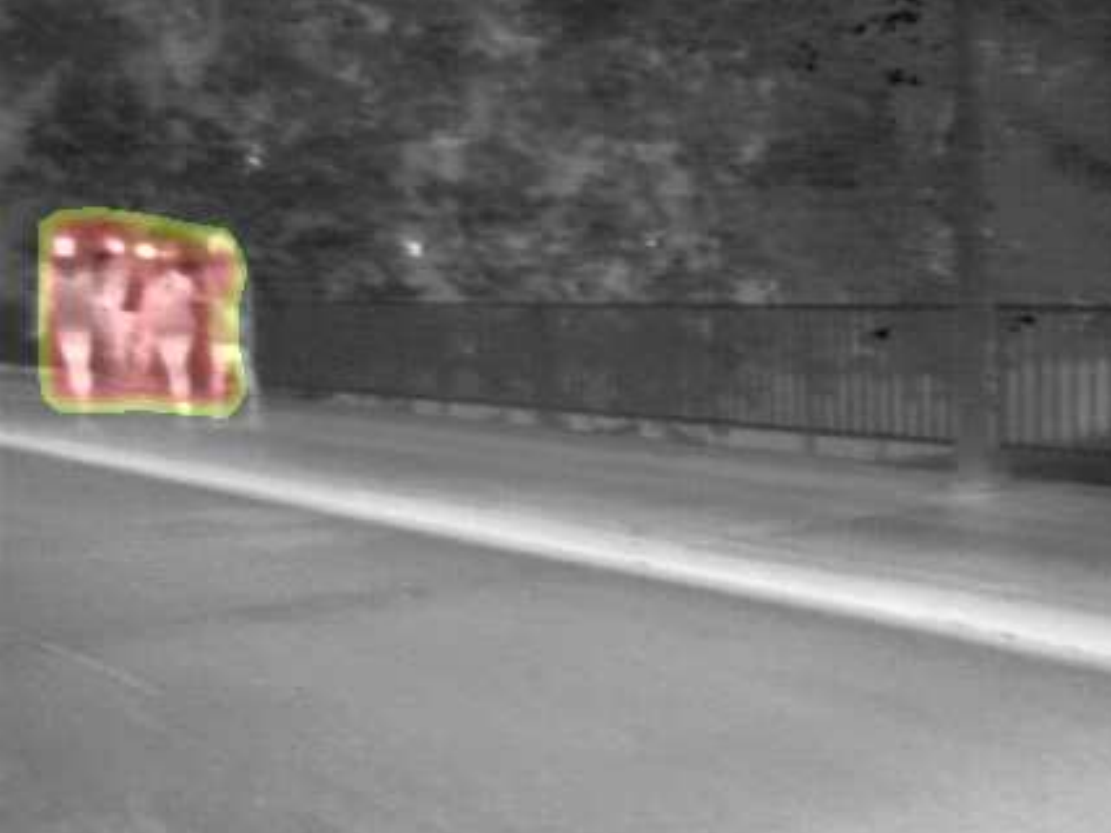}
		\end{minipage}
		\hspace{0.5mm}
		\begin{minipage}{0.24\linewidth}
			\includegraphics[width=1\linewidth,trim=0 120 160 0,clip]{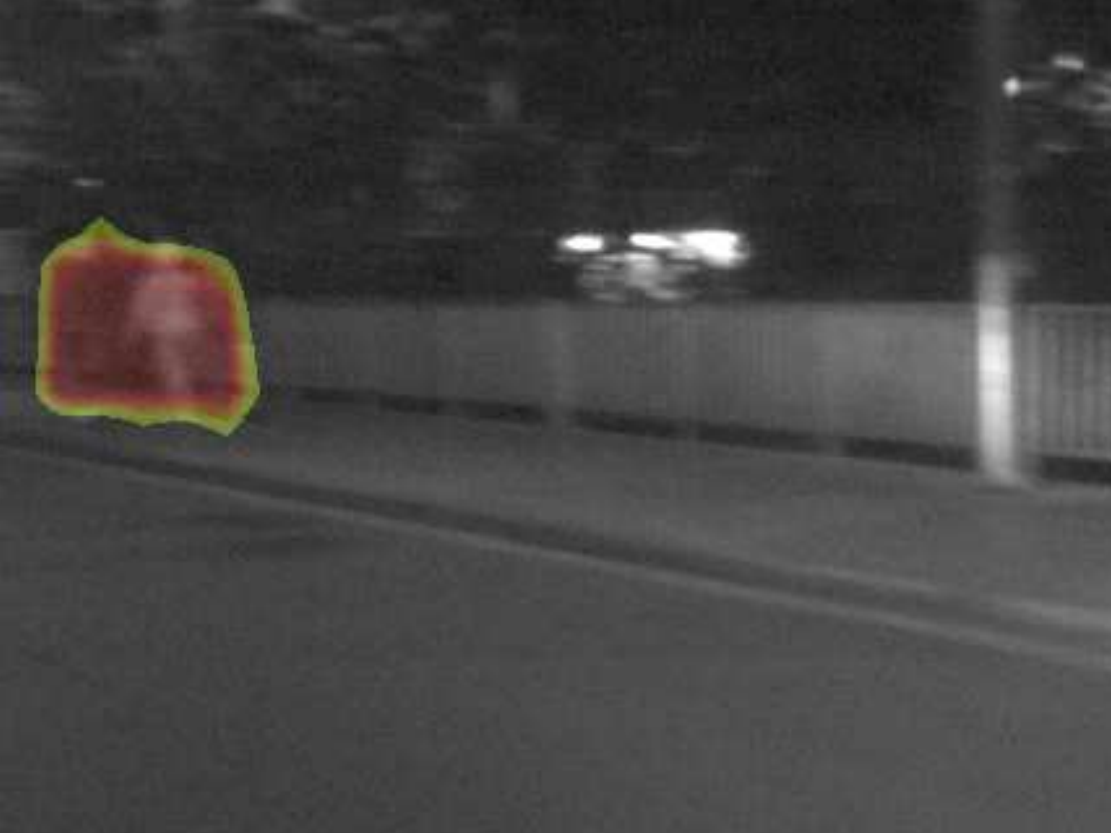}
		\end{minipage}
		\begin{minipage}{0.24\linewidth}
			\includegraphics[width=1\linewidth,trim=0 120 160 0,clip]{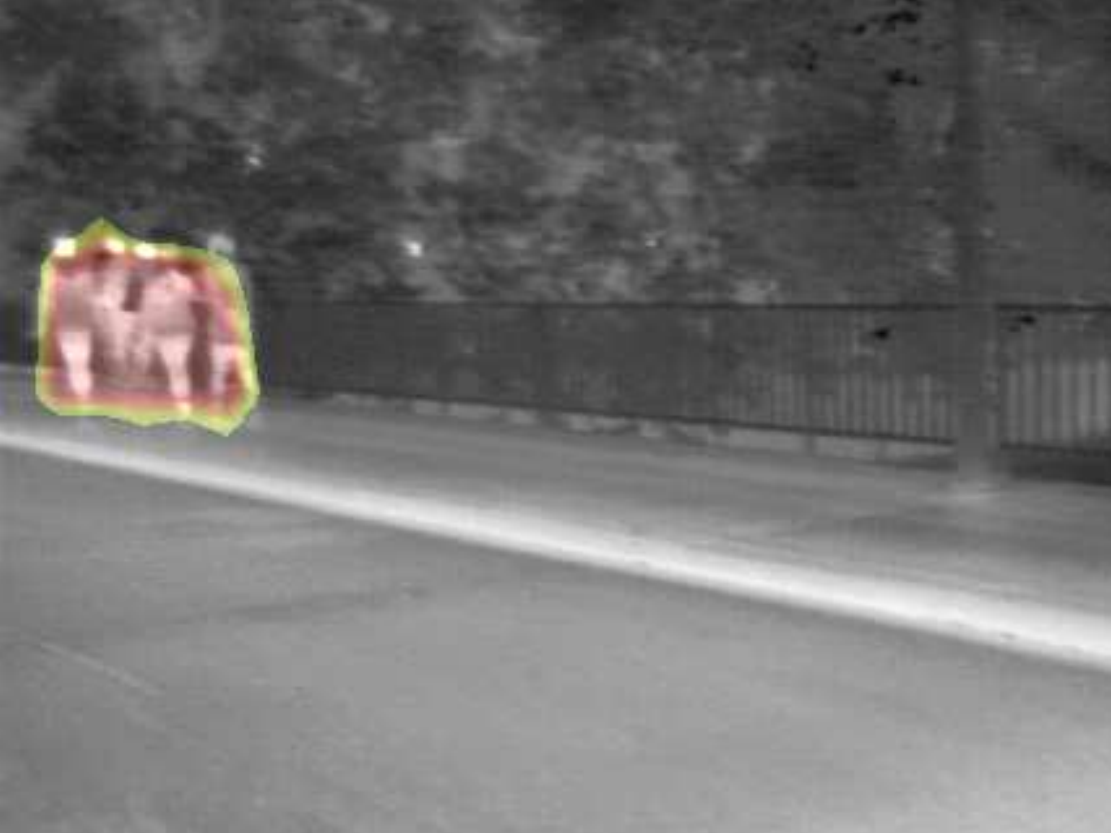}
		\end{minipage}
		\vspace{0.5mm}
	\end{minipage}
	\centering{(a)}
	\begin{minipage}{0.99\linewidth}
		\begin{minipage}{0.24\linewidth}
			\includegraphics[width=1\linewidth,trim=160 120 0 0,clip]{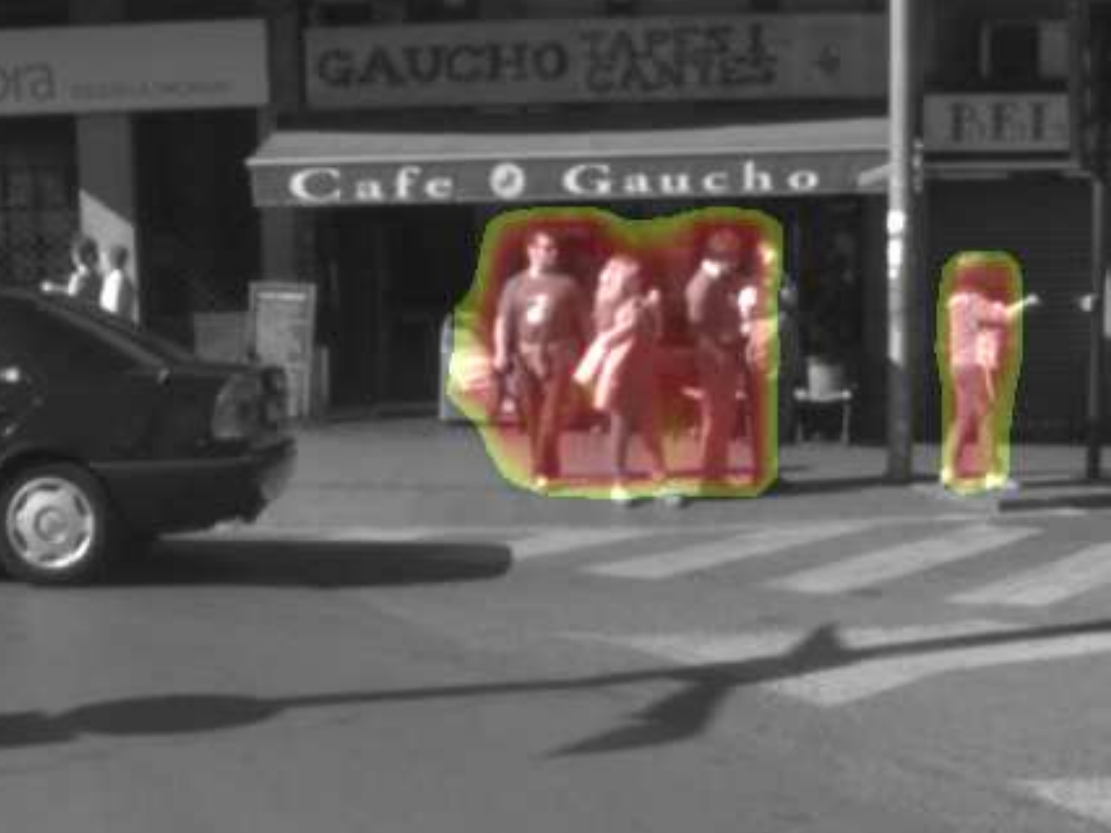}
		\end{minipage}
		\begin{minipage}{0.24\linewidth}
			\includegraphics[width=1\linewidth,trim=160 120 0 0,clip]{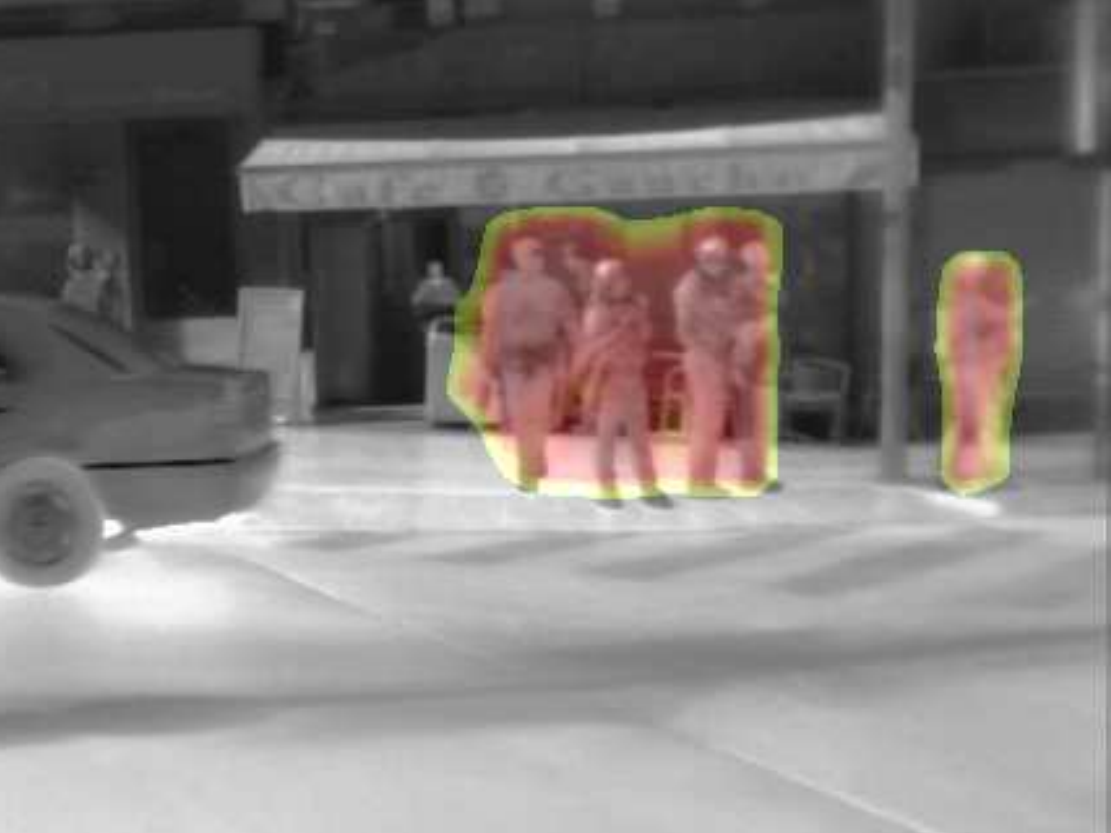}
		\end{minipage}
		\hspace{0.5mm}
		\begin{minipage}{0.24\linewidth}
			\includegraphics[width=1\linewidth,trim=160 120 0 0,clip]{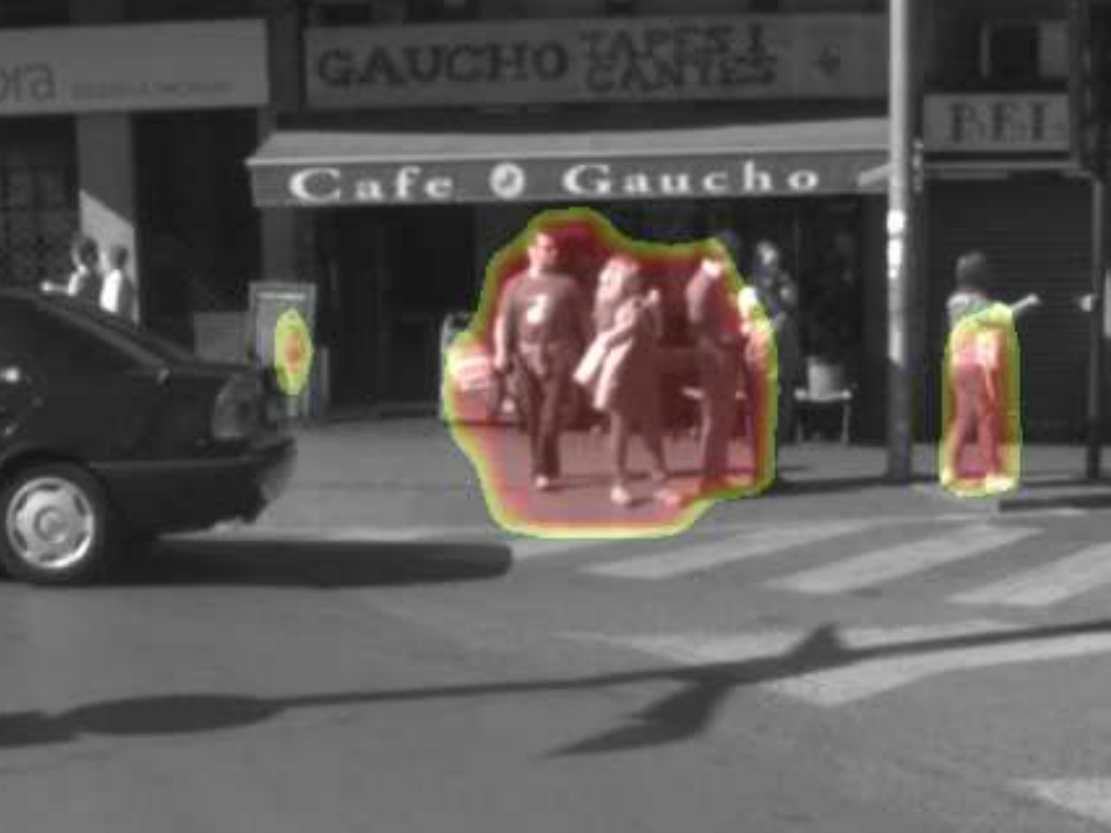}
		\end{minipage}
		\begin{minipage}{0.24\linewidth}
			\includegraphics[width=1\linewidth,trim=160 120 0 0,clip]{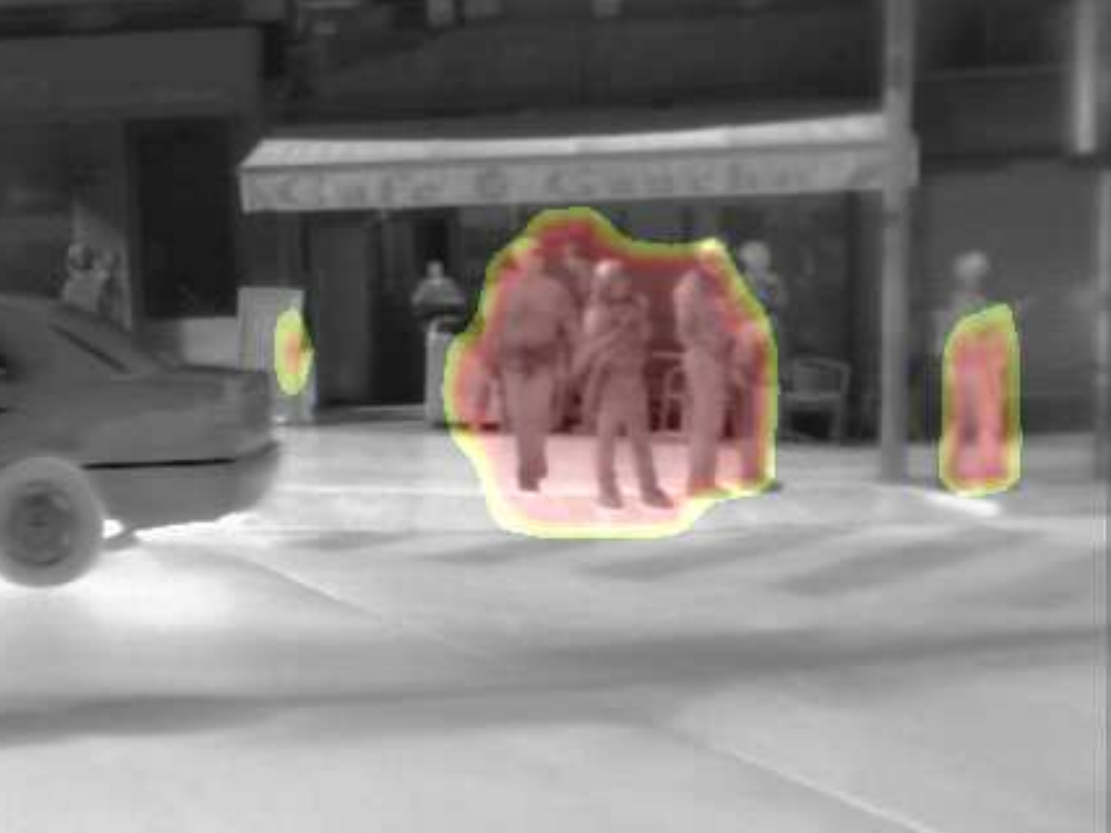}
		\end{minipage}
		\vspace{0.5mm}
	\end{minipage}
	\begin{minipage}{0.99\linewidth}
		\begin{minipage}{0.24\linewidth}
			\includegraphics[width=1\linewidth,trim=0 120 160 0,clip]{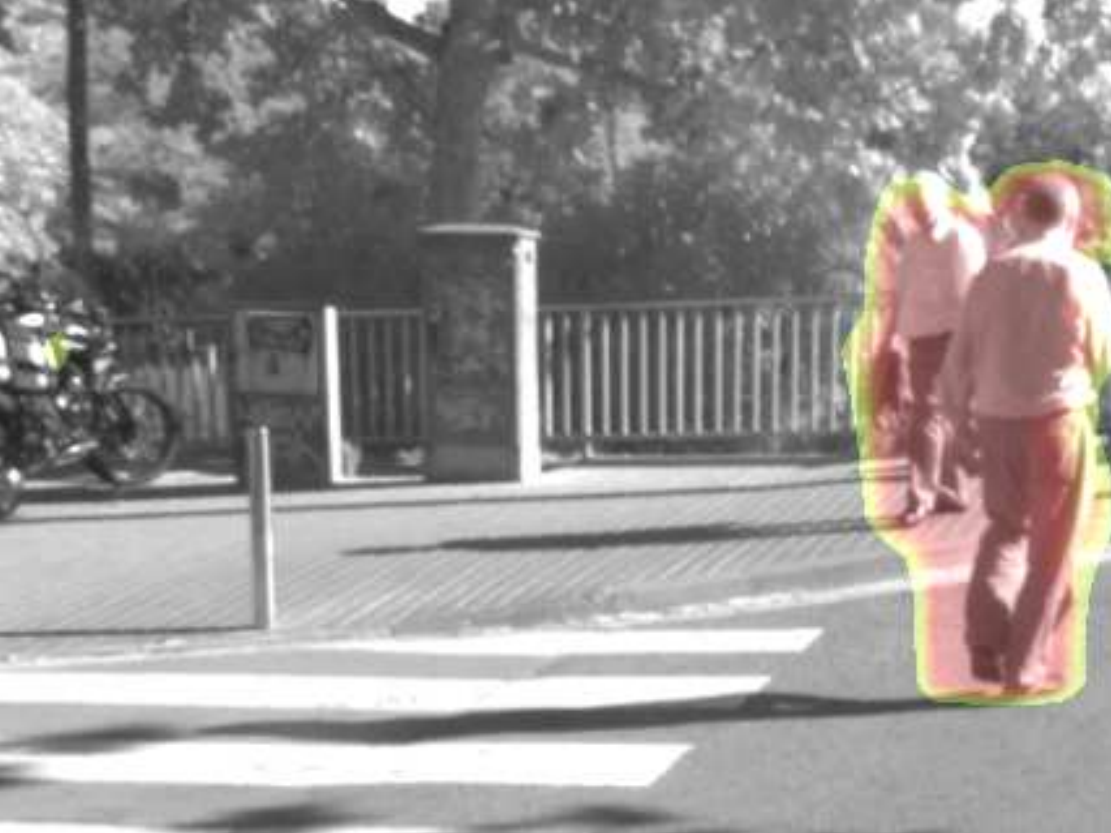}
		\end{minipage}
		\begin{minipage}{0.24\linewidth}
			\includegraphics[width=1\linewidth,trim=0 120 160 0,clip]{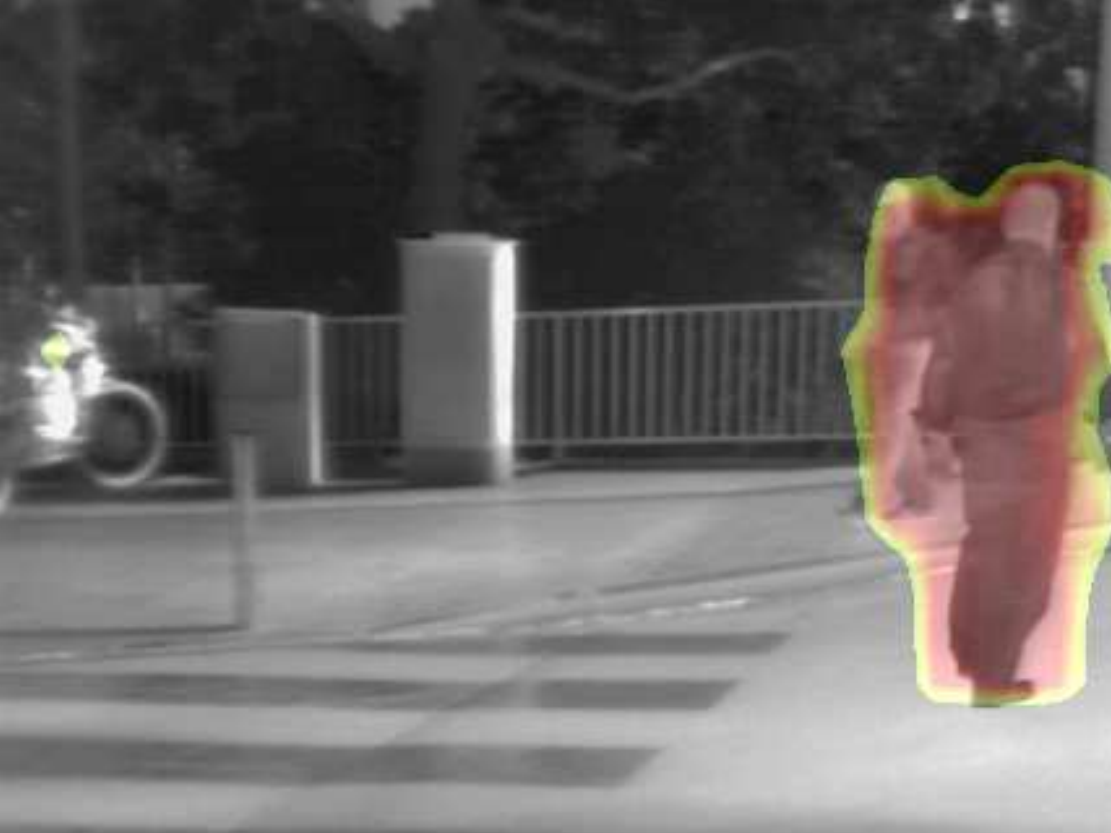}
		\end{minipage}
		\hspace{0.5mm}
		\begin{minipage}{0.24\linewidth}
			\includegraphics[width=1\linewidth,trim=0 120 160 0,clip]{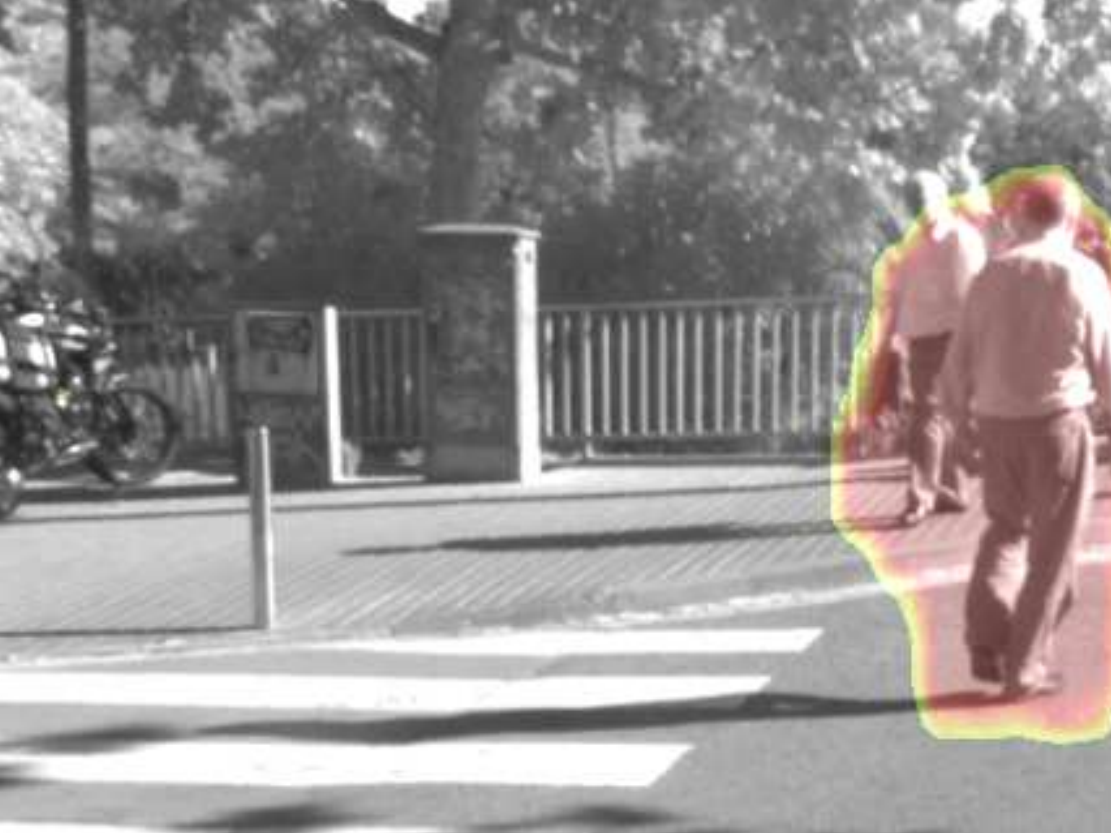}
		\end{minipage}
		\begin{minipage}{0.24\linewidth}
			\includegraphics[width=1\linewidth,trim=0 120 160 0,clip]{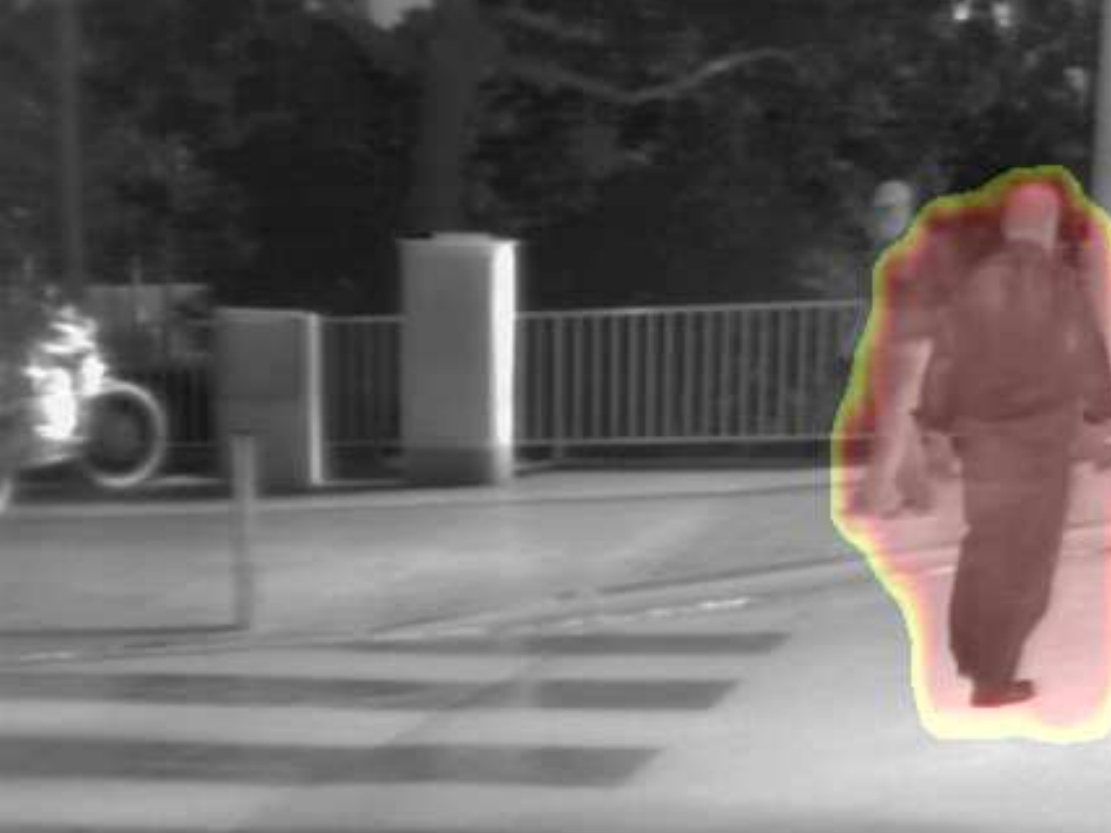}
		\end{minipage}
		\vspace{0.5mm}
	\end{minipage}
	\begin{minipage}{0.99\linewidth}
		\begin{minipage}{0.24\linewidth}
			\includegraphics[width=1\linewidth,trim=80 120 80 0,clip]{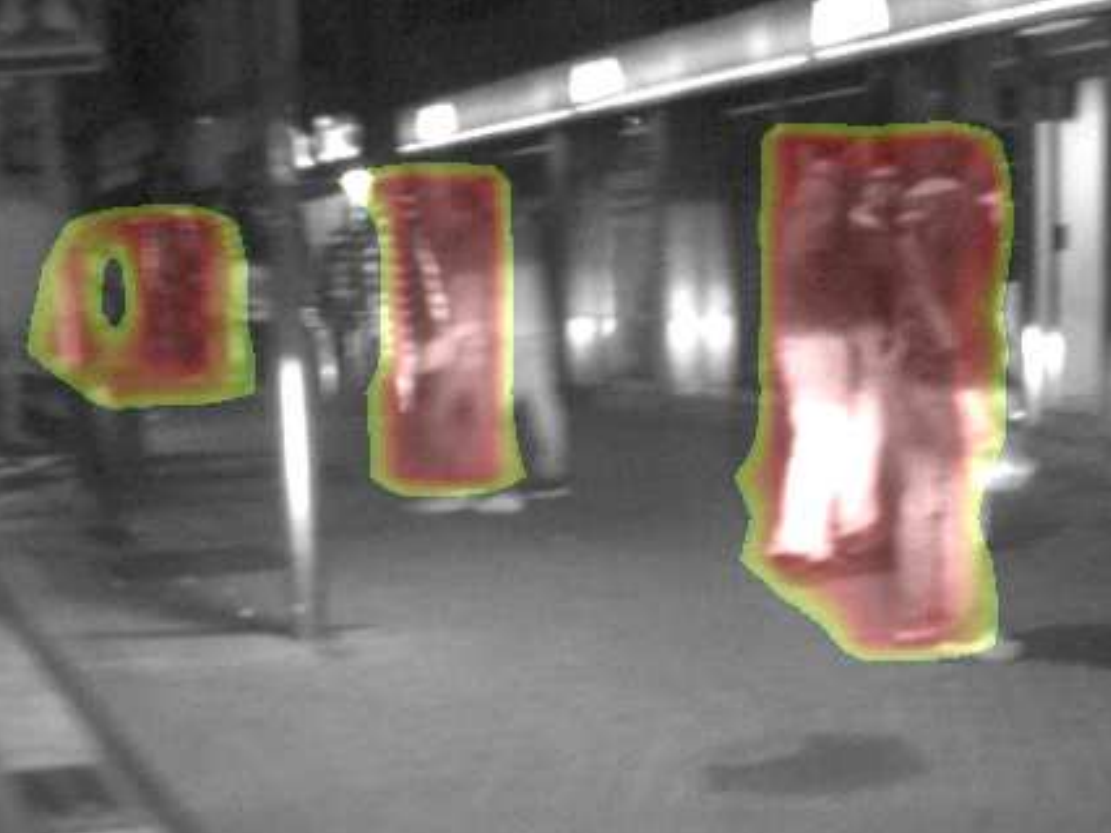}
		\end{minipage}
		\begin{minipage}{0.24\linewidth}
			\includegraphics[width=1\linewidth,trim=80 120 80 0,clip]{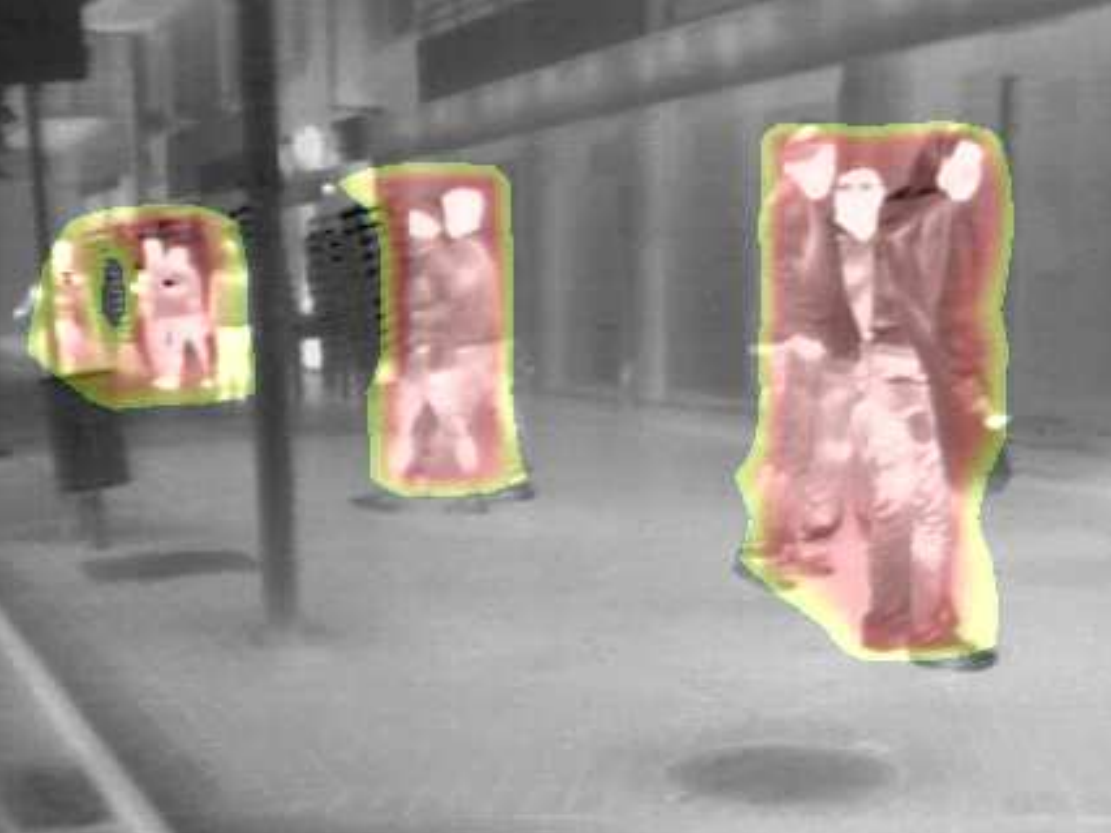}
		\end{minipage}
		\hspace{0.5mm}
		\begin{minipage}{0.24\linewidth}
			\includegraphics[width=1\linewidth,trim=80 120 80 0,clip]{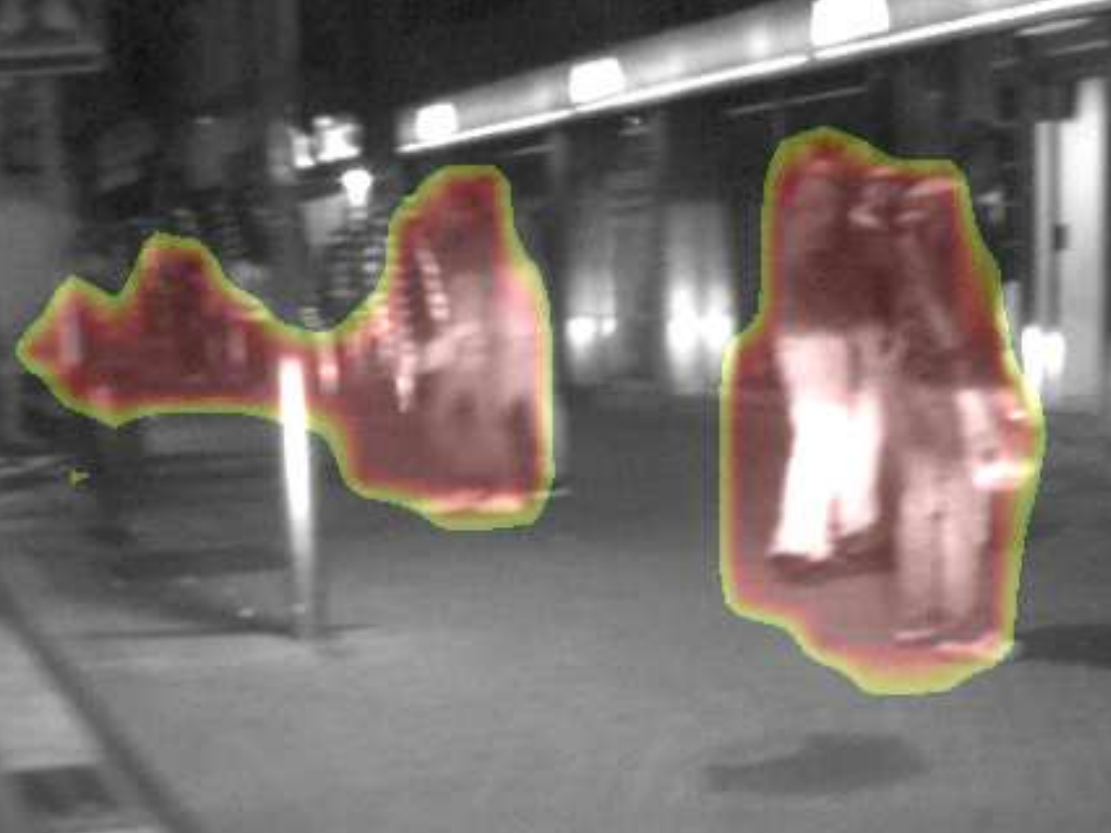}
		\end{minipage}
		\begin{minipage}{0.24\linewidth}
			\includegraphics[width=1\linewidth,trim=80 120 80 0,clip]{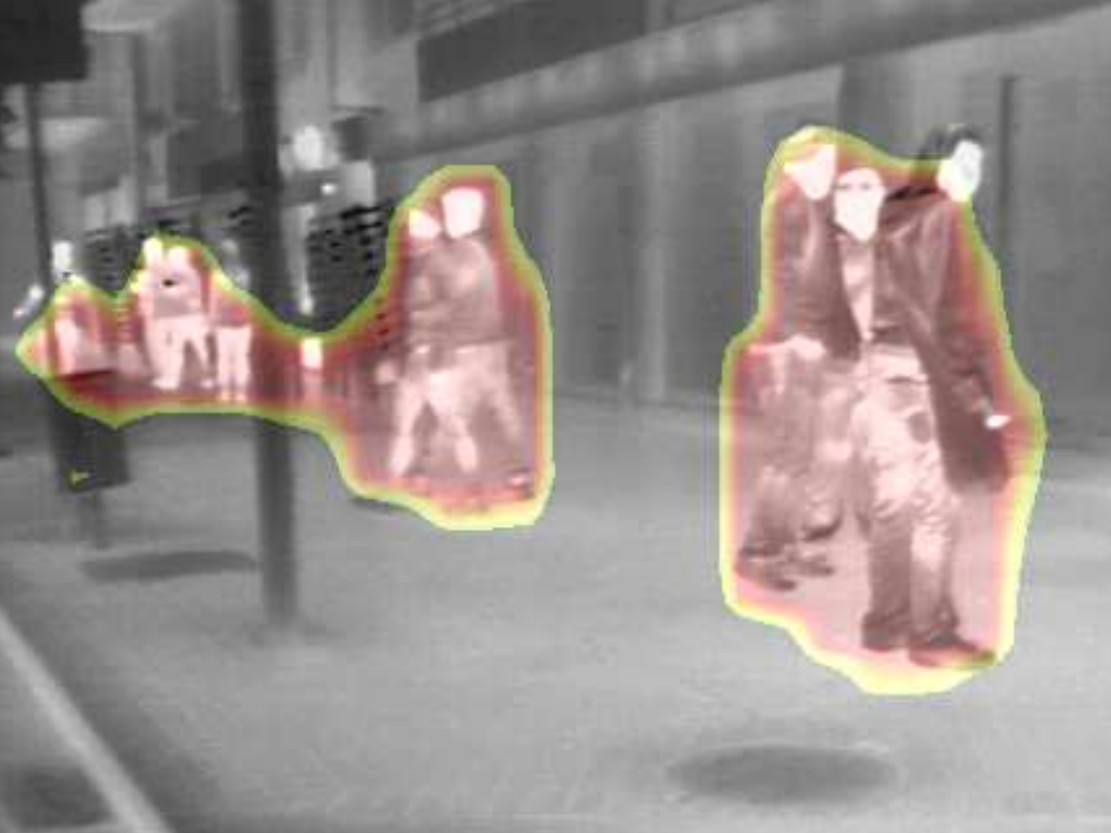}
		\end{minipage}
		\vspace{0.5mm}
	\end{minipage}
	\begin{minipage}{0.99\linewidth}
		\begin{minipage}{0.24\linewidth}
			\includegraphics[width=1\linewidth,trim=160 120 0 0,clip]{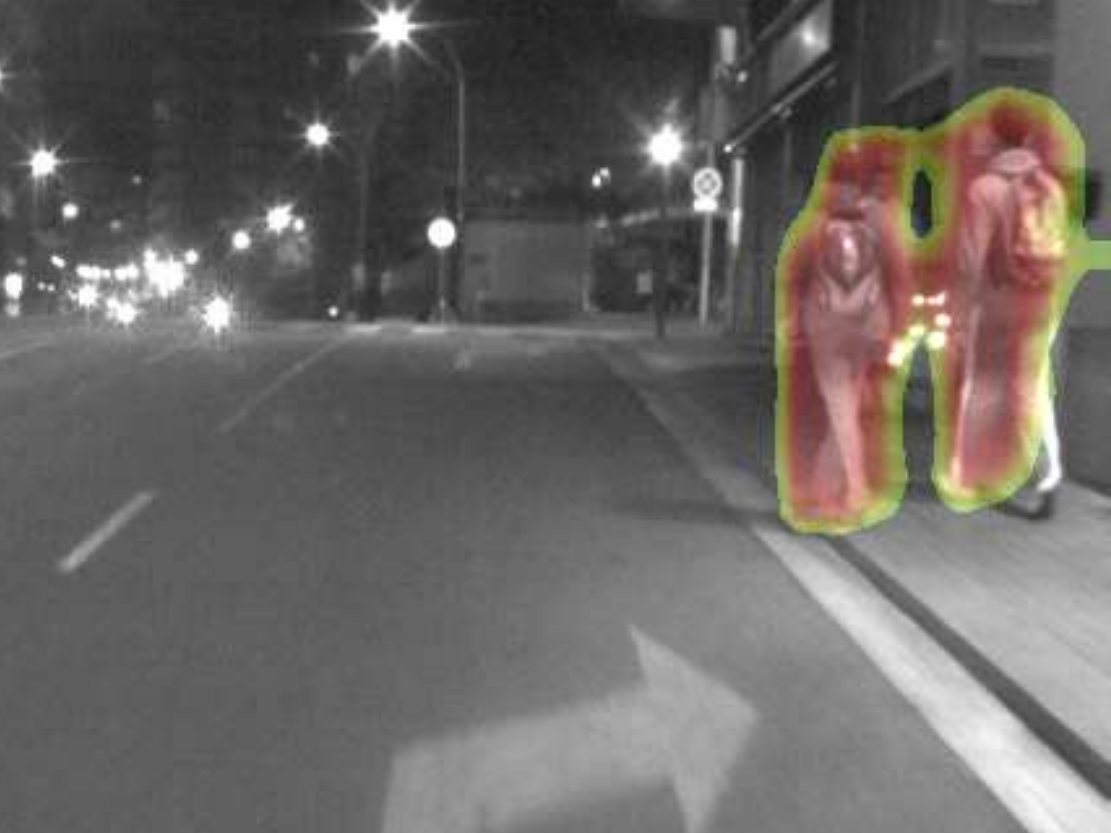}
		\end{minipage}
		\begin{minipage}{0.24\linewidth}
			\includegraphics[width=1\linewidth,trim=160 120 0 0,clip]{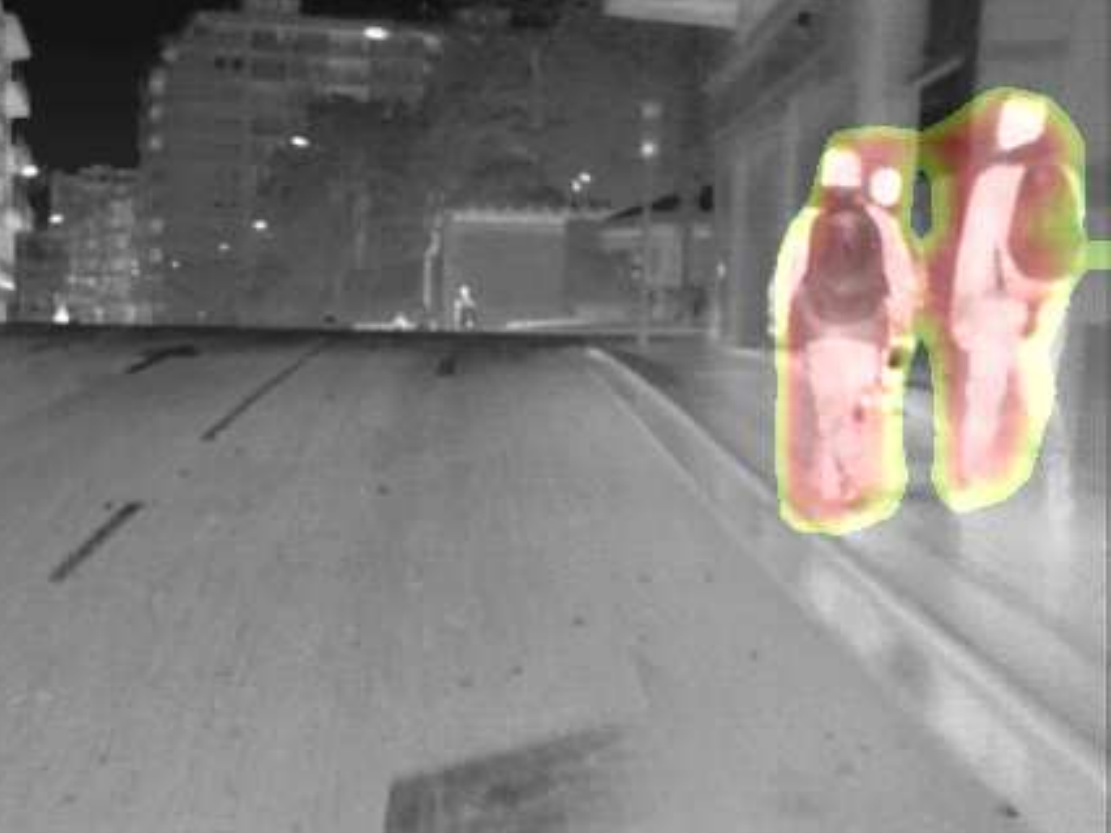}
		\end{minipage}
		\hspace{0.5mm}
		\begin{minipage}{0.24\linewidth}
			\includegraphics[width=1\linewidth,trim=160 120 0 0,clip]{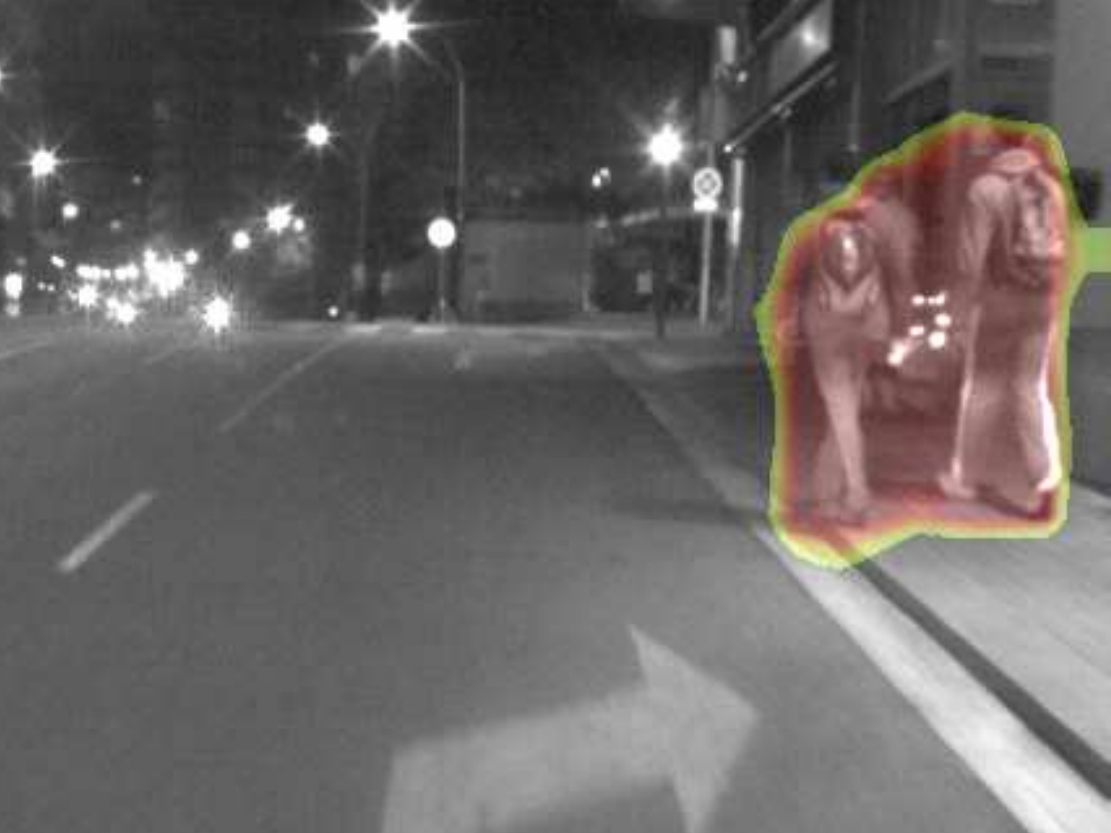}
		\end{minipage}
		\begin{minipage}{0.24\linewidth}
			\includegraphics[width=1\linewidth,trim=160 120 0 0,clip]{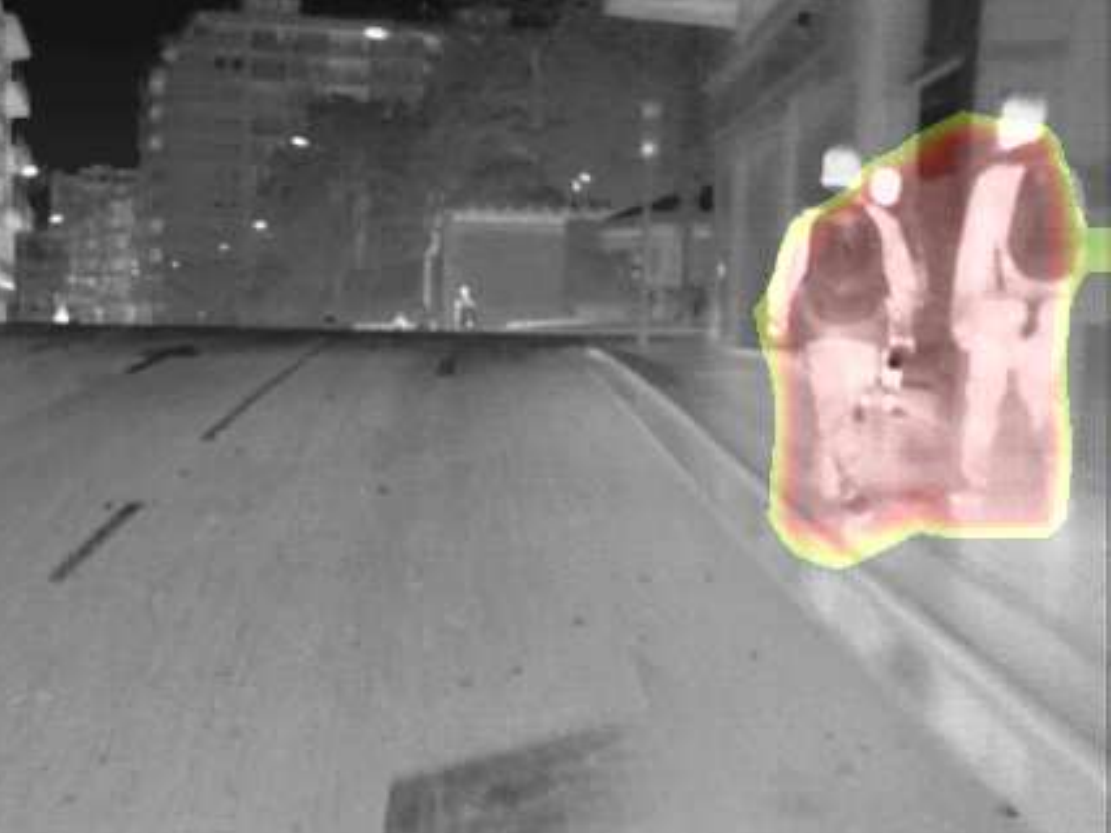}
		\end{minipage}
		\vspace{0.5mm}
	\end{minipage}
	\centering{(b)}
	\caption{Comparing the qualitative performance of multispectral pedestrian detections of UMDA and SMDA. (a) The results of UMDA is comparable with SMDA; (b) The results of UMDA is not satisfactory comparing with SMDA. }
	\label{fig6}
\end{figure}

In Tab.~\ref{tab2}, we observe that our proposed unsupervised multimodal domain adaptation (UMDA) model achieves competitive detection accuracy comparing with the supervised multispectral pedestrian detectors, pixel-level AP~\cite{cao2019box} of UMDA is 6.67\% lower than the results of SMDA and 8.84\% higher than the ones of ACF+T+THOG~\cite{hwang2015multispectral}. In order to investigate the gap between unsupervised and supervised multimodal domain adaptation models, we also compare the qualitative performance of multispectral pedestrian detection results of UMDA and SMDA in Fig.~\ref{fig6}. When human-related characteristics are distinct in either visible or thermal image as illustrated in Fig.~\ref{fig6}~(a), the multispectral pedestrian detection results of UMDA are comparable with the ones of SMDA. However, the UMDA may generate unsatisfactory results comparing with the SMDA when either the pedestrian samples appear indistinct or the background is clutter. Our future research will focus on enhancing the multispectral pedestrian detection methods to separate the human-related features with background ones. 

It should be mentioned that manual annotating of large-scale multispectral pedestrian dataset is extremely time-consuming. As mentioned in \cite{cao2019pedestrian}, it takes more than 80 hours to annotate the visible and infrared image pairs on the KAIST training dataset, which contains 50172 aligned visible and infrared sequential image pairs. Considering that the CVC-14 training dataset consist of 7085 aligned multispectral sequential image pairs, we consider that the annotating time is more than 11 hours. In comparison, our proposed unsupervised multimodal domain adaptation (UMAD) framework can be used to train the multispectral pedestrian detector without manual annotating effort. It is worth mentioning that comparing with the training procedure of supervised multispectral pedestrian detection approach, the additional processing time of our proposed UMAD method is the optimization of visible and thermal pseudo annotations without back-propagation, which barely increases the training time.

\section{Conclusion}

In this paper, we present an unsupervised multimodal domain adaptation (UMAD) framework for multispectral pedestrian detection, by iteratively generating pseudo annotations and updating the parameters of our designed multispectral pedestrian detector on target domain without manual annotating effort. Our proposed UMAD method achieves multispectral pedestrian detection performance significantly higher than the approach without multi-modal domain adaptation (pixel-level AP~\cite{cao2019box} of UMDA is 31.37\% higher than the results of WMDA) , and is competitive with the supervised multispectral pedestrian detectors (pixel-level AP~\cite{cao2019box} of UMDA is 6.67\% lower than the results of SMDA and 8.84\% higher than the ones of ACF+T+THOG~\cite{hwang2015multispectral}). It is worth mentioning that the training time of our proposed UMAD framework is barely the same as the training time of supervised approach. Our proposed method can be adapted to other multimodal computer vision tasks on unsupervised domain adaptation without manual annotating effort.

\section*{Acknowledgment}
The work is funded by DFG (German Research Foundation) YA 351/2-1, RO 4804/2-1 within SPP 1894, and the National Natural Science Foundation of China (No.51605428, No.51575486 and U1664264). The authors gratefully acknowledge the support.
The authors also acknowledge NVIDIA Corporation for the donated GPUs.

{\small
\bibliographystyle{ieee}
\bibliography{egbib}
}

\end{document}